\documentclass[conference]{IEEEtran}
\IEEEoverridecommandlockouts
\usepackage[utf8]{inputenc}

\usepackage{graphicx}
\usepackage{subcaption}
\usepackage{hyperref}
\usepackage{amsmath}
\usepackage{amsfonts}
\usepackage[version=4]{mhchem}
\usepackage{siunitx}
\usepackage{longtable,tabularx}
\usepackage{nicefrac}
\usepackage{placeins}
\usepackage{tikz}
\usepackage{makecell}
\newcommand{\vect}[1]{\boldsymbol{\mathbf{#1}}}
\def\realR{\mathbb{R}}
\def\sref{\operatorname{ref}}
\def\lpf{\operatorname{lpf}}
\def\hpf{\operatorname{hpf}}
\def\ext{\operatorname{ext}}
\def\ie{i.e.}
\def\eg{e.g.}
\newcommand{\ccos}[1]{\operatorname{c}\!{#1}\;}
\newcommand{\ssin}[1]{\operatorname{s}\!{#1}\;}
\newcommand{\coss}[1]{\cos{#1}\;}
\newcommand{\sinn}[1]{\sin{#1}\;}

\def\xelem{\vect{i}_x^\top}
\def\yelem{\vect{i}_y^\top}
\def\zelem{\vect{i}_z^\top}
\def\atan2{\operatorname{atan2}}

\usepackage{xcolor}

\newcommand{\revised}[1]{{#1}}
\newcommand{\rrevised}[1]{{#1}}

\title{Global Incremental Flight Control for\\ Agile Maneuvering of a Tailsitter Flying Wing}

\author{Ezra Tal and Sertac Karaman
	\thanks{E. Tal and S. Karaman are with the Laboratory for Information and Decision Systems (LIDS) and the Department of Aeronautics and Astronautics, Massachusetts Institute of Technology. 
		{\tt\footnotesize \{eatal, sertac\}@mit.edu}}%
}

\begin{document}

\maketitle

\begin{abstract}
This paper proposes a novel control law for accurate tracking of agile trajectories using a tailsitter flying wing unmanned aerial vehicle (UAV) that transitions between vertical take-off and landing (VTOL) and forward flight.
The global control formulation enables maneuvering throughout the flight envelope, including uncoordinated flight with sideslip.
Differential flatness of the nonlinear tailsitter dynamics with a simplified aerodynamics model is shown.
Using the flatness transform, the proposed controller incorporates tracking of the position reference along with its derivatives velocity, acceleration and jerk, as well as the yaw reference and yaw rate.
The inclusion of jerk and yaw rate references through an angular velocity feedforward term improves tracking of trajectories with fast-changing accelerations.
\rrevised{The controller does not depend on extensive aerodynamic modeling but instead uses incremental nonlinear dynamic inversion (INDI) to compute control updates based on only a local input-output relation, resulting in robustness against discrepancies in the simplified aerodynamics equations.
Exact inversion of the nonlinear input-output relation is achieved through the derived flatness transform.}
The resulting control algorithm is extensively evaluated in flight tests, where it demonstrates accurate trajectory tracking and challenging agile maneuvers, such as sideways flight and aggressive transitions while turning.
\end{abstract}
\section*{Supplemental Material}
Video of the experiments can be found on the project website \url{https://aera.mit.edu/projects/TailsitterAerobatics}.

\section{Introduction}

\revised{Transitioning powered-lift aircraft combine the vertical take-off and landing (VTOL) and hover capability of rotorcraft with the increased speed, range, and endurance of fixed-wing aircraft.
Tailsitter aircraft pitch down during transition, so that their rotors naturally shift from lift generation for take-off to propulsion for forward flight.}
While the large attitude envelope of tailsitters may render them less suitable for manned flight, their relative mechanical simplicity makes them an appealing option for unmanned aerial vehicle (UAV) applications.
\revised{Increased range and endurance with the capability to maneuver in confined spaces make these UAVs relevant to many applications.}
For example, in search and rescue, unmanned tailsitter aircraft could quickly reach remote locations using \revised{forward} flight, and inspect structures or enter buildings in hovering flight.

\begin{figure}
	\centering
	\includegraphics[trim={15em 30em 5em 20em},clip,width=\linewidth]{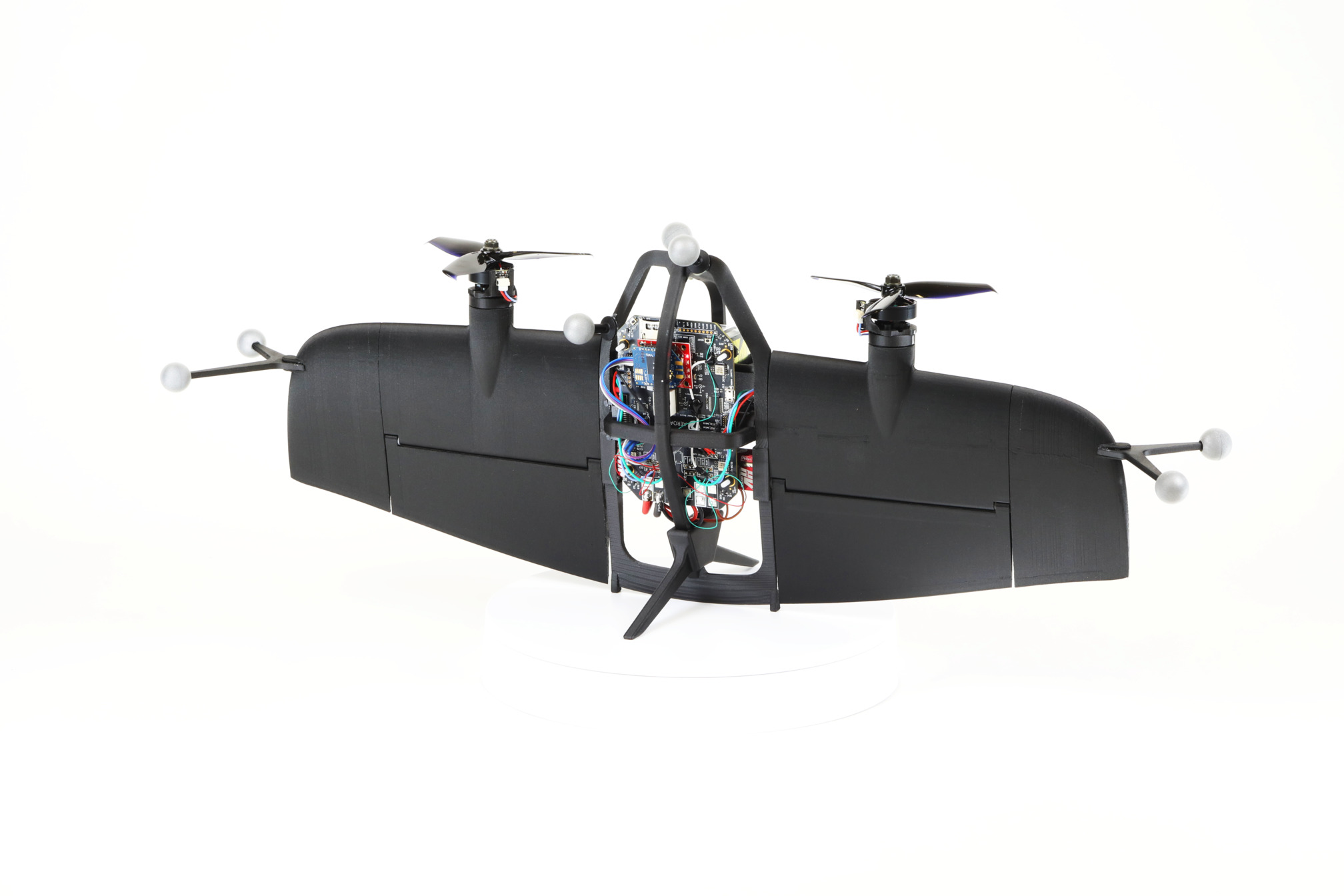}
	\caption{Tailsitter flying wing aircraft.}
	\label{fig:aircraft}
\end{figure}

A tailsitter flying wing is a tailsitter aircraft without fuselage, tail, and vertical stabilizers or control surfaces.
Forgoing these structures simplifies the aerodynamic and mechanical design of the aircraft and potentially improves performance by lowering mass and aerodynamic drag.
Due to the lack of vertical aerodynamic surfaces, flying wing aircraft often require active directional stabilization.
The fast and relatively powerful brushless motors found on many small UAVs are particularly suitable to fulfill this task through differential thrust.
By placing flaps that act as elevator \revised{(deflecting collectively)} and as aileron \revised{(deflecting differentially)}, \ie, \textit{elevons}, in the rotor wash, the aircraft remains controllable throughout its flight envelope, including static hover conditions.
The reduced stability of flying wing aircraft may also result in increased maneuverability.
Specifically, the lack of vertical surfaces enables maneuvers such as fast skidding turns and knife edge flight where the wing points in the direction of travel.
In general, it permits uncoordinated flight, where the vehicle incurs nonzero lateral velocity.

In this paper, we propose a novel flight control algorithm that is specifically designed for tracking of agile trajectories using the tailsitter flying wing aircraft shown in Fig. \ref{fig:aircraft}.
The proposed controller uses differential flatness to track the reference position, velocity, acceleration, and jerk (the third derivative of position), as well as yaw angle and yaw rate.
It is based on a global formulation, without mode switching or blending, and able to exploit the entire flight envelope, including uncoordinated flight conditions, for agile maneuvering.
We derive the controller based on a simplified aerodynamics model and apply incremental nonlinear dynamic inversion (INDI) to achieve accurate trajectory tracking despite model discrepancies.

We use $\varphi$-theory to model the aerodynamic force and moment~\cite{lustosa2019global}.
The method captures dominant contributions over the entire flight envelope, including post-stall and uncoordinated flight conditions\rrevised{, in a single, global formulation.}
The $\varphi$-theory model does not suffer from singularities that methods based on the angle of attack and sideslip angle may incur around hover, where these angles are not defined.

Incremental, or sensor-based, nonlinear dynamic inversion is a version of nonlinear dynamic inversion (NDI) control~\cite{snell1992nonlinear} that alleviates the lack of robustness associated with NDI~\cite{papageorgiou2005robustness,lee2009feedback,wang2019stability} by incrementally updating control inputs based on inertial measurements~\cite{smith1998simplified,bacon2000reconfigurable}.
Instead of directly computing control inputs from the inverted dynamics model, it only considers the input-output relation around the current operating point and computes the required control increment relative to this point~\cite{indi,smeur2015adaptive}.
As such, it only relies on local accuracy of the dynamics model and can correct for discrepancies by further incrementing control inputs in subsequent updates.

Differential flatness is a property of nonlinear dynamics systems that guarantees the existence of an equivalent controllable linear system~\cite{fliess1992lessystemesnon,fliess1993linearisation, fliessintro}.
The state variables and control inputs of a flat system can be expressed as a function of a (ficticious) flat output and a finite number of its time-derivatives.
Using this function, trajectories can be generated in the flat output space and transformed to the state space for tracking control~\cite{martin1996aircraft, martin2003flat}.
This enables tracking higher-order derivatives of the output, which has been demonstrated to improve trajectory-tracking performance in fast and agile flight~\cite{ferrin2011differential,rivera2010flatness, faessler2018differential,tal2020accurate}.
The differential flatness property has been shown to hold for idealized aircraft dynamics~\cite{martin1992contribution} and for aggressive fighter maneuvers in coordinated flight~\cite{hauser1997aggressive}.

Existing flight control designs for tailsitter aircraft are based on various approaches.
Blending of separate controllers~\cite{ke2018design,forshaw2014transitional}, gain scheduling~\cite{jung2012development,lustosa2017varphi}, or pre-planned transition maneuvers~\cite{chiappinelli2018modeling} can be used to handle the change of dynamics between hover and forward flight.
However, when performing agile maneuvering at large angle of attack, the aircraft continuously enters and exits the transition regime, and it is preferable to utilize a controller without blending or switching~\cite{hartmann2017unified}.
A global formulation for trajectory tracking in coordinated flight is proposed by \cite{ritz2017global}.
The controller is based on numerical inversion of a global first-principles model, but does not account for model discrepancies, leading to a systematic pitch tracking error.
Wind tunnel testing can be used to improve accuracy of the dynamics model~\cite{lyu2018simulation,verling2016full}.
However, building an accurate model from measurements can be a time-consuming process that may need to be repeated if the controller is transferred to a different vehicle.

Robustification can be used to design a performant controller that does not rely on an accurate model of the vehicle dynamics, \eg, by using model-free control~\cite{barth2020model}.
INDI has also been leveraged for robust control of various types of transitioning aircraft, such as tiltrotor aircraft~\cite{di2015modeling, raab2018proposal} and aircraft with dedicated lift rotors~\cite{liu2018vtol, lombaerts2020dynamic, pfeifle2021energy}.
The additional control inputs of these configurations may lead to over-actuation, requiring specific attention to control allocation that is typically not necessary for tailsitter aircraft.
On the other hand, tailsitter aircraft operate over a more extensive attitude range and at increased angle of attack, leading to distinct challenges in control design.
Robustness of an INDI longitudinal flight controller for tracking pre-designed transition maneuvers is shown through analysis and simulation by \cite{yang2020indi}, and an INDI attitude control design for a tailsitter with varying fuselage shape is proposed by \cite{xia2021transition}.
An exhaustive framework for INDI-based linear and angular acceleration control of a tailsitter is presented by \cite{smeur2020incremental}.
This work includes flight test results, focusing on coordinated flight at small flight path angles.

The trajectory generation algorithm by \cite{mcintosh2021optimal} utilizes differential flatness of a simplified longitudinal dynamics model to design transitions for a quadrotor biplane.
The resulting trajectories are limited to forward motion and consider acceleration but no higher-order derivatives.
The algorithm by \cite{verling2016full} employs a pre-designed constant angular velocity feedforward input to improve transition.
Theoretically, this feedforward signal corresponds to the acceleration rate of change, \ie, jerk.
However, it is not applied beyond the pre-designed transition maneuver.

The main contribution of this paper is a global control design for tracking agile trajectories using a flying wing tailsitter.
Our proposed control design is novel in several ways.
Firstly, we derive a differential flatness transform for the tailsitter flight dynamics with simplified $\varphi$-theory aerodynamics model.
Secondly, we present a method to incorporate jerk tracking as an angular velocity feedforward input in tailsitter control design.
As far as we are aware, this is the first tailsitter controller that achieves jerk tracking, making it suitable to fly agile trajectories with fast-changing acceleration references.
Thirdly, we apply INDI to control a tailsitter aircraft in agile maneuvers that include large flight path angles and uncoordinated flight conditions.
Our control design is based on direct nonlinear inversion and, contrary to existing \revised{INDI} implementations, does not rely on \revised{local} linearization of the dynamics for inversion.
Fourthly, we detail our methodology for analytical and experimental estimation of the $\varphi$-theory aerodynamic parameters used by the controller.
Finally, we demonstrate the proposed controller in extensive flight experiments reaching up to 8 m/s in an indoor flight space measuring 18 m $\times$ 8 m.
The flight experiments include agile maneuvers, such as aggressive transitions while turning, differential thrust turning, and uncoordinated flight.
In order to explicitly show the advantages provided by the aforementioned contributions, we also present an experimental comparison to a baseline controller.

The paper is structured as follows: 
An overview of the main nomenclature is given in Table \ref{tab:ts_nomenclature}.
In Section \ref{sec:model}, we provide an overview of the flight dynamics and aerodynamics model.
We derive the corresponding differential flatness transform in Section \ref{sec:invdyn}.
The design of the trajectory-tracking controller is presented in Section \ref{sec:control}.
Section \ref{sec:paramest} details our methodology for analytical and experimental estimation of the aerodynamic parameters used by the controller. 
Extensive experimental flight results are presented in Section \ref{sec:experiments}.
Additional details on the flatness transform and an experimental comparison that explicitly shows the advantages of incremental control and feedforward references can be found in Appendix \ref{app:snap} and Appendix \ref{sec:baselinecomp}, respectively.
An initial version of this paper, not including Section \ref{sec:paramest} and the appendices, was presented at AIAA Aviation Forum 2021~\cite{tal2021global}.
\begin{table*}
	\centering
	\caption{%
		Main nomenclature.}
	\label{tab:ts_nomenclature}

{\footnotesize
		\begin{tabular}{ll}
			\hline
			$\vect{a}$& linear coordinate acceleration (in world-fixed frame, unless noted otherwise), {m/s\textsuperscript{2}}\\
			$\vect{b}_x$, $\vect{b}_y$, $\vect{b}_z$& basis vectors of body-fixed frame\\
			$c$ & propulsion and $\varphi$-theory aerodynamic coefficients\\
			$\bar c$ & Buckingham $\pi$ aerodynamic coefficients\\
			$\vect{f}$ & force vector (in world-fixed frame, unless noted otherwise), {N}\\
			$g$& gravitational acceleration, {m/s\textsuperscript{2}}\\
			$\vect{i}_x$, $\vect{i}_y$, $\vect{i}_z$& standard basis vectors\\
			$\vect{j}$& jerk in world-fixed frame, {m/s\textsuperscript{3}}\\	
			$\vect{J}$ & vehicle moment of inertia tensor, {kg m\textsuperscript{2}}\\
			$\vect{K}$& diagonal control gain matrix\\		
			$l$ & flap/rotor location\\
			$m$& vehicle mass, {kg}\\
			$\vect{m}$ & moment vector in body-fixed frame, {Nm}\\
			$\vect{q}$ & throttle input\\
			$\vect{R}^i_b$& rotation matrix from frame $b$ to frame $i$\\
			$T$ & total thrust, {N}\\
			$T_i$ & thrust by rotor $i$, {N}\\
			$\vect{v}$& velocity (in world-fixed frame, unless noted otherwise), {m/s}\\
			$\vect{x}$& position in world-fixed frame, {m}\\
			$\bar \alpha$ & sum of zero-lift angle of attack and thrust angle, {rad}\\
			$\alpha_0$& zero-lift angle of attack, {rad}\\ 
			$\alpha_T$& thrust angle, {rad}\\
			$\vect{\alpha}_x$, $\vect{\alpha}_y$, $\vect{\alpha}_z$& basis vectors of zero-lift frame\\
			$\delta$ & sum of flap deflections, {rad}\\
			$\delta_i$ & deflection of flap $i$, {rad}\\
			$\Delta T$ & differential thrust, {N}\\
			$\theta$& vehicle pitch angle, {rad}\\
			$\bar \theta$& zero-lift reference frame pitch angle, {rad}\\
			$\mu_i$ & torque by rotor $i$, {Nm}\\
			$\vect{\xi}$&normed quaternion attitude vector\\
			$\vect{\sigma}_{\sref}$ & reference trajectory\\
			$\phi$& vehicle roll angle, {rad}\\
			$\psi$& vehicle yaw angle, {rad}\\
			$\omega_i$& angular speed of rotor $i$, {rad/s}\\
			$\vect{\Omega}$& vehicle angular velocity in body-fixed frame, \si{rad/s}\\
			&\\
			\multicolumn{2}{l}{\textit{Subscript and superscript}}\\
			$b$ & body-fixed reference frame\\
			$c$ & control command\\
			$i$ & world-fixed reference frame, or flap/rotor index\\
			$T$ & thrust force or moment\\
			$w$ & wing force or moment\\
			$\alpha$ & zero-lift reference frame\\
			$\delta$ & flap force or moment\\
			$\bar\theta$ & intermediate control reference frame after pitch rotation \\
			$\phi$ & intermediate control reference frame after roll rotation\\
			$\psi$ & intermediate control reference frame after yaw rotation\\
			$\ext$ & unmodeled force or moment\\
			$\hpf$ & high-pass filtered signal\\
			$\lpf$ & low-pass filtered signal \\
			$\sref$ & (derived from) reference trajectory\\
			\hline

		\end{tabular}
			}
\end{table*}

\section{Flight Dynamics Model}\label{sec:model}
This section provides a detailed overview of the flight dynamics model employed in our proposed control algorithm.
The algorithm, described in Section \ref{sec:control}, is based on the notion of incremental control action and therefore utilizes the dynamics model solely as a local approximation of the flight dynamics.
Unlike conventional inversion-based controllers, it does not require a globally accurate dynamics model.

The model is employed by the incremental controller to predict (i) the change in linear acceleration due to increments in attitude and collective thrust, and (ii) the change in angular acceleration due to increments in differential thrust and flap deflections.
By inversion of these relationships, the control algorithm computes the increments required to attain the commanded changes in linear and angular acceleration.
In order to maintain analytical invertibility and avoid undue complexity, the dynamics model omits any contributions that do not directly affect the aforementioned incremental relationships.
For example, the velocity of the aircraft relative to the atmosphere may result in a significant aerodynamic moment.
However, when compared to the fast dynamics of the motors and servos controlling the propellers and flaps, this moment contribution is relatively slow-changing, as it relates to the orientation and velocity of the entire vehicle.
Consequently, it is assumed to be constant between control updates and does not need to be included in the incremental dynamics model.
It is nonetheless accounted for in the control algorithm together with other unmodeled contributions to linear and angular acceleration through inertial measurement feedback, as described in Section \ref{sec:control}.

\subsection{Reference Frame Conventions}
\begin{figure*}[t!]
	\centering
	\begin{subfigure}[t]{0.485\textwidth}
		\centering
	\includegraphics[trim={12em 0em 12em 0em},clip,width=\linewidth]{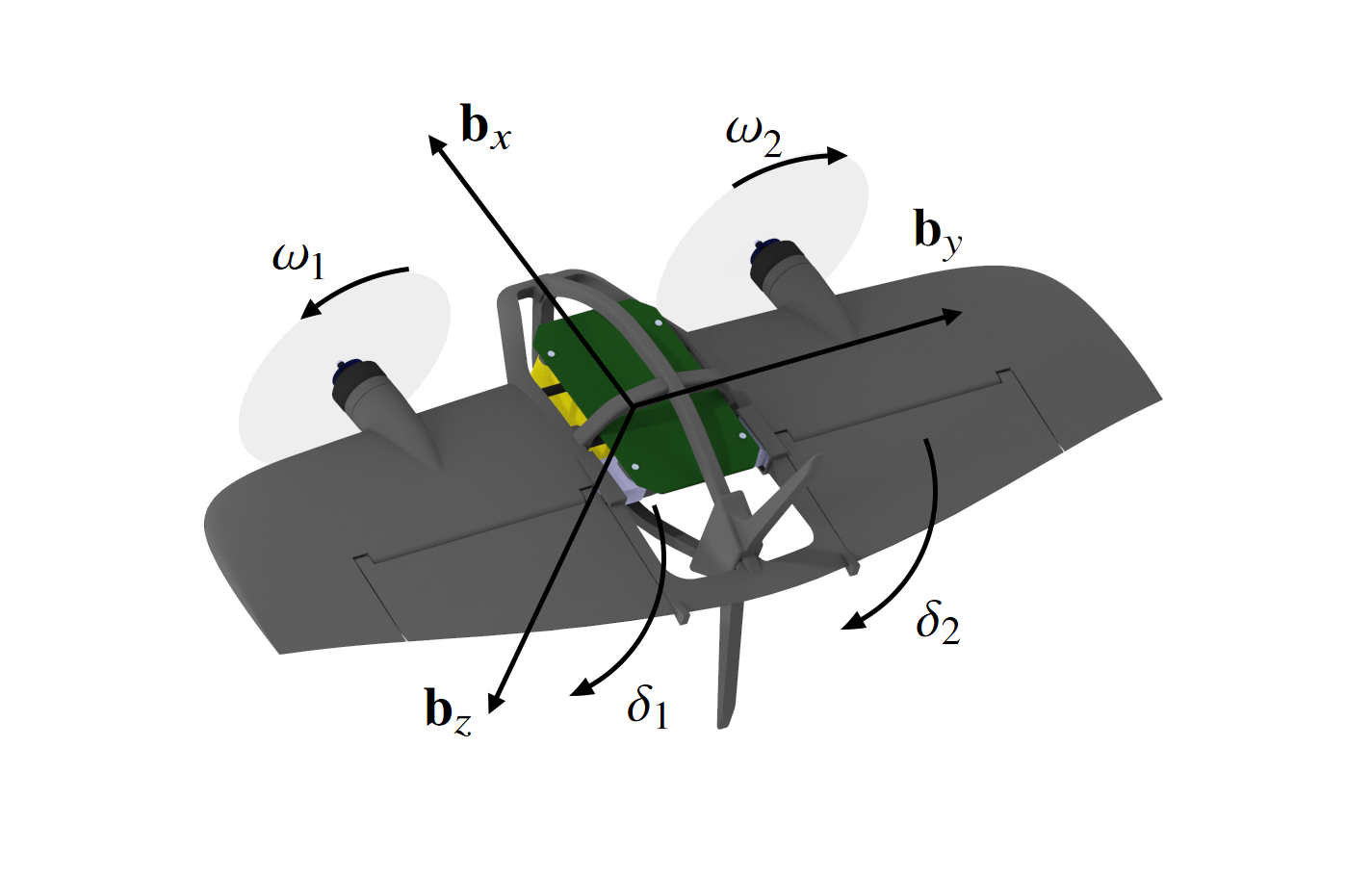}
	\caption{\revised{Body-fixed reference frame $\vect{b}$, and control inputs, \ie, rotor speeds $\omega_1$ and $\omega_2$, and flap deflections $\delta_1$ and $\delta_2$.}}
	\label{fig:bodyref_system}
	\end{subfigure}%
	\quad
	\begin{subfigure}[t]{0.485\textwidth}
		\centering
	\includegraphics[trim={18em 0em 12emem 0em},clip,width=\linewidth]{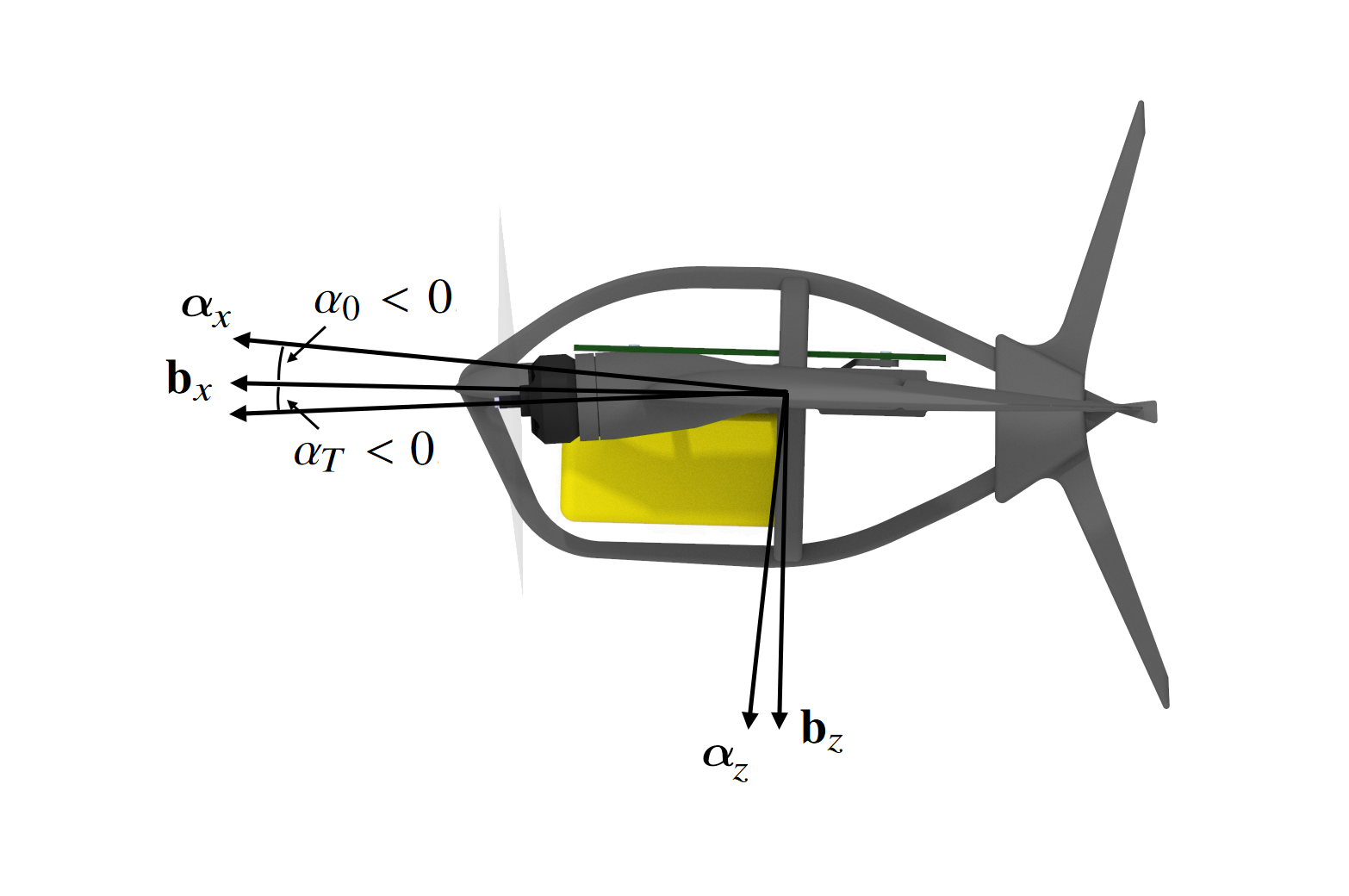}
	\caption{\revised{Zero-lift reference frame $\vect{\alpha}$, zero-lift angle of attack $\alpha_0$, and thrust angle $\alpha_T$.}}
	\label{fig:zeroliftref_system}
	\end{subfigure}
	\caption{Reference frame and control input conventions.}
	\label{fig:ref_systems}
\end{figure*}

We define the basis of the body-fixed reference frame\rrevised{, shown in Fig. \ref{fig:bodyref_system},} as the vectors
$\vect{b}_x$, which coincides with the chord line and the wing symmetry plane;
$\vect{b}_y$, which is perpendicular to this symmetry plane; and
$\vect{b}_z$, which is defined to satisfy the right-hand rule.
These vectors form the rotation matrix $\vect{R}_b^i=[
\vect{b}_x\;\vect{b}_y\;\vect{b}_z
] \in SO(3)$, which gives the transformation from the body-fixed reference frame (indicated by the subscript $b$) to the world-fixed north-east-down (NED) reference frame (indicated by the superscript $i$) \rrevised{consisting of the columns of the identity matrix $[
\vect{i}_x\;\vect{i}_y\;\vect{i}_z]$.}
The zero-lift axis system, depicted in Fig. \ref{fig:zeroliftref_system}, is obtained by rotating the body-fixed axis system around its negative $\vect{b}_y$-axis by the zero-lift angle of attack $\alpha_0$, which is defined as the angle of attack for which the aircraft produces zero lift.
For symmetric airfoils $\alpha_0 = 0$, while most cambered airfoils have $\alpha_0 < 0$.
Finally, the thrust angle $\alpha_T$ is defined as the angle of the thrust line with regard to the $\vect{b}_x$-$\vect{b}_y$ plane.
\revised{The motors may be slightly tilted down to obtain a horizontal thrust vector in forward flight with positive angle of attack, leading to $\alpha_T < 0$.}

\subsection{Vehicle Equations of Motion}\label{sec:eom}
\revised{We apply the Newton-Euler equations to describe the vehicle translational and rotational dynamics.
Their translational component is given by}
\begin{align}
\dot{\vect{x}} &= \vect{v},\\
\dot{\vect{v}} &= g\vect{i}_z + m^{-1}\left(\vect{R}^i_\alpha\vect{f}^\alpha + \vect{f}_{\ext}\right),\label{eq:vdot}
\end{align}
where $\vect{x}$ and $\vect{v}$ are respectively the position and velocity \revised{of the vehicle center of mass} in the world-fixed reference frame, $g$ is the gravitational acceleration, and $m$ is the vehicle mass.
The vector $\vect{f}^\alpha$ represents the modeled aerodynamic and thrust force in the zero-lift reference frame.
Any unmodeled forces are represented by the external force vector $\vect{f}_{\ext}$, which is defined in the world-fixed reference frame.

The rotational dynamics are given by
\begin{align}
\dot{\vect{ \xi}} &= \frac{1}{2}\vect{\xi}\circ\vect{\Omega},\\
\dot{\vect{\Omega}} &= \vect{J}^{-1}(\vect{m} + \vect{m}_{\ext}-\vect{\Omega} \times \vect{J}\vect{\Omega}), \label{eq:Omegadot}
\end{align}
where
$\vect{\Omega}$ is the angular velocity in the body-fixed reference frame, and
$\vect{\xi}$ is the normed quaternion attitude vector.
The symbol $\circ$ denotes the Hamilton quaternion product~\cite{hamilton1866elements}, such that $\vect{v}^b = \vect{R}^b_i \vect{v} = \vect{\xi}^{-1}\circ\vect{v}\circ\vect{\xi}$.
The matrix $\vect{J}$ is the vehicle moment of inertia tensor, and $\vect{m}$ represents the aerodynamic and thrust moment in the body-fixed reference frame.
\revised{The final term of \eqref{eq:Omegadot} accounts for the rotation of the body-fixed reference frame.}
The external moment vector $\vect{m}_{\ext}$ represents unmodeled moment contributions, similar to the force vector $\vect{f}_{\ext}$.
We note that the term \textit{external} with regard to $\vect{f}_{\ext}$ and $\vect{m}_{\ext}$ refers to unmodeled force and moment contributions, \ie, external to the model, but not necessarily due to physically external influences, such as gusts.

\subsection{Aerodynamic and Thrust Force and Moment}\label{sec:aeromodel}
We employ $\varphi$-theory parametrization to model the aerodynamic force and moment~\cite{lustosa2019global}.
This parametrization has several advantages over standard expressions for aerodynamic coefficients.
Firstly, it provides a simple global model that includes dominant contributions over the entire flight envelope, including post-stall conditions.
Our simplified model relies on only nine scalar aerodynamic coefficients: two for the wing, two for the flaps, two for the propellers, and three for propeller-wing interaction.
\revised{We use experimental flight data to determine these coefficients, leading to improved model accuracy, as described in Section \ref{sec:paramest}.}
Secondly, $\varphi$-theory parametrization avoids the singularities that methods based on angle of attack and sideslip angle incur near hover conditions, where these angles are undefined.

We obtain the force in the zero-lift axis system by summing contributions due to thrust, flaps, and wings.
The thrust force is given by
\begin{equation}\label{eq:force_thrust}
\vect{f}^\alpha_T = \sum_{i=1}^2 \underbrace{\left[\begin{array}{c}
\coss{\bar \alpha} (1 - c_{D_T})\\
0\\
\sinn{\bar \alpha}(c_{L_T} - 1)
\end{array}\right]T_i}_{\vect{f}^\alpha_{T_i}},
\end{equation}
\revised{where $\bar \alpha$ is the sum of the zero-lift angle of attack and the thrust angle (\ie, $\bar \alpha = \alpha_0 + \alpha_{T}$),} $T_i$ is the thrust due to motor $i$, and the coefficients $c_{D_T}$ and $c_{L_T}$ represent drag and lift due to thrust vector components in the zero-lift axis system, respectively.
The motor thrust is \revised{obtained from the following quadratic approximation:}
\begin{equation}\label{eq:thrust}
T_i = c_T \omega_i^2\;\;\;\text{with}\;\;i = 1, 2,
\end{equation}
where $\omega_i \geq 0$ is the speed of motor $i$.
The thrust coefficient $c_T$ is a function of propeller geometry and can be obtained from bench tests using a force balance.
Intuitively, $c_{D_T}$ mostly represents the loss of propeller efficiency due to the presence of the wing in the propwash, while $c_{L_T}$ represents the propwash-induced lift.
Note that the lift component vanishes if the thrust line coincides with the zero-lift axis, \ie, $\alpha_0 + \alpha_T = 0$.
For convenience, all aerodynamic coefficients incorporate the effects of air density.
If the coefficients are applied for flight in significantly varying conditions,
their values may be recomputed using a simple scaling with air density, as shown in Section \ref{sec:paramest}.
The force contribution by the flaps is given by
\begin{equation}\label{eq:force_flap}
\vect{f}^\alpha_\delta = \sum_{i=1}^{2}\underbrace{-\left[\begin{array}{c}
0\\
0\\
c^\delta_{L_T} \coss{\bar\alpha} T_i + c^\delta_{L_V}\|\vect{v}\| \xelem\vect{v}^\alpha
\end{array}\right]\delta_i}_{\vect{f}^\alpha_{\delta_i}},
\end{equation}
where $\delta_i$ is the deflection angle of flap $i$\revised{, the superscript $\top$ indicates the transpose, and $\|\cdot\|$ is the Euclidean norm.}
The first term of \eqref{eq:force_flap}, scaled with the coefficient $c^\delta_{L_T}$, is the flap lift due to the prop-wash induced airspeed. The second term, scaled with $c^\delta_{L_V}$, is the flap lift due to the airspeed along the zero-lift line.
Finally, the wing force contribution is obtained as
\begin{equation}\label{eq:force_body}
\vect{f}^\alpha_w = -\left[\begin{array}{c}
c_{D_V}\xelem \vect{v}^\alpha\\
0\\
c_{L_V}\zelem\vect{v}^\alpha
\end{array}\right]\|\vect{v}\|,
\end{equation}
where $c_{D_V}$ and $c_{L_V}$ are the wing drag and lift coefficients, respectively.
The total force in the zero-lift axis system is now obtained as
\begin{equation}\label{eq:force_total}
\vect{f}^\alpha = \vect{f}^\alpha_T + \vect{f}^\alpha_\delta +\vect{f}^\alpha_w.
\end{equation}
We note that \eqref{eq:force_total} does not contain any lateral force component.
Due to the lack of a fuselage and vertical tail surface, the lateral force is relatively much smaller than the lift and drag components.
Any incurred lateral force is captured in the unmodeled force $\vect{f}_{\ext}$, and accounted for by the controller through accelerometer feedback, as described in Section \ref{sec:linacc_control}.

The moment is obtained by summation of contributions due to motor thrust and torque, and flap deflections.
We ignore the wing moment due to velocity, attitude, and rotation rates, as these state variables are relatively slow-changing compared to the motor speeds and flap deflections.
The corresponding contributions are incorporated in the unmodeled moment $\vect{m}_{\ext}$ and accounted for through angular acceleration feedback.
The moment due to motor thrust is given by
\begin{equation}
\vect{m}_T = \left[\begin{array}{c}
l_{T_y} \zelem \vect{R}^b_\alpha(\vect{f}^\alpha_{T_2} - \vect{f}^\alpha_{T_1})\\
c_{\mu_T} (T_1 + T_2)\\
l_{T_y} \xelem \vect{R}^b_\alpha(\vect{f}^\alpha_{T_1} - \vect{f}^\alpha_{T_2})
\end{array}\right],
\end{equation}
where $l_{T_y}$ is the absolute distance along $\vect{b}_y$ between the vehicle center of gravity and each motor, and $c_{\mu_T}$ is the pitch moment coefficient due to thrust.
The moment due to motor torque is obtained as follows:
\begin{equation}
\vect{m}_\mu = \left[\begin{array}{c}
\cos\alpha_T\\
0\\
-\sin\alpha_T
\end{array}\right]\sum_{i=1}^2 \mu_i,
\end{equation}
where $\mu_i$ is the motor torque around the thrust-fixed $x$-axis given by
\begin{equation}\label{eq:motortorque}
\mu_i = -(-1)^i c_\mu \omega_i^2
\end{equation}
with $c_\mu$ the propeller torque coefficient. The signs in \eqref{eq:motortorque} correspond to the rotation directions defined in Fig. \ref{fig:zeroliftref_system}.
The flap contribution to the aerodynamic moment is given by
\begin{equation}\label{eq:moment_flap}
\vect{m}_\delta = \left[\begin{array}{c}
l_{\delta_y} \coss{\alpha_0}\zelem (\vect{f}^\alpha_{\delta_2} - \vect{f}^\alpha_{\delta_1})\\
l_{\delta_x} \zelem (\vect{f}^\alpha_{\delta_1} + \vect{f}^\alpha_{\delta_2})\\
l_{\delta_y} \sinn{\alpha_0}\zelem (\vect{f}^\alpha_{\delta_2} - \vect{f}^\alpha_{\delta_1})
\end{array}\right],
\end{equation}
where $l_{\delta_y}$ is the absolute distance between the vehicle center of gravity and each flap center along the $\vect{b}_y$ axis, and $l_{\delta_x}$ is the distance from this axis to the aerodynamic center of both flaps.
The total aerodynamic moment can now be obtained by summing the contributions, as follows:
\begin{equation}\label{eq:moment_total}
\vect{m} = \vect{m}_T+ \vect{m}_\mu+\vect{m}_\delta.
\end{equation}
\section{Differential Flatness}\label{sec:invdyn}
The purpose of our control design is to accurately track the trajectory reference
\begin{equation}\label{eq:traj}
\vect{\sigma}_{\sref}(t)=[\vect{x}_{\sref}(t)^\top\;\psi_{\sref}(t)]^\top,
\end{equation}
which consists of four elements: the vehicle position in the world-fixed reference frame $\vect{x}_{\sref}(t)\in \realR^3$ and the yaw angle $\psi_{\sref}(t)\in \mathbb{T}$, where $\mathbb{T}$ denotes the circle group. 
The reference $\vect{\sigma}_{\sref}(t)$ may be provided by a pre-planned trajectory or by an online motion planning algorithm.
Henceforward, we do not explicitly write the time argument $t$ everywhere.
By taking the derivative of $\vect{x}_{\sref}$, we obtain continuous references for velocity $\vect{v}_{\sref}$, acceleration $\vect{a}_{\sref}$, and jerk $\vect{j}_{\sref}$.
Similarly, we obtain a continuous yaw rate reference $\dot \psi_{\sref}$ from the yaw reference $\psi_{\sref}$.
For dynamic feasibility\revised{, \ie, for the trajectory to be attainable subject to the vehicle dynamics}, it is required that the position reference $\vect{x}_{\sref}$ is at least fourth-order continuous and that the yaw reference $\psi_{\sref}$ is at least second-order continuous~\cite{tal2021algorithms}.
\revised{As we will show in this section, tracking a discontinuous trajectory (derivative) requires physically infeasible discontinuities in the state or control input.
Practically speaking, our control design can handle (mild) discontinuities without issues because stabilizing feedback control will ensure that the tracking error due to the discontinuity is quickly removed.}

\rrevised{Differential flatness of a nonlinear dynamics system entails the existence of an equivalent controllable linear system.}
An important property of flat systems is that their state and input variables can be directly expressed as a function of the flat output and a finite number of its derivatives.
This property is of major importance when developing trajectory generation and tracking algorithms, as it allows one to readily obtain state and input trajectories corresponding to an output trajectory, effectively transforming the output tracking problem into a state tracking problem.
In practice, the state trajectory can serve as a feedforward control input that enables tracking of higher-order derivatives of the flat output.
Inclusion of these feedforward inputs improves trajectory tracking performance by reducing the phase lag in response to rapid changes in the flat output.
\rrevised{For details on differential flatness and its applications in general, we refer the reader to \cite{fliess1992lessystemesnon,fliess1993linearisation, fliessintro}.}

In this section, we show differential flatness of the dynamics system described in Section \ref{sec:model}---with some simplifications---by deriving expressions of the state and control inputs as a function of the flat output defined by \eqref{eq:traj}.
The expression for angular velocity is used in our trajectory-tracking controller to obtain a feedforward input based on the reference jerk and yaw rate\rrevised{, and the expressions} for attitude and the control inputs are used for linear and angular acceleration control, respectively.

\subsection{Attitude and Collective Thrust}\label{sec:diff_attthrust}
The position and velocity states are trivially obtained from \eqref{eq:traj}.
We arrive at expressions for the attitude and collective thrust by rewriting \eqref{eq:vdot} as
\begin{equation}\label{eq:fi}
\vect{f}^i = m\left(\vect{a} - g\vect{i}_z\right) - \vect{f}_{\ext},
\end{equation}
where we assume that $\vect{f}_{\ext}$ is constant.
In practice $\vect{f}_{\ext}$ may not be constant, but it is implicitly estimated and corrected for by incremental control, as described in Section \ref{sec:linacc_control}.
Given \eqref{eq:fi}, the vehicle attitude and collective thrust are uniquely defined by three major constraints:
\begin{enumerate}
	\item[(i)] the yaw angle reference $\psi$,
	\item[(ii)] the fact that $\yelem \vect{f}^\alpha = 0$ according to \eqref{eq:force_thrust}, and
	\item[(iii)] the forces in the vehicle symmetry plane, \ie, $\xelem \vect{f}^\alpha$ and $\zelem\vect{f}^\alpha$.
\end{enumerate}
Additionally, we exploit the continuity of yaw and the fact that the collective thrust must be non-negative.

\begin{figure*}[t!]
	\centering
	\begin{subfigure}[t]{0.3\textwidth}
		\centering
		\includegraphics[trim={45em 0em 60em 30em},clip,width=\linewidth]{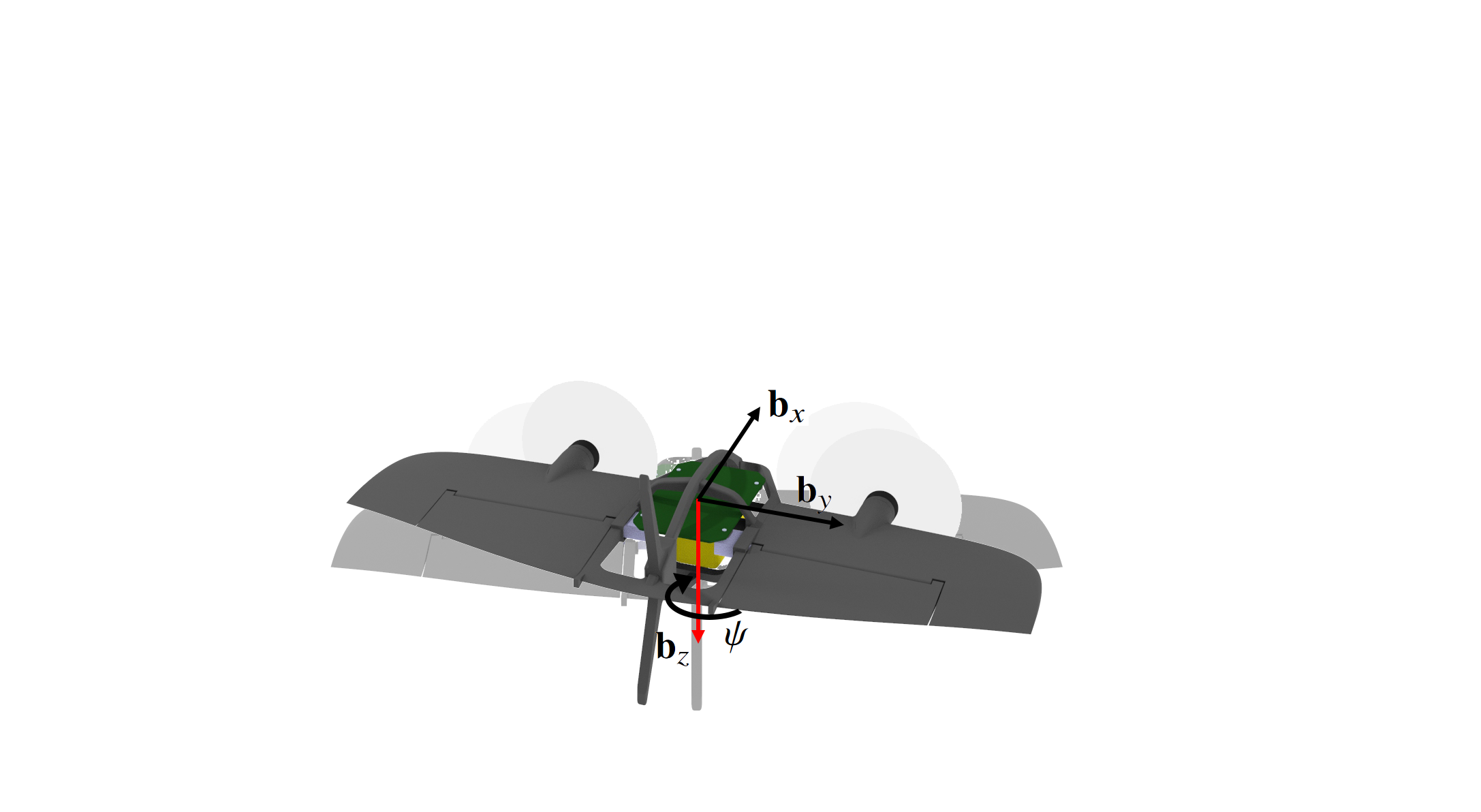}
		\caption{Yaw rotation to satisfy the yaw reference.}
		\label{fig:yaw}
	\end{subfigure}%
	~ 
	\begin{subfigure}[t]{0.3\textwidth}
		\centering
		\includegraphics[trim={45em 0em 60em 30em},clip,width=\linewidth]{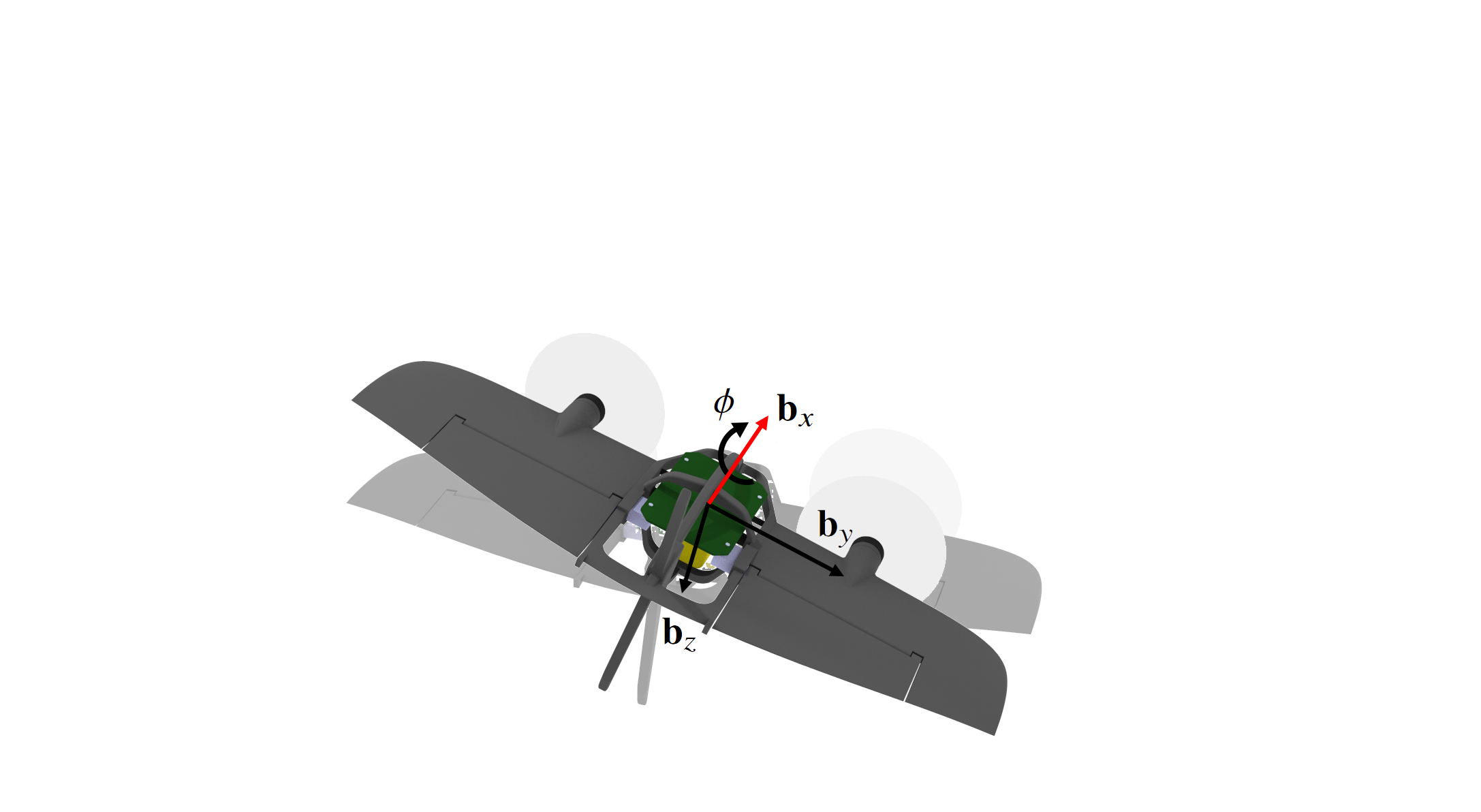}
		\caption{Roll rotation to satisfy $\yelem \vect{f}^\alpha = 0$.}
		\label{fig:roll}
	\end{subfigure}
	~
	\begin{subfigure}[t]{0.3\textwidth}
		\centering
		\includegraphics[trim={45em 0em 60em 30em},clip,width=\linewidth]{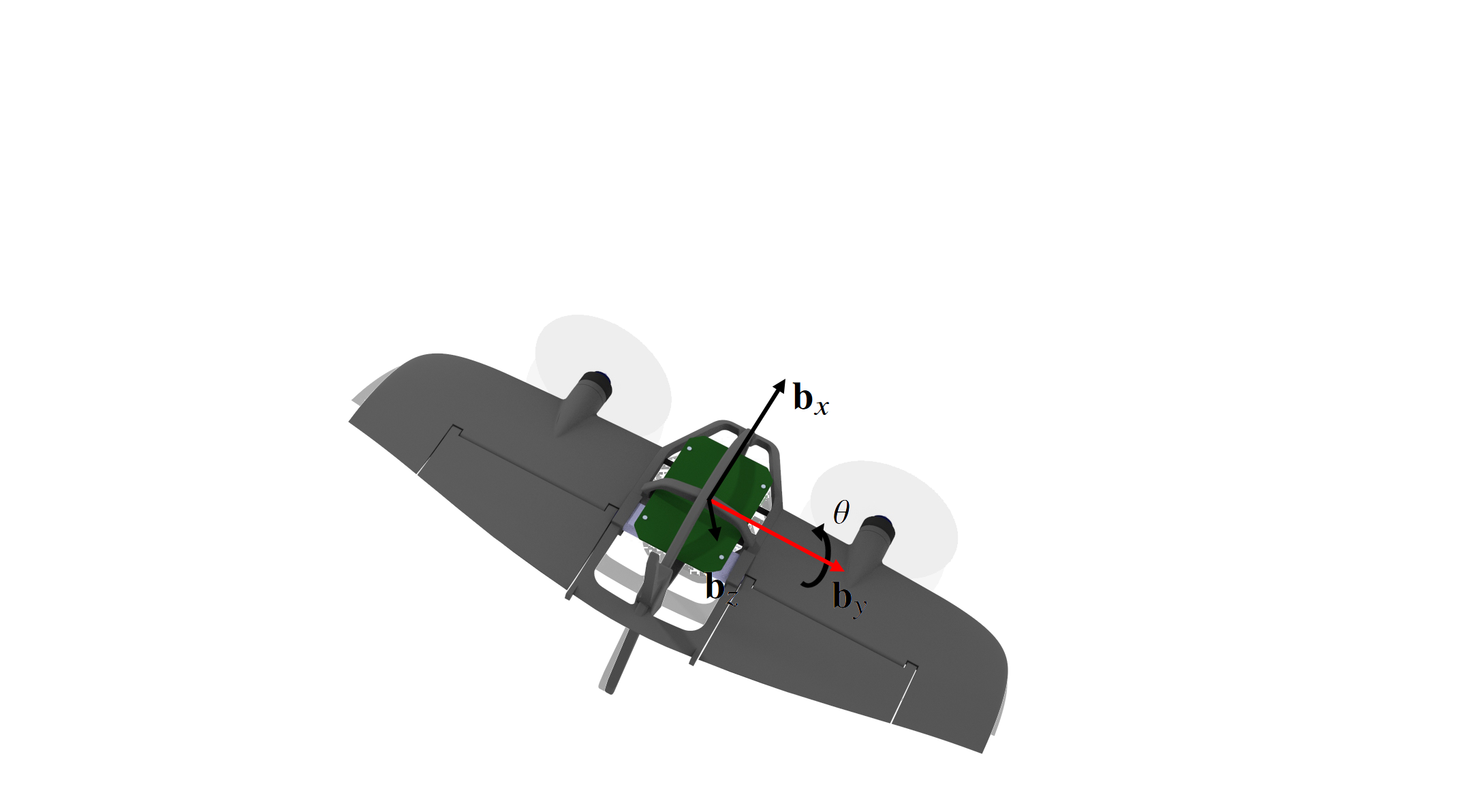}
		\caption{Pitch rotation to attain $\xelem \vect{f}^\alpha$ and $\zelem\vect{f}^\alpha$.}
		\label{fig:pitch}
	\end{subfigure}
	\caption{\revised{ZXY Euler angle rotation sequence} for attitude flatness transform.}
	\label{fig:attcmd_rotations}
\end{figure*}

In this section, \revised{we express the attitude using the ZXY, or 3-1-2, Euler angles $\psi$, $\phi$, and $\theta$}.
The angle symbols are also used to refer to rotation matrices between intermediate frames, \eg, the rotation matrix $\vect{R}^\phi_i$ represents the rotations by $\psi$ and $\phi$.
The ZXY Euler angles form a valid and universal attitude representation that is suitable for the flat transform because each of the angles is uniquely defined by one of the constraints, as shown in Fig. \ref{fig:attcmd_rotations}.
In order to avoid the well-known issues with Euler angles, we convert the obtained attitude to quaternion format before it is used by the flight controller.

We define the yaw angle $\psi$ as the angle between the world-fixed $\vect{i}_y$-axis and the projection of the body-fixed $\vect{b}_y$-axis onto the horizontal plane, \ie, the plane perpendicular to $\vect{i}_z$.
While this angle is undefined if the wingtips are pointing straight up/down (\ie, $\zelem{\vect{b}_y} = \pm1$), we avoid ambiguity by performing the yaw rotation $\psi \vect{i}_z$ from the identity rotation (\ie, from $\vect{R}^b_i = \vect{I}$).

Next, we satisfy constraint (ii) by rotation around the yawed $x$-axis $\vect{R}^i_\psi \vect{i}_x$ by
\begin{equation}\label{eq:roll}
\phi = -\atan2\left(\yelem \vect{R}^\psi_i \vect{f}^i,\zelem \vect{f}^i\right) + k\pi,
\end{equation}
where $\atan2$ is the four-quadrant inverse tangent function. In the second term, $k\in\{0,1\}$ is set such that $\vect{b}_y \bullet (\vect{R}^i_\phi \vect{i}_y) > 0$, \ie, such that the obtained $y$-axis corresponds as closely as possible to the current $\vect{b}_y$-axis.
This results in the equivalence $\psi \equiv \psi + \pi$,
which enables continuous yaw tracking through discontinuities, such as a roll maneuver where the yaw angle instantly switches to $\psi + \pi$.
Unwanted switching does not occur due to continuity of the yaw reference.
If the commanded force is entirely in the horizontal plane and perpendicular to the yaw direction, both arguments of the tangent function are zero, and any $\phi$ satisfies the constraint.
This condition is highly unlikely to occur in actual flight, but can be resolved in practice by setting $\phi$ to match the current direction of $\vect{b}_y$ as closely as possible.

To satisfy constraint (iii), we solve \eqref{eq:force_total} for the collective thrust $T = T_1 + T_2$ and for the rotation angle $\bar \theta$ around the vehicle $y$-axis.
In order to find these expressions, we assume that the flap angles are constant and known.
We can make this assumption without consequence because of a limitation of the INDI acceleration controller.
As described in Section \ref{sec:control}, we only consider the low-frequency component of the flap deflection when controlling the linear acceleration.
This slow-changing component is virtually constant between control updates.

Since the individual thrust values are still undetermined, we assume that the difference between the steady-state flap deflections is negligible so that
\begin{equation}
\delta_1T_1+\delta_2T_2 \approx \frac{T}{2}  \delta,
\end{equation}
where $ \delta = \delta_1 + \delta_2$.
This assumption may be violated during sustained maneuvers or flight with sideslip, but we have found that it typically does not lead to large discrepancies.
We substitute into \eqref{eq:force_total} $\vect{f}^\alpha = \vect{R}^{\bar\theta}_\phi \vect{f}^\phi$ and $\vect{v}^\alpha = \vect{R}^{\bar\theta}_\phi \vect{v}^\phi$ with $\vect{f}^\phi = \vect{R}^\phi_i \vect{f}^i$, $\vect{v}^\phi = \vect{R}^\phi_i \vect{v}^i$.
Note that $\bar \theta$ refers to the rotation from $\phi$ to the zero-lift reference frame, while $\theta$ is the rotation to the body-fixed reference frame.
We obtain the following two equalities:
\begin{multline}
\ccos{\bar\alpha} \left(1 - c_{D_T}\right) T - c_{D_V} \|\vect{v}\|\left(\ccos {\bar \theta} \xelem{\vect{v}^\phi} - \ssin {\bar \theta} \zelem \vect{v}^\phi \right)\\ = \ccos{{\bar \theta}}\xelem\vect{f}^\phi -\ssin{{\bar \theta}}\zelem{\vect{f}^\phi},\label{eq:thetaTunsolved1}
\end{multline}
\begin{multline}
\left(\ssin{\bar\alpha} (c_{L_T} - 1) - \ccos{\bar\alpha}c^\delta_{L_T}\nicefrac{\delta}{2}\right)T \\
- c^\delta_{L_V}\delta \|\vect{v}\|\left(\ccos{{\bar \theta}} \xelem\vect{v}^\phi - \ssin{{\bar \theta}}\zelem \vect{v}^\phi\right) \\- c_{L_V} \|\vect{v}\| \left(\ssin{{\bar \theta}}\xelem\vect{v}^\phi + \ccos{{\bar \theta}}\zelem{\vect{v}^\phi}\right) = \ssin{{\bar \theta}}\xelem\vect{f}^\phi + \ccos{{\bar \theta}}\zelem\vect{f}^\phi,\label{eq:thetaTunsolved2} 
\end{multline}
where $\operatorname{c}$ and $\operatorname{s}$ represent respectively cosine and sine, and $\bar \alpha = \alpha_0 + \alpha_T$.
Solving \eqref{eq:thetaTunsolved1} and \eqref{eq:thetaTunsolved2} for $\bar \theta$ and $T$ gives
\begin{multline}
{\bar \theta} = \atan2\left(\eta \left(\xelem\vect{f}^\phi+c_{D_V}\|\vect{v}\|\xelem \vect{v}^\phi \right)\right.\\
- c^\delta_{L_V}\delta\|\vect{v}\|\xelem \vect{v}^\phi - c_{L_V}\|\vect{v}\|\zelem\vect{v}^\phi-\zelem\vect{f}^\phi,\\
\hfill \eta \left(\zelem\vect{f}^\phi+c_{D_V}\|\vect{v}\|\zelem \vect{v}^\phi \right) - c^\delta_{L_V}\delta\|\vect{v}\|\zelem\vect{v}^\phi \;\;\\
\hfill\left.+ c_{L_V}\|\vect{v}\|\xelem\vect{v}^\phi+\xelem\vect{f}^\phi \right) + k\pi,\label{eq:theta}\\
\end{multline}
\begin{multline}
T = {\ccos{\bar\alpha}\left(1-c_{D_T}\right)}^{-1}\\
\left(\ccos{{\bar \theta}}\xelem\vect{f}^\phi - \ssin{{\bar \theta}}\zelem\vect{f}^\phi + c_{D_V} \|\vect{v}\|\left(\ccos {\bar \theta} \xelem{\vect{v}^\phi} - \ssin {\bar \theta} \zelem \vect{v}^\phi \right)\right),\label{eq:T}
\end{multline}
where
\begin{equation}
\eta = \frac{\ssin{\bar \alpha} \left(c_{L_T} - 1\right) - \ccos{\bar \alpha} c^\delta_{L_T}\nicefrac{\delta}{2}}{\ccos{\bar\alpha}\left(1-c_{D_T}\right)}.
\end{equation}
is the ratio of lift and forward force due to thrust.
We set $k\in \{0,1\}$ such that $T\geq 0$. In the unlikely event that constraint (iii) is satisfied for any $\bar \theta$, both arguments of the $\atan2$ function equal zero and, in practice, we can set $\bar \theta$ to match the current attitude as closely as possible.
Since the angle $\bar \theta$ is the rotation to the zero-lift axis system, we use $\theta = \bar\theta + \alpha_0$ to obtain the corresponding rotation to the body-fixed reference frame.

Note that we purposely selected the ZXY rotation sequence and the definition of yaw, such that $\phi$ and $\theta$ do not affect the satisfaction of constraint (i), and $\theta$ does not affect the satisfaction of constraint (ii).
Given that the Euler angles are uniquely defined (up to addition of $\pi$) by the yaw reference, \eqref{eq:roll}, and \eqref{eq:theta}, this implies that these expressions give the attitude as a function of $\vect{\sigma}_{\sref}$.

\subsection{Angular Velocity}\label{sec:diffangvel}
By taking the derivative of \eqref{eq:roll}, we obtain
\begin{equation}\label{eq:phidot}
\dot \phi = -\frac{\dot\beta_x\beta_z-\beta_x\dot\beta_z}{\beta_x^2 + \beta_z^2},
\end{equation}
where $\beta_x$ and $\beta_z$ are respectively the first and second arguments of the $\atan2$ function in \eqref{eq:roll}, and
\begin{align}
	\dot\beta_x &= -\ccos{\psi} \dot \psi \xelem\vect{f}^i - \ssin{\psi} \xelem\dot{\vect{f}}^i - \ssin{\psi}\dot\psi\yelem\vect{f}^i + \ccos{\psi}\yelem\dot{\vect{f}}^i,\label{eq:betadot}\\
	\dot\beta_z &= \zelem\dot{\vect{f}}^i.\label{eq:beta2dot}
\end{align}
The force derivative is obtained as the derivative of \eqref{eq:fi}, as follows:
\begin{equation}\label{eq:forceder}
\dot{\vect{f}}^i = m\vect{j}.
\end{equation}
We take the derivative of \eqref{eq:theta} to obtain
\begin{equation}\label{eq:dottheta}
\dot \theta = \frac{\dot\sigma_x\sigma_z - \sigma_x\dot\sigma_z }{\sigma_x^2 + \sigma_z^2},
\end{equation}
where $\sigma_x$ and $\sigma_z$ are respectively the first and second arguments of the $\atan2$ function in \eqref{eq:theta}, and
\begin{align}
	\dot\sigma_x &= \eta \left(\xelem\dot{\vect{f}}^\phi + c_{D_V} \tau_x \right) - c^\delta_{L_V}  \delta \tau_x - c_{L_V}\tau_z - \zelem\dot{\vect{f}}^\phi,\label{eq:sigma1dot}\\
	\dot\sigma_z &= \eta \left(\zelem\dot{\vect{f}}^\phi + c_{D_V} \tau_z \right) - c^\delta_{L_V}  \delta \tau_z + c_{L_V}\tau_x + \xelem\dot{\vect{f}}^\phi\label{eq:sigma2dot}
\end{align}
with
\begin{align}
	\tau_x &= \dot{\|\vect{v}\|} \xelem \vect{v}^\phi + \|\vect{v}\|\xelem\dot{\vect{v}}^\phi,\\
	\tau_z &= \dot{\|\vect{v}\|} \zelem \vect{v}^\phi + \|\vect{v}\|\zelem\dot{\vect{v}}^\phi
\end{align}
and
\begin{align}
	\dot{\|\vect{v}\|} &= \frac{\vect{v}^\top\vect{a}}{\|\vect{v}\|},\\
	\dot{\vect{v}}^\phi &= \dot{\vect{R}}^\phi_i\vect{v} + \vect{R}^\phi_i\vect{a}.\label{eq:accel}
\end{align}
The expression for the force derivative $\dot{\vect{f}}^\phi$ is similar to \eqref{eq:accel}.
In the force equations, we assume that the flap deflection is known and that its temporal derivatives are negligible, as described in Section \ref{sec:diff_attthrust}.
Finally, we obtain the angular velocity in the body-fixed reference frame, as follows:
\begin{equation}\label{eq:angrate}
\vect{\Omega} = \left[\begin{array}{c}
0\\\dot \theta\\0
\end{array}\right] + \vect{R}^{ \theta}_\phi \left[\begin{array}{c}
\dot \phi \\0\\0
\end{array}\right] + \vect{R}^{ \theta}_\psi \left[\begin{array}{c}
0\\0\\\dot \psi
\end{array}\right].
\end{equation}

\subsection{Control Inputs}\label{sec:diffcontrol}
At this point, we have expressed the state variables as a function of the flat output \eqref{eq:traj}.
To obtain an expression for the control inputs, an expression for angular acceleration is obtained as the derivative of \eqref{eq:angrate} and substituted into \eqref{eq:Omegadot} to obtain an expression for $\vect{m}$.
The angular acceleration can be utilized as a feedforward input corresponding to snap, the fourth derivative of position, and yaw acceleration~\cite{tal2020accurate}.
However, calculation of this feedforward input significantly complicates the controller expressions, and its benefit may be marginal given how challenging it is to perform accurate feedforward control of the angular acceleration of a fixed-wing aircraft.
Hence, we do not incorporate the angular acceleration feedforward input in our control design.
For the sake of completeness, we still include the corresponding expressions in Appendix \ref{app:snap}.

As described in Section \ref{sec:angacc_control}, our control design obtains a moment command using INDI.
To find the corresponding control inputs, \ie, flap deflections and differential thrust $\Delta T = T_1 - T_2$, we solve \eqref{eq:moment_total} for these inputs.
We find an expression for the differential thrust $\Delta T$ by equating 
\begin{equation}
\zelem\left( \vect{m}_T + \vect{m}_\mu\right) = \zelem \vect{m},
\end{equation}
which assumes that the contribution by $\zelem \vect{m}_\delta$ is negligible.
Due to the multiplication with $\sin{\alpha_0}$, this assumption typically does not result in significant discrepancies.
Using $\mu_1 + \mu_2 = \nicefrac{c_\mu}{c_T}\Delta T$, we obtain
\begin{equation}\label{eq:DeltaT}
\Delta T = \frac{\zelem \vect{m}}{l_{T_y}\left(\ccos{\alpha_0}\ccos{\bar \alpha}(1 - c_{D_T}) - \ssin{\alpha_0}\ssin{\bar\alpha}(c_{L_T}- 1)\right) - \ssin{\alpha_T}\frac{c_\mu}{c_T}}.
\end{equation}
The individual thrust values are then given by
\begin{equation}\label{eq:T1T2}
	T_1 = \frac{T + \Delta T}{2},\;\;\;\;\;\;\;\;\;\;\;\;
	T_2 = \frac{T - \Delta T}{2}.
\end{equation}
After obtaining the motor speeds from \eqref{eq:thrust}, we can deduct $\vect{m}_T$ and $\vect{m}_\mu$ from $\vect{m}$ to obtain $\vect{m}_\delta$.
Finally, the flap deflections are computed by inversion of \eqref{eq:moment_flap}, as follows:
\begin{equation}\label{eq:d1d2}
\left[\begin{array}{c}
\delta_1\\
\delta_2
\end{array}\right] = \left[\begin{array}{cc}
-l_{\delta_y} \ccos{\alpha_0} {\nu}_1& l_{\delta_y} \ccos{\alpha_0} {\nu}_2\\
l_{\delta_x}  {\nu}_1&l_{\delta_x}  {\nu}_2
\end{array}\right]^{-1}\left[\begin{array}{c}
\xelem \vect{m}_\delta\\
\yelem \vect{m}_\delta
\end{array}\right]
\end{equation}
with
\begin{equation}
{\nu}_i = -c^\delta_{L_T} \coss{\bar\alpha} T_i - c^\delta_{L_V}\|\vect{v}\| \xelem\vect{v}^\alpha.
\end{equation}
\section{Trajectory-tracking Control}\label{sec:control}
Our proposed controller is designed to accurately track the dynamic position reference $\vect{\sigma}_{\sref}$.
It consists of several components based on various control methodologies, as shown in Fig. \ref{fig:diagram}.
Each component employs a global formulation that enables seamless maneuvering throughout the flight envelope.
By separating kinematics and dynamics, we are able to employ proportional-derivative (PD) control on the translational and rotational kinematics.
Application of the resulting linear and angular acceleration commands is performed using INDI control.
INDI enables accurate control by incremental adjustment of control inputs, based on the inverted dynamics model derived in Section \ref{sec:invdyn}.
Due to its incremental formulation, the controller only depends on local accuracy of the input-output relation, resulting in favorable robustness against modeling errors and external disturbances.
As we will detail in this section, these errors and disturbances (\ie, $\vect{f}_{\ext}$ and $\vect{m}_{\ext}$) are implicitly estimated and corrected for based on the difference between sensor-based and model-based force and moment estimates.
By directly incorporating linear and angular acceleration measurements to obtain the sensor-based estimates, the controller is able to quickly and wholly counteract errors and disturbances, without relying on integral action.

Our proposed control design uses a state estimate consisting of position $\vect{x}$, velocity $\vect{v}$, and attitude $\vect{\xi}$.
Additionally, linear acceleration $\vect{a}^b$ and angular velocity $\vect{\Omega}$ measurements in the body-fixed reference frame are obtained from the inertial measurement unit (IMU).
Motor speeds $\vect{\omega}$ and flap deflections $\vect{\delta}$ are measured and utilized as well.

\tikzset{
	block/.style = {draw, fill=white, rectangle, minimum height=3em, minimum width=3em},
	tmp/.style  = {coordinate},
}

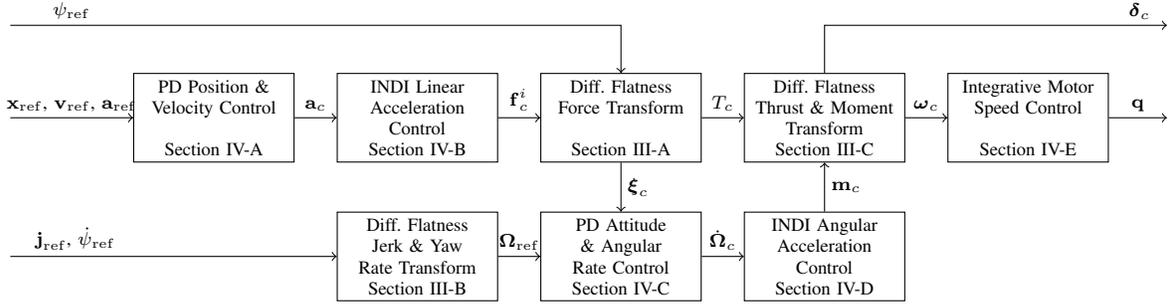
\begin{figure*}
\scriptsize
\centering
{\begin{tikzpicture}[auto, node distance=6em]
	
	\node[tmp](in1){};
	\node[tmp,below of=in1, node distance=7.5em](in3){};
	\node[tmp, above of=in1, node distance=5em](in2){};
	\node [block,right of=in1, node distance = 11 em,text width=8em,align=center](PDPos){PD Position \& Velocity Control\\~\\Section \ref{sec:posvel_control}};
	
	\node [block,right of=PDPos, node distance = 11 em,text width=8em,align=center](INDIAcc){INDI Linear Acceleration Control\\Section \ref{sec:linacc_control}};
	
	\node [block,below of=INDIAcc, node distance = 7.5 em,text width=8em,align=center](Jerk){Diff. Flatness Jerk \& Yaw Rate Transform\\Section \ref{sec:diffangvel}};
	
	\node [block,right of=INDIAcc, node distance =11 em,text width=8em,align=center](Force){Diff. Flatness Force Transform\\~\\Section \ref{sec:diff_attthrust}};
	
	\node [block,below of=Force, node distance = 7.5 em,text width=8em,align=center](PDAtt){PD Attitude \& Angular Rate Control\\Section \ref{sec:att_control}};
	
	\node [block,right of=PDAtt, node distance = 11 em,text width=8em,align=center](INDIAng){INDI Angular Acceleration Control\\Section \ref{sec:angacc_control}};
		
	\node [block,right of=Force, node distance = 11 em,text width=8em,align=center](Moment){Diff. Flatness Thrust \& Moment Transform\\Section \ref{sec:diffcontrol}};
	
	\node [block,right of=Moment, node distance = 11 em,text width=8em,align=center](Motor){Integrative Motor Speed Control\\~\\Section \ref{sec:motorspeed_control}};
	
	\node[tmp,right of=Motor, node distance = 7.5em](out1){};
	\node[tmp,above of=out1, node distance = 5em](out2){};
	
	\draw [->] (in2) -| (Force)
	node [above,pos=0.05]{${\psi}_{\sref}$};
	
	\draw [->] (in3) -- (Jerk)
	node [above,pos=0.2]{$\vect{j}_{\sref}$, $\dot\psi_{\sref}$};
	
	\draw [->] (Jerk) -- (PDAtt)
	node [above,pos=0.5]{$\vect{\Omega}_{\sref}$};
	
	\draw [->] (in1) -- (PDPos)
	node [above,pos=0.5]{$\vect{x}_{\sref}$, $\vect{v}_{\sref}$, $\vect{a}_{\sref}$};
	
	\draw [->] (PDPos) -- (INDIAcc)
	node [above,pos=0.5]{$\vect{a}_{c}$};
	
	\draw [->] (INDIAcc) -- (Force)
	node [above,pos=0.5]{$\vect{f}_{c}^i$};
	
	\draw [->] (Force) -- (PDAtt)
	node [right,pos=0.5]{$\vect{\xi}_{c}$};
	
	\draw [->] (Force) -- (Moment)
	node [above,pos=0.5]{$T_{c}$};
	
	\draw [->] (PDAtt) -- (INDIAng)
	node [above,pos=0.5]{$\dot{\vect{\Omega}}_c$};
	
	\draw [->] (INDIAng) -- (Moment)
	node [right,pos=0.5]{$\vect{m}_{c}$};
	
	\draw [->] (Moment) -- (Motor)
	node [above,pos=0.5]{$\vect{\omega}_c$};
	
	\draw [->] (Motor) -- (out1)
	node [above,pos=0.5]{$\vect{q}$};
	
	\draw [->] (Moment) |- (out2)
	node [above,pos=0.96]{$\vect{\delta}_c$};
	
\end{tikzpicture}}
\caption{Overview of trajectory-tracking control architecture.} \label{fig:diagram}
\end{figure*}

\subsection{PD Position and Velocity Control}\label{sec:posvel_control}
We use cascaded proportional-derivative (PD) controllers for position and velocity control, resulting in the following expression for the acceleration command:
\begin{multline}\label{eq:tspd}
\vect{a}_c =\vect{R}_b^i \left(\vect{K}_{\vect{x}} \vect{R}_i^b \left(\vect{x}_{\sref}-\vect{x}\right) + \vect{K}_{\vect{v}}\vect{R}_i^b \left(\vect{v}_{\sref}-\vect{v}\right)\right.\\
\left.+ \vect{K}_{\vect{a}}\vect{R}_i^b \left(\vect{a}_{\sref}-\tilde{\vect{a}}_{\lpf}\right)\right) + \vect{a}_{\sref}
\end{multline}
with $\vect{K_{\bullet}}$ indicating diagonal gain matrices.
The first term of \eqref{eq:tspd} aims to null the position and velocity errors, while the second term is a feedforward input that ensures the acceleration reference is accurately tracked.
Since the vehicle has different acceleration capabilities along its body-fixed axes, we define the control gains in the body-fixed reference frame and transform them to the world-fixed frame for each control update.

The gravity-corrected linear acceleration in the world-fixed reference frame is obtained as
\begin{equation}
\vect{a}_{\lpf} = (\vect{R}_b^i\vect{a}^b +g\vect{i}_z )_{\lpf},
\end{equation}
where ${\lpf}$ indicates low-pass filtering that is applied to IMU measurements to alleviate measurement noise, e.g., due to vibrations.
We follow the method by \cite{smeur2020incremental} and deduct acceleration contributions due to the transient flap movements.
This correction helps eliminate pitch oscillations that may result from the non-minimum phase response of acceleration to flap deflections.
To isolate transient movement, we first filter the measured flap deflections using the low-pass filter and then using a high-pass filter, resulting in a band-pass filtered signal.
The low-pass filter helps to match the phase delay between accelerometer and flap deflection measurements, and ensures that we do not (re-)introduce high-frequency noise in the resulting acceleration signal
\begin{equation}\label{eq:acc_correction}
\tilde{\vect{a}}_{\lpf} = {\vect{a}}_{\lpf} - m^{-1}\vect{R}^i_\alpha \vect{f}^\alpha_{\delta_{\hpf}}.
\end{equation}

\subsection{INDI Linear Acceleration Control}\label{sec:linacc_control}
INDI control incrementally updates the attitude and collective thrust to track the acceleration command $\vect{a}_c$.
The controller estimates the unmodeled force $\vect{f}_{\ext}$ by comparing the measured acceleration to the expected acceleration according to the vehicle aerodynamics model and motor speed measurements.
By rewriting \eqref{eq:vdot}, we obtain
\begin{equation}\label{eq:tsfext}
\vect{f}_{\ext} = m\left(\tilde{\vect{a}}_{\lpf} - g \vect{i}_z\right) - \vect{R}^i_\alpha \vect{f}^\alpha_{\lpf},
\end{equation}
where, for consistency with $\tilde{\vect{a}}_{\lpf}$, low-pass filtered motor speeds and the filtered flap deflections without transient component are used in the computation of $\vect{f}^\alpha_{\lpf}$ according to \eqref{eq:force_total}.
Substitution of \eqref{eq:tsfext} into \eqref{eq:vdot} gives
\begin{align}
\begin{split}
\label{eq:indi_force}
{\vect{a}} &= g\vect{i}_z + m^{-1}\left(\vect{R}^i_\alpha\vect{f}^\alpha + \vect{f}_{\ext}\right)\\
 &= g\vect{i}_z + m^{-1}\left(\vect{R}^i_\alpha\vect{f}^\alpha + \left(m\left(\tilde{\vect{a}}_{\lpf} - g \vect{i}_z\right) - \vect{R}^i_\alpha \vect{f}^\alpha_{\lpf}\right)\right)\\
 &=\tilde{\vect{a}}_{\lpf} +  m^{-1}\left(\vect{f}^i - \vect{f}^i_{\lpf}\right).
 \end{split}
\end{align}
Solving \eqref{eq:indi_force} for $\vect{f}^i$ gives an incremental expression for the force command that corresponds to the commanded acceleration, as follows:
\begin{equation}\label{eq:force_command}
\vect{f}^i_c = m(\vect{a}_c - \tilde{\vect{a}}_{\lpf}) + \vect{f}^i_{\lpf}.
\end{equation}
This incremental control law enables the controller to achieve the commanded acceleration despite potential modeling discrepancies and external forces, without depending on integral action.
If the commanded acceleration is not yet attained, the force command will be adjusted further in subsequent control updates until the first term in \eqref{eq:force_command} vanishes.
Based on the force command $\vect{f}^i_c$, the commanded attitude $\vect{\xi}_c$ is obtained from $\psi_{\sref}$, \eqref{eq:roll} and \eqref{eq:theta}, and the collective thrust command $T_c$ is obtained from \eqref{eq:T}.
Note that it is through the flatness transform described in Section \ref{sec:invdyn} that our INDI algorithm can perform fully nonlinear inversion, without linearization of the dynamics \eqref{eq:indi_force}.
The nonlinear inversion provides more accurate control commands when large acceleration deviations occur, such as may happen during aggressive maneuvers with quickly changing acceleration references.

\subsection{PD Attitude and Angular Rate Control}\label{sec:att_control}
Given the extensive attitude envelope of the tailsitter vehicle, our attitude controller employs quaternion representation to avoid kinematic singularities.
The attitude error quaternion is obtained as
\begin{equation}
\vect{\xi}_e = \vect{\xi}^{-1}\circ\vect{\xi}_c,
\end{equation}
and the corresponding three-element error angle vector is given by
\begin{equation}
\vect{\zeta}_e = \frac{2 \arccos\xi_e^w}{\sqrt{1- {\left(\xi_e^w\right)}^2}}\left[\begin{array}{ccc}
\xi_e^x&\xi_e^y&\xi_e^z
\end{array}\right]^\top,
\end{equation}
where the superscript refers to individual components of $\vect{\xi}_e$.
The angular acceleration command is obtained using the PD controller
\begin{equation}\label{eq:tsPD}
\dot{\vect{\Omega}}_c = \vect{K}_{\vect{\xi}} \vect{\zeta}_e + \vect{K}_{\vect{\Omega}} \left(\vect{\Omega}_{\sref}-\vect{\Omega}_{\lpf}\right),
\end{equation}
where $\vect{\Omega}_{\lpf}$ is the low-pass filtered angular velocity measurement from the IMU, and $\vect{\Omega}_{\sref}$ is the feedforward angular velocity reference obtained by \eqref{eq:angrate} based on $\dot{\psi}_{\sref}$ and $\vect{j}_{\sref}$.
By including this feedforward jerk term, the controller improves trajectory-tracking accuracy, especially on agile trajectories with fast-changing acceleration references.

\subsection{INDI Angular Acceleration Control}\label{sec:angacc_control}
The angular acceleration controller has a similar construction as its linear acceleration counterpart described in Section \ref{sec:linacc_control}.
By rewriting \eqref{eq:Omegadot}, we obtain the following expression for the unmodeled moment:
\begin{equation}\label{eq:moment_ext}
\vect{m}_{\ext} = \vect{J}\dot{\vect{\Omega}}_{\lpf} - \vect{m}_{\lpf} + \vect{\Omega}_{\lpf} \times \vect{J}\vect{\Omega}_{\lpf},
\end{equation}
where $\dot{\vect{\Omega}}_{\lpf}$ is obtained by numerical differentiation of ${\vect{\Omega}}_{\lpf}$, and $\vect{m}_{\lpf}$ is calculated using \eqref{eq:moment_total} and based on the low-pass filtered flap deflection and motor speed measurements.
Substitution of \eqref{eq:moment_ext} into \eqref{eq:Omegadot} gives
\begin{align}\label{eq:indi_moment}
\begin{split}
\dot{\vect{\Omega}} &= \vect{J}^{-1}\left(\vect{m} + \vect{m}_{\ext}-\vect{\Omega} \times \vect{J}\vect{\Omega}\right)\\
&= \vect{J}^{-1}\left(\vect{m} + \left(\vect{J}\dot{\vect{\Omega}}_{\lpf} - \vect{m}_{\lpf} + \vect{\Omega}_{\lpf} \times \vect{J}\vect{\Omega}_{\lpf}\right)-\vect{\Omega} \times \vect{J}\vect{\Omega}\right)\\
&=\dot{\vect{\Omega}}_{\lpf} + \vect{J}^{-1}\left(\vect{m} - \vect{m}_{\lpf} \right),
\end{split}
\end{align}
where it is assumed that the angular momentum term is relatively slow changing, so that the difference with its filtered version may be neglected.
Solving \eqref{eq:indi_moment} for $\vect{m}$ gives the incremental control law
\begin{equation}\label{moment_command}
\vect{m}_c = \vect{J}(\dot{\vect{\Omega}}_c - \dot{\vect{\Omega}}_{\lpf}) + \vect{m}_{\lpf}.
\end{equation}
Based on the commanded moment $\vect{m}_c$, the thrust and flap deflection commands can now be calculated by \eqref{eq:DeltaT} and \eqref{eq:T1T2}, and \eqref{eq:d1d2}, respectively.
Finally, the commanded motor speeds $\vect{\omega}_c$ are calculated by inversion of \eqref{eq:thrust}.

\subsection{Integrative Motor Speed Control}\label{sec:motorspeed_control}
While the flaps are controlled by servos equipped with closed-loop position control, the propellers are driven by brushless motors that cannot directly track motor speed commands.
Instead, we use the second-order polynomial ${p}$ to find the corresponding throttle input.
This function was obtained from regression analysis of static test data relating motor speed to throttle input.
We add integral action to account for changes due to the fluctuating battery voltage,
so that the throttle command that is sent to the motor electronic speed controller (ESC) is obtained as
\begin{equation}
q_i = {p}(\omega_{i,c}) + {k_{I_\omega}}\int \left(\omega_{i,c}-{\omega_i}\right) \;\;d t,
\end{equation}
where ${k_{I_\omega}}$ is the integrator gain.
\section{Estimation of Aerodynamic Parameters}\label{sec:paramest}
The controller utilizes several mass, geometric, and aerodynamic properties of the vehicle.
The vehicle mass $m$, moment of inertia $\vect{J}$, motor position $l_{T_y}$, motor incidence angle $\alpha_{T}$, and flap position $l_{\delta_y}$ can be measured using standard methods or determined based on the design.
The propeller thrust coefficient $c_T$, torque coefficient $c_\mu$, and throttle response curve $p$ are obtained using static bench tests.
For a simple wing without twist, the zero-lift angle of attack $\alpha_0$ is determined by the airfoil and can thus be obtained from literature or two-dimensional analysis.

\begin{table}[!ht]
	\centering
	\caption{Tailsitter aircraft properties.}
	\label{tab:wing_shape}
	\begin{tabular}{llll}
		\hline
		Property & Symbol & Value&\\
		\hline
		Airfoil lift slope & $\bar c_{l_\alpha}$ & 5.73 &rad\textsuperscript{-1}\\
		Wing surface area & $S$ & 0.070 &m\textsuperscript{2}\\
		Aspect ratio & $AR$ & 4.3&\\
		Taper ratio & $\lambda$ & 0.59&\\
		Flap chord ratio & $\nicefrac{c_f}{c}$ & 0.5 & \\
		Propeller diameter & $D$ & 0.13 &m\\
		Thrust angle & $\alpha_{T}$ & -5$\frac{\pi}{180}$ &rad\\
		Monoplane circulation coefficient & $\tau$ & 0.14 & \\
		Span efficiency factor & $e$ & 0.97 &\\
		\hline
	\end{tabular}
\end{table}

What remains are the $\varphi$-theory aerodynamic coefficients $c_{L_V}$, $c_{D_V}$, $c_{L_T}$, $c_{D_T}$, $c^\delta_{L_V}$, $c^\delta_{L_T}$, and $c_{\mu_T}$, as well as the position of the aerodynamic center of the flaps $l_{\delta_x}$.
Instead of relying on extensive analysis (\eg, through computational fluid dynamics (CFD) or wind tunnel tests), we initially estimate these parameters using back-of-the-envelope calculations and then refine them using flight test data.

\subsection{Analytical Model}

We approximate the $\varphi$-theory coefficients based on conventional Buckingham $\pi$ dimensionless aerodynamic coefficients obtained using Prandtl's classical lifting-line theory.
In order to avoid confusion, we use $\bar c$ to denote the conventional Buckingham $\pi$ coefficients, whereas the $\varphi$-theory and propulsion coefficients used by the model described in Section \ref{sec:model} are written without an overbar.
The lift slope of a finite wing without twist or sweep is given by
\begin{equation}\label{eq:cl}
\bar c_{L_\alpha} = \frac{\bar c_{l_\alpha}}{1 + \frac{\bar c_{l_\alpha}}{\pi AR}(1+\tau)},
\end{equation}
where $\tau$ is a function of the Fourier coefficients used to represent the circulation distribution~\cite{glauert}.
In symmetric flight, small angles of attack can be approximated as
\begin{equation}\label{eq:aoa}
\alpha - \alpha_0 = \frac{\zelem\vect{v}^\alpha}{\|\vect{v}\|}.
\end{equation}
By substituting \eqref{eq:aoa} into \eqref{eq:cl} and equating to the expression in \eqref{eq:force_body}, we obtain
\begin{multline}\label{eq:lift}
\frac{1}{2}\rho \|\vect{v}\|^2 S \bar c_{L_\alpha} (\alpha - \alpha_0) = \frac{1}{2}\rho  S\frac{\bar c_{l_\alpha}}{1 + \frac{\bar c_{l_\alpha}}{\pi AR}(1+\tau)}\zelem\vect{v}^\alpha \|\vect{v}\|\\
= c_{L_V}\zelem\vect{v}^\alpha\|\vect{v}\|\Rightarrow c_{L_V} = \frac{1}{2}\rho  S\frac{\bar c_{l_\alpha}}{1 + \frac{\bar c_{l_\alpha}}{\pi AR}(1+\tau)},
\end{multline}
where $\rho$ is the air density.
We note that the coefficient $c_{L_V}$ is not dimensionless, as described in Section \ref{sec:aeromodel}.
The drag coefficient $c_{D_V}$ represents the force along the zero-lift axis.
Since we assume inviscid flow, this force is fully lift-induced and due to the tilting of the lift vector with regard to the zero-lift line by
\begin{equation}\label{eq:aoa_eff}
\alpha_{\operatorname{eff}} = \alpha - \alpha_0 - \alpha_i,
\end{equation}
where $\alpha_{\operatorname{eff}}$ is the effective angle of attack and $\alpha_i$ is the induced angle of attack due to downwash.
The lift vector is perpendicular to the local velocity, so that its dimensionless component along the zero-lift axis (in positive $\alpha_x$ direction) is given by
\begin{align}\label{eq:ca}
\bar c_A &= \bar c_L \alpha_{\operatorname{eff}} = \bar c_L (\alpha - \alpha_0 - \alpha_i) = \bar c_L (\alpha - \alpha_0) - \bar c_{D_i}\\
&= \left(\bar c_{L_\alpha} - \frac{\bar{c}^2_{L_\alpha}}{\pi AR e}\right)(\alpha - \alpha_0)^2,
\end{align}
where $\bar c_{D_i}$ is the induced drag coefficient and we again use small angle assumptions.
The span efficiency factor $e\leq 1$ is a function of the taper ratio and aspect ratio and represents the deviation from a perfectly elliptical lift distribution~\cite{barnes}.
For anything but very small aspect ratios, the factor on the righthand side of \eqref{eq:ca} is positive, resulting in $\bar c_A > 0$.
This means that we can expect a forward force along the zero-lift axis, especially at low speeds where drag is dominated by the lift-induced contribution.
The format of \eqref{eq:ca} does not correspond to the $c_{D_V}$ component of \eqref{eq:force_body}, since the latter is meant to model viscous drag contributions.
At the relatively low speeds that can be achieved in our indoor flight area, viscous drag is likely small, so we leave $c_{D_V} = 0$ kg/m for now.

From momentum disc theory, we obtain
\begin{equation}\label{eq:Ti}
T_i = \frac{\pi D^2}{4}\frac{\rho}{2}v_{e,i}^2,
\end{equation}
where $D$ is the propeller diameter, $v_{e,i}$ is the downstream velocity of propeller $i$, and we assume negligible upstream velocity.
For simplicity, we assume that the wake of each propeller results in a uniform velocity over a third of the corresponding semi-wing at a geometric angle of attack of $-\alpha_T$.
In reality, the flow is not uniform and significant interaction between the wing and propeller may occur~\cite{leng2020experimental}.
In order to keep the control algorithm tractable, we neglect these effects in our dynamics model.
Substituting \eqref{eq:Ti} into the lefthand side of \eqref{eq:lift} and equating to the corresponding component of \eqref{eq:force_thrust} gives
\begin{equation}\label{eq:clt}
-\sum_{i}T_i \frac{2}{3}\frac{S}{\pi D^2} \bar c_{L_\alpha} \bar{\alpha} = -\sum_{i}T_i c_{L_T} \bar{\alpha} \Rightarrow c_{L_T} = \frac{2}{3}\frac{S}{\pi D^2} \bar c_{L_\alpha},
\end{equation}
where we assume small $\bar \alpha$.
We set $c_{D_T} = 0$, similar to $c_{D_V}$.

Flap deflection changes the lift coefficient in two major ways: change in angle of attack and change in airfoil camber~\cite{anderson1999aircraft}.
For simplicity, we only consider the change in angle of attack due to flap deflection, so that for small angles
\begin{multline}\label{eq:flaplift}
\frac{1}{2}\rho \|\vect{v}\|^2 S \bar c_{L_\alpha} \frac{c_f}{2c}\sum_{i}\delta_i = c_{L_V}^\delta \|\vect{v}\|\xelem \vect{v}^\alpha \sum_{i}\delta_i \\
\Rightarrow c_{L_V}^\delta = \frac{1}{2}\rho S \bar c_{L_\alpha} \frac{c_f}{2c}= \frac{c_f}{2c} c_{L_V}
\end{multline}
where $c$ is the wing chord, $c_f$ is the flap chord, and the division by 2 in the righthand side is due to the fact that each flap is attached to half of the wing.
Similarly, we obtain $c_{L_T}^\delta$ based on $c_{L_T}$, as follows
\begin{equation}\label{eq:flaptrustlift}
c_{L_T}^\delta = \frac{c_f}{c}c_{L_T}.
\end{equation}

Wing properties corresponding to the tailsitter aircraft shown in Fig. \ref{fig:aircraft} are given in Table \ref{tab:wing_shape}.
Using these values and $\rho = 1.225$ kg/m\textsuperscript{3} for standard sea-level conditions, we obtain the aerodynamic parameters given in the middle column of Table \ref{tab:aeroparams}.
We set the thrust pitch moment coefficient $c_{\mu_T}$ to zero for now, because its analytical estimation is complicated by the propeller-wing interaction.
The flap pitch moment effectiveness $l_{\delta_x}$ is set to 0.075 m based on the location of the flap quarter-chord point.

\begin{table}[!ht]
	\centering
	\caption{Aerodynamic force parameters.}
	\label{tab:aeroparams}
	\begin{tabular}{llll}
		\hline
		Parameter & Analytical & Experimental& \\
		\hline
		$c_{L_V}$& 0.17 & 0.29 & kg/m\\
		$c_{D_V}$& 0& 0& kg/m\\
		$c_{L_T}$& 3.4& 2.23 \\
		$c_{D_T}$& 0& 0& \\
		$c^\delta_{L_V}$& 0.041 & 0.18& kg/m\\
		$c^\delta_{L_T}$& 1.7 & 1.25& \\
		\hline
	\end{tabular}
\end{table}

\subsection{Experimental Data}
We found that our control algorithm is able to stabilize the tailsitter in flight when using the analytically obtained aerodynamic parameters from Table \ref{tab:aeroparams}.
The incremental nature of the algorithm enables the controller to compensate for discrepancies in the dynamics model, such as inaccurate aerodynamics parameters.
Despite the fact that the parameters were obtained using small angle of attack assumptions, we found that they also enable stable flight at large angles of attack and in static hover.
A description of the experimental setup is given in Section \ref{sec:setup}.

We use experimental data to improve our estimate of the aerodynamic parameters.
Specifically, we use the fact that the force contributions in the zero-lift reference frame $\vect{f}^\alpha_T$, $\vect{f}^\alpha_\delta$, and $\vect{f}^\alpha_w$ are linear in the parameters.
This allows us to formulate the parameter estimation problem as a multiple linear regression.
For the drag coefficients we have
\begin{equation}\label{eq:ls_x}
\xelem \vect{f}^\alpha - \cos \bar\alpha \sum_{i}T_i = \left[
\begin{array}{cc}
c_{D_V} & c_{D_T}\\
\end{array}
\right]\left[\begin{array}{c}
-\xelem \vect{v}^\alpha \|\vect{v}\|\\
-\cos \bar\alpha \sum_{i} T_i
\end{array}\right],
\end{equation}
and for the lift coefficients we have
\begin{multline}\label{eq:ls_z}
\zelem \vect{f}^\alpha + \sin \bar\alpha \sum_{i}T_i \\
= \left[\begin{array}{cccc}
c_{L_V}&c_{L_T}&c^\delta_{L_V}&c^\delta_{L_T}\\
\end{array}\right]
\left[\begin{array}{c}
-\zelem \vect{v}^\alpha \|\vect{v}\|\\
\sin\bar\alpha \sum_{i} T_i\\
-\xelem \vect{v}^\alpha \|\vect{v}\| \sum_{i}\delta_i\\
-\cos \bar\alpha \sum_{i}\delta_i T_i\\
\end{array}\right].
\end{multline}
We perform a flight with coordinated turns in both directions.
The flight starts from static hover, after which the vehicle reaches speeds up to 4.5 m/s in between turns.
We estimate the forces in the zero-lift reference frame based on gravity-corrected acceleration measurements.

\begin{figure*}
	\centering

	\begin{subfigure}[t]{0.31\textwidth}
		\centering
		\includegraphics[width=\linewidth]{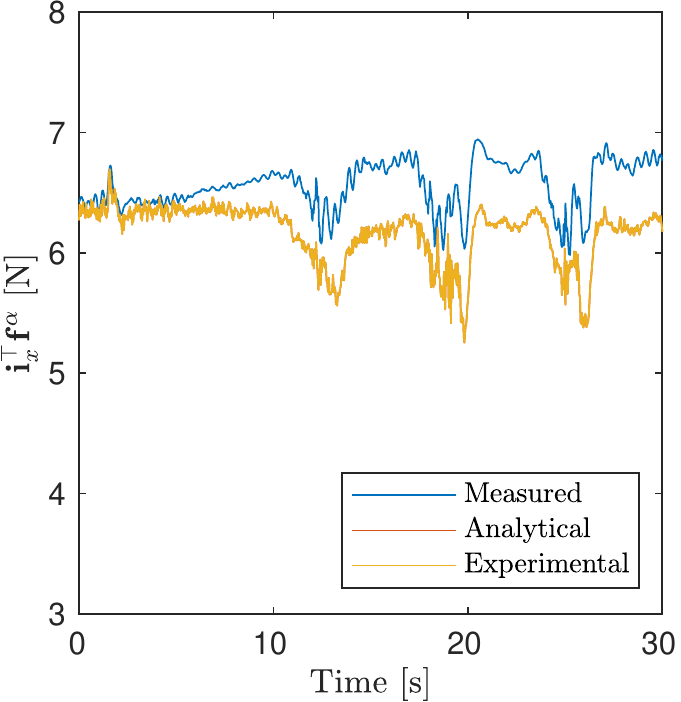}
		\caption{Component along $\vect{\alpha}_x$.}
		\label{fig:force_x}
	\end{subfigure}%
	\quad
	\begin{subfigure}[t]{0.31\textwidth}
		\centering
		\includegraphics[width=\linewidth]{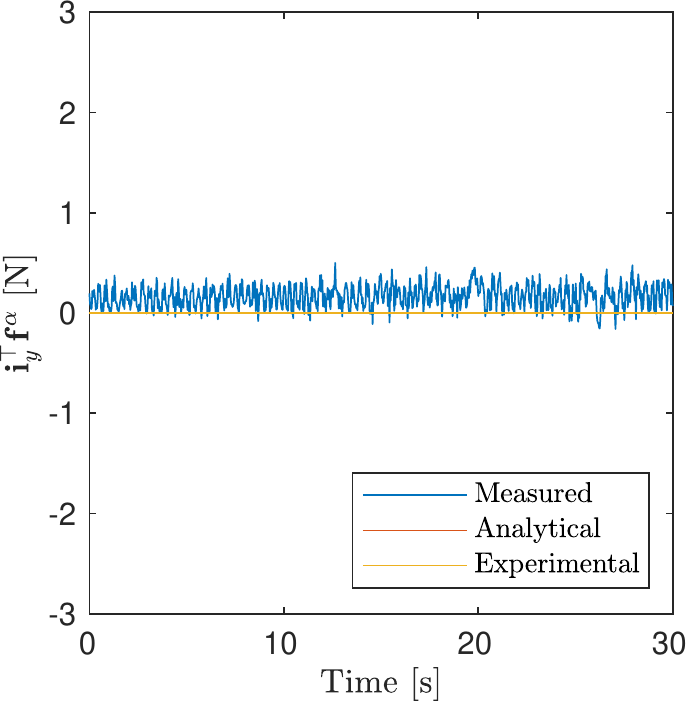}
		\caption{Component along $\vect{\alpha}_y$.}
		\label{fig:force_y}
	\end{subfigure}%
	\quad
	\begin{subfigure}[t]{0.31\textwidth}
		\centering
		\includegraphics[width=\linewidth]{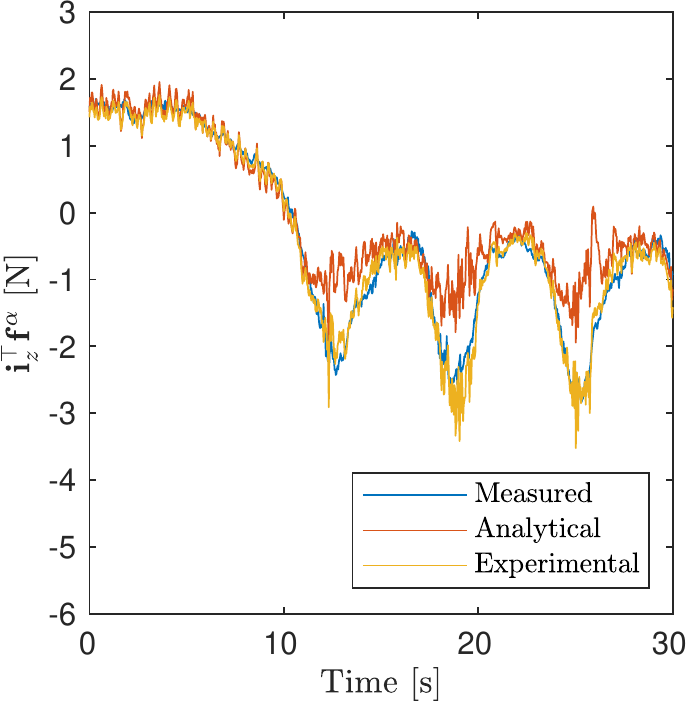}
		\caption{Component along $\vect{\alpha}_z$.}
		\label{fig:force_z}
	\end{subfigure}%
	\caption{Forces in zero-lift reference frame based on measurements, and on analytical and experimental estimates of aerodynamic parameters. The analytical and experimental lines coincide in (a) and (b).}
	\label{fig:force}
\end{figure*}

The force measurements are shown in Fig. \ref{fig:force}, along with estimates based on analytical and experimental aerodynamic parameters.
Since we set both $c_{D_V}$ and $c_{D_T}$ to zero, the analytical force estimate along $\vect{\alpha}_x$, shown in Fig. \ref{fig:force_x}, consists solely of the direct thrust contribution.
At the beginning of the trajectory, where the vehicle is in static hover, this analytical estimate closely matches the measured force, meaning that the presence of the wing has no significant effect on the thrust magnitude and $c_{D_T}$ is indeed close to zero.
As the speed increases, the measured forward force increases beyond the analytical thrust force estimate.
This may be due to (a combination of) various reasons.
The actual thrust may be underestimated because of increasing efficiency of the high-pitch propellers as the blade angle of attack is reduced in forward flight.
Another reason may be the lift-induced forward force given by \eqref{eq:ca}.
Possibly, this forward force eclipses the parasitic drag force acting in the opposite direction at the speeds we achieve in the indoor flight space.
Regardless of its exact cause, the increasing forward force drives us to set $c_{D_V}$ to zero.
When flying at higher speeds, it may be possible to obtain an accurate nonzero estimate of $c_{D_V}$ using the regression equation.
We found that, in practice, the discrepancy in the forward force estimate has little influence on controller performance.
It is closely aligned with the thrust force and can thus be counteracted quickly and accurately by the incremental controller.
The lateral force component, shown in Fig. \ref{fig:force_y}, is close to zero, as expected.
The small bias in its measurement may be due to an imbalance or misalignment.
The measured and estimated force component along $\vect{\alpha}_z$ is shown in Fig. \ref{fig:force_z}.
It can be seen that the estimate based on the analytical parameters has a significant deviation at increased speeds.
The estimate based on the experimental regression parameters closely matches the measurements throughout the trajectory with \revised{a coefficient of determination, \ie, $R^2$ value,} of 0.97.
The resulting parameters are given in the final column of Table \ref{tab:aeroparams} and have similar order of magnitude as the corresponding analytical parameters.
The analytical underestimation of $c^\delta_{L_V}$ may be because the change of camber due to flap deflection is not considered.

We also attempted to use multiple linear regression on the pitch moment $\yelem \vect{m}$ to obtain experimental estimates for $c_{\mu_T}$ and $l_{\delta_x}$.
This approach did not result in consistent parameters and good moment predictions;
most likely due to significant moment contributions that are not captured by the simplified aerodynamic model, \eg, due to airspeed, angle of attack, and angular rates.
Modeling these contributions is not required for our control design, which relies only on an incremental expression that relates the change in moment to changes in differential thrust and flap deflections.
However, their absence in the aerodynamic model makes accurate estimation of $c_{\mu_T}$ and $l_{\delta_x}$ more difficult.
We found that the initial estimate of $l_{\delta_x} = 0.075$ m results in satisfactory pitch control, so we leave this value unchanged.
Finally, we set $c_{\mu_T}$ based on the trim flap deflections in static hover, as follows:
\begin{equation}\label{eq:pitchcoeff}
c_{\mu_T} = \frac{\delta}{2} l_{\delta_x} c^\delta_{L_T} \cos \bar\alpha.
\end{equation}
For our vehicle, we found $\nicefrac{\delta}{2} = -0.27$ rad, resulting in $c_{\mu_T} = -0.025$ m.

The flight test results presented in Section \ref{sec:experiments} were obtained using the experimental parameter estimates.
We found that these parameters result in improved trajectory-tracking performance when compared to the analytical parameters.
As expected, the difference is most significant at increased speeds where the discrepancy between the force estimates is largest.
Specifically, we found that altitude oscillations may occur due to the underestimated $c_{L_V}$ and $c^\delta_{L_V}$ analytical parameters.

\section{Experimental Results}\label{sec:experiments}
In this section, we evaluate controller performance on various trajectories that include challenging flight conditions, such as large accelerations, transition on curved trajectories, and uncoordinated flight.
\revised{An overview of the trajectories is given in Table \ref{tab:control_exp_error}.}
Additionally, in Appendix \ref{sec:baselinecomp} we present a comparison to a baseline version of the controller that illustrates the necessity of feedforward jerk and yaw rate tracking and of robustification using incremental control.
Further flight experiments, focusing on aerobatic maneuvers, can be found in \cite{tal2022aerobatic}.
Video of the experiments is available at \url{https://aera.mit.edu/projects/TailsitterAerobatics}.

\begin{table*}[!ht]
	\centering
	\footnotesize
	\caption{Maximum speed, load, and angular rate; and maximum and root mean square (RMS) position tracking error for flight experiments.}
	\label{tab:control_exp_error}
	\begin{tabular}{lccccc}
		\hline
		& max$\|\vect{v}\|$ [m/s]&max$\|\vect{a} - \vect{i}_z g\|$ [$g$]& max$\|\vect{\Omega}\|$ [deg/s] & max$\|\vect{x}_{\sref}-\vect{x}\|$ [m] & RMS$\|\vect{x}_{\sref}-\vect{x}\|$ [m]\\
		\hline
		Lemniscate               & 6.2 & 1.7 & 133 & 0.17 & 0.33\\
		Knife Edge Transitioning & 6.6 & 2.1 & 277 & 0.20 & 0.48\\
		Circular (Coordinated)   & 7.8 & 2.2 & 154 & 0.15 & 0.18\\
		Circular (Knife Edge)    & 7.9 & 2.1 & 140 & 0.15 & 0.17\\
		Transition from Hover    & 8.3 & 2.1 & 146 & 0.10 & 0.15\\
		Transition to Hover      & 8.0 & 2.2 & 164 & 0.15 & 0.24\\
		Differential Thrust Turn & 7.3 & 2.2 & 661 & 0.63 & 0.96\\
		\hline
	\end{tabular}
\end{table*}

\subsection{Experimental Setup}\label{sec:setup}
Experiments were performed in an 18 m $\times$ 8 m indoor flight space using the tailsitter vehicle shown in Fig. \ref{fig:aircraft} and described in \cite{bronz2020mission}.
The vehicle is 3D-printed using Onyx filament with carbon fiber reinforcement.
It weighs 0.7 kg and has a wingspan of 55 cm from tip to tip.
It is equipped with two T-Motor F40 2400 KV motors \revised{and Gemfan Hulkie 5055 propellers with 5.0 inch tip diameter and 5.5 inch advance pitch.}
The motor speeds are measured using optical encoders at one measurement per rotation, \ie, at about 200 Hz in hover.
MKS HV93i micro servos are used to control the flaps.
We obtain the flap deflection measurement as an analog signal by connecting a wire to the potentiometer in the servo.

The flight control algorithm runs onboard on an STM32 microcontroller using custom firmware.
The microcontroller has a clock speed of 400 MHz and takes 25 $\mu$s to compute a control update at 32-bit floating point precision.
Accelerometer and gyroscope measurements are obtained from an Analog Devices ADIS16475-3 IMU at 2000 Hz, and control updates are performed at the same frequency.
Position and attitude measurements are provided by a motion capture tracking system at 360 Hz and sent to the vehicle over Wi-Fi with an average latency of 18 ms.
\rrevised{The motion capture data is propagated using IMU measurements to correct for the latency.}

\revised{We used time-domain flight data to find control gains that attain the fast rise time required for aggressive maneuvering while maintaining small overshoot.
Second-order Butterworth filters are applied for signal processing, because this type of filter provides the most uniform sensitivity in the passband.
In order to achieve high-bandwidth disturbance rejection, we set the cutoff frequency of the low-pass filters to the maximum frequency where controller performance is not affected by inertial measurement noise; in our case, 15 Hz.
While the motor speed and flap deflection measurements contain relatively much less noise, we apply identical low-pass filters to these signals to avoid phase difference.
The transient flap deflections are obtained by subsequent filtering using a second-order Butterworth high-pass filter with cutoff frequency of 1 Hz.
We raised this cutoff to increase linear acceleration control authority, until we observed pitch oscillations due to the non-minimum phase acceleration response.}

\subsection{Lemniscate Trajectory}

\begin{figure*}
	\centering
	
	\begin{subfigure}[t]{0.9\textwidth}
		\centering
		\includegraphics[width=\linewidth]{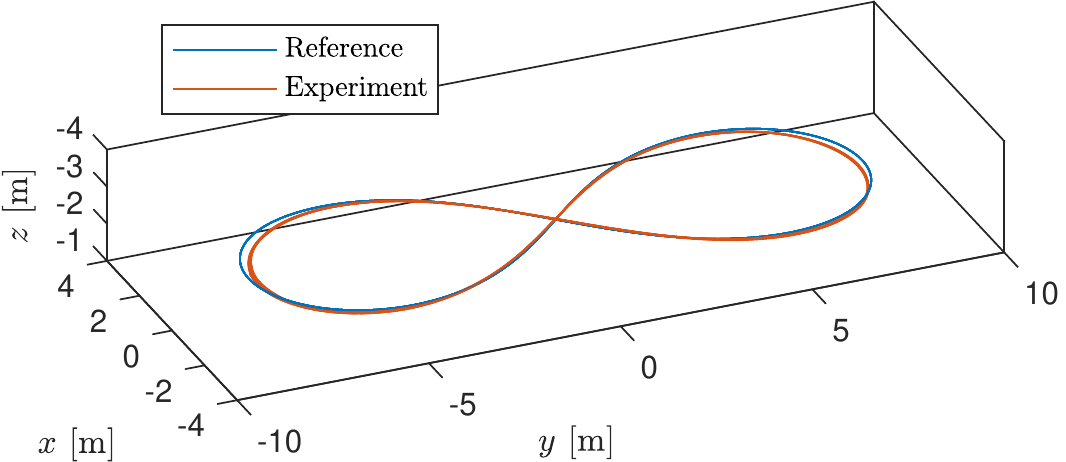}
		\caption{Position.}
		\label{fig:lemniscate_position}
	\end{subfigure}%
	
	\begin{subfigure}[t]{0.485\textwidth}
		\centering
		\includegraphics[width=\linewidth]{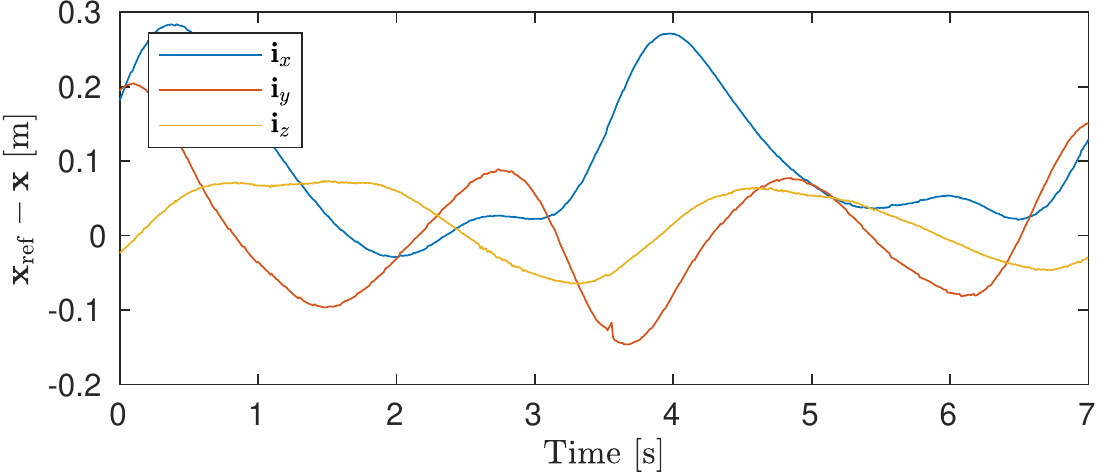}
		\caption{Position tracking error.}
		\label{fig:lemniscate_poserror}
	\end{subfigure}%
	\quad
	\begin{subfigure}[t]{0.485\textwidth}
		\centering
		\includegraphics[width=\linewidth]{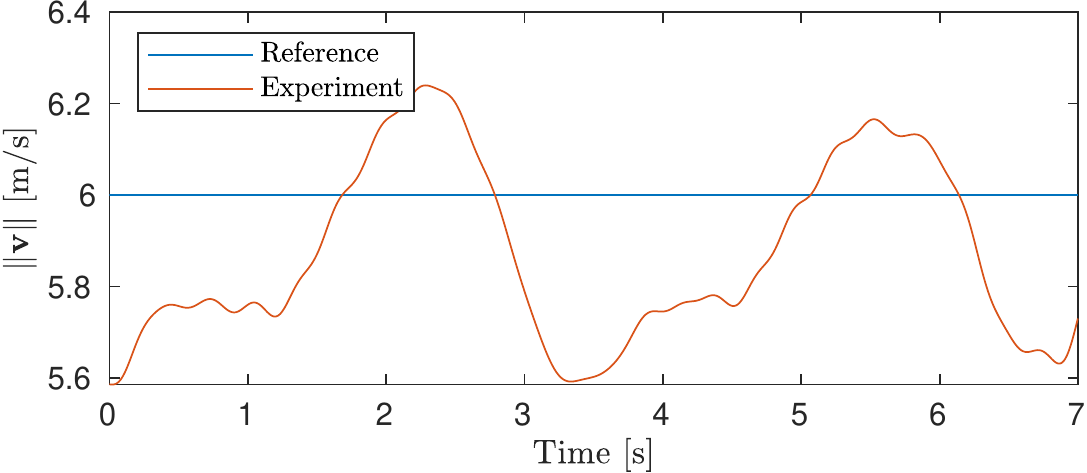}
		\caption{Speed.}
		\label{fig:lemniscate_speed}
	\end{subfigure}

	\begin{subfigure}[t]{0.485\textwidth}
	\centering
	\includegraphics[width=\linewidth]{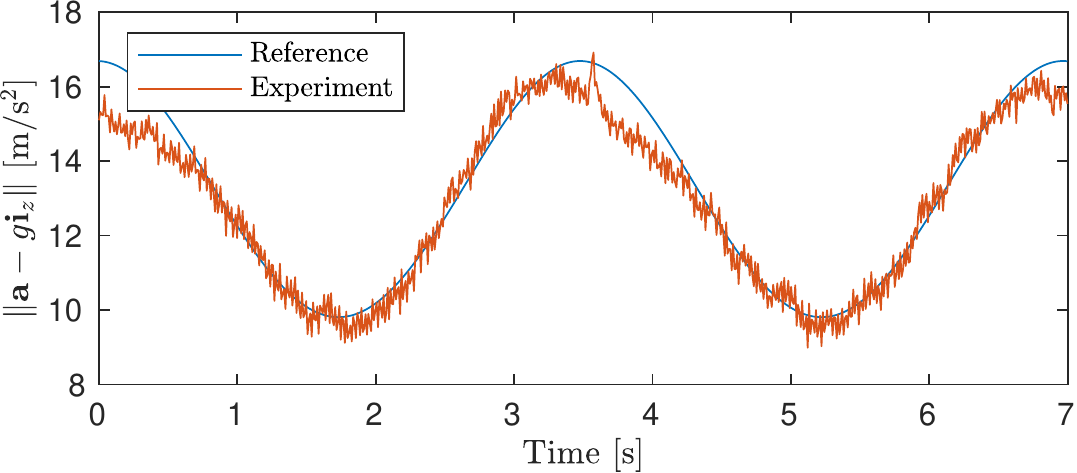}
	\caption{Acceleration.}
	\label{fig:lemniscate_acceleration}
\end{subfigure}%
\quad
\begin{subfigure}[t]{0.485\textwidth}
	\centering
	\includegraphics[width=\linewidth]{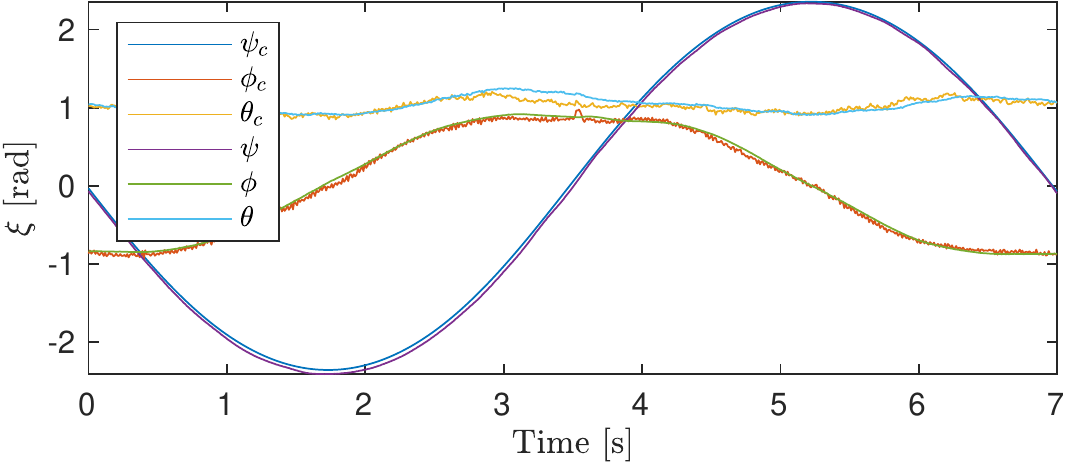}
	\caption{Attitude.}
	\label{fig:lemniscate_attitude}
\end{subfigure}
	\caption{Experimental results for lemniscate trajectory at 6 m/s.}
	\label{fig:lemniscate}
\end{figure*}

Figure \ref{fig:lemniscate} shows experimental results for tracking of a Lemniscate of Bernouilli with a constant speed of 6 m/s.
Throughout the trajectory, $\psi_{\sref}$ is set perpendicular to the velocity, leading to coordinated flight.
The reference and flown trajectories over eight consecutive laps (of 7 s each) starting at the $y$ extreme are shown in Fig. \ref{fig:lemniscate_position}.
It can be seen that the tracking performance is very consistent between laps.
Figure \ref{fig:lemniscate_poserror} shows that the largest position deviation occurs at the end of the circular parts where the vehicle does not fully maintain the reference acceleration of almost 2 $g$, as can be seen in Fig. \ref{fig:lemniscate_acceleration}.
Over the middle part of the trajectory, the vehicle increases speed to catch up (see Fig. \ref{fig:lemniscate_speed}), and the position error is reduced again.
Overall, the controller achieves a position tracking error of 17 cm root mean square (RMS) with a maximum error of 33 cm. 

Figure \ref{fig:lemniscate_attitude} shows the commanded and flown attitude.
The attitude controller uses quaternion representation, but to ease interpretation the figure uses the ZXY Euler angles described in Section \ref{sec:invdyn}.
For this trajectory, $\phi$ corresponds to the bank angle and reaches almost 1 rad, which matches the acceleration nearing 2 $g$ in Fig. \ref{fig:lemniscate_acceleration}.
The maximum attitude error occurs during a small period in the turn, where the vehicle incurs a pitch error of 6 deg.
Controlling the pitch angle of a flying wing during aggressive maneuvers is generally challenging due to the lack of an elevator, and the pitch deviation is likely the cause of the relatively large position deviation at the exit of the turn.
Overall, the controller is able to track the dynamic attitude command well, and it maintains an attitude error of less than 2 deg on each axis during the rest of the trajectory.

\subsection{Knife Edge Transitioning Flight}\label{sec:exp_ke}
\begin{figure*}
	\centering
	
	\begin{subfigure}[t]{0.9\textwidth}
		\centering
		\includegraphics[width=\linewidth]{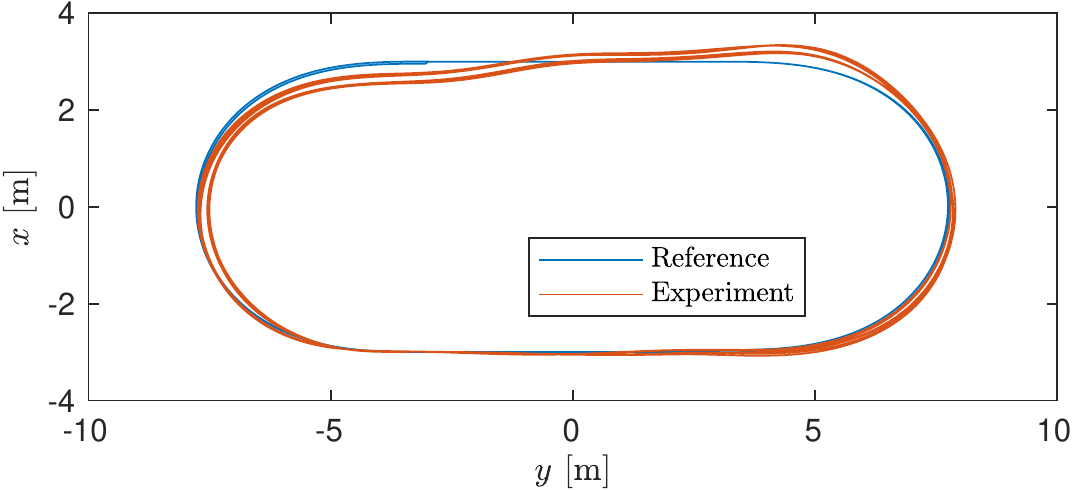}
		\caption{Position.}
		\label{fig:knifeedge_position}
	\end{subfigure}%
	
	\begin{subfigure}[t]{0.485\textwidth}
		\centering
		\includegraphics[width=\linewidth]{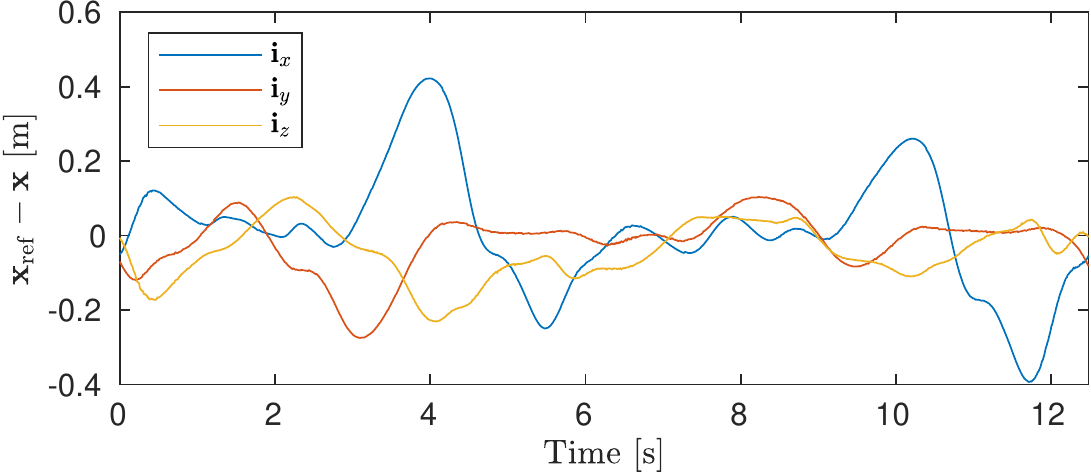}
		\caption{Position tracking error.}
		\label{fig:knifeedge_poserror}
	\end{subfigure}%
	\quad
	\begin{subfigure}[t]{0.485\textwidth}
		\centering
		\includegraphics[width=\linewidth]{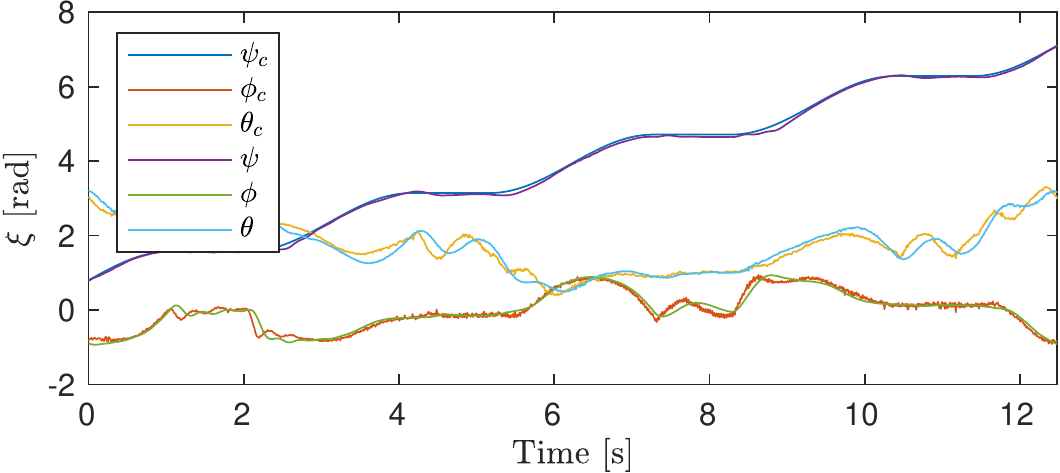}
		\caption{Attitude.}
		\label{fig:knifeedge_attitude}
	\end{subfigure}
	\caption{Experimental results for knife edge transitioning trajectory at 6 m/s.}
	\label{fig:knifeedge}
\end{figure*}

Our proposed algorithm is able to control the vehicle in uncoordinated flight conditions where it has significant lateral velocity.
In knife edge flight, the wingtip is pointing in the velocity direction, leading to instability due to the location of the center of gravity behind the quarter span point~\cite{smeur2020incremental}.
Figure \ref{fig:knifeedge} shows experimental results for a trajectory where $\psi_{\sref}$ is set to enforce knife edge flight on one side and coordinated flight on the other side.
The results show that our controller is able to stabilize the knife edge flight condition, and that it is able to transition between knife edge and coordinated flight while at the same time performing a 1.6 $g$ turn.

One lap of the oval trajectory takes 6.25 s to complete at a constant speed of 6 m/s.
Referring to the view from above in Fig. \ref{fig:knifeedge_position}, the trajectory is flown in clockwise direction with the straight at the top in knife edge configuration and the straight at the bottom in coordinated flight.
During each turn, the yaw reference is rotated by $\nicefrac{\pi}{2}$ rad to enforce the switch between configurations.
Consequently, the coordinated flight segment is flown in inverted orientation every other lap.
Figure \ref{fig:knifeedge_position} shows the reference and flown position over eight successive laps, and
Fig. \ref{fig:knifeedge_poserror} and Fig. \ref{fig:knifeedge_attitude} show data corresponding to two laps starting and ending at the transition from knife edge flight to coordinated flight in inverted orientation.
It can be seen that the reference is followed accurately during both regular and inverted coordinated flight.
Even while performing the transition from knife edge to coordinated flight, the controller is able to track the turning trajectory.
At the transition from coordinated flight to knife edge, we see that the trajectory is shifted during transitions from inverted orientation.
This leads to a position tracking error of at most 48 cm at the start of the knife edge segment.
During transition from regular coordinated flight to knife edge a position error of at most 25 cm is incurred.
Flight during the knife edge segment is consistent and stable, and the yaw reference is tracked within 3 deg.
Over the whole trajectory, the controller achieves tracking of the position reference within 20 cm RMS and the yaw reference within 1.7 deg RMS.

\subsection{Circular Trajectory}

\begin{figure*}
	\centering
	
	\begin{subfigure}[t]{0.485\textwidth}
		\centering
		\includegraphics[width=\linewidth]{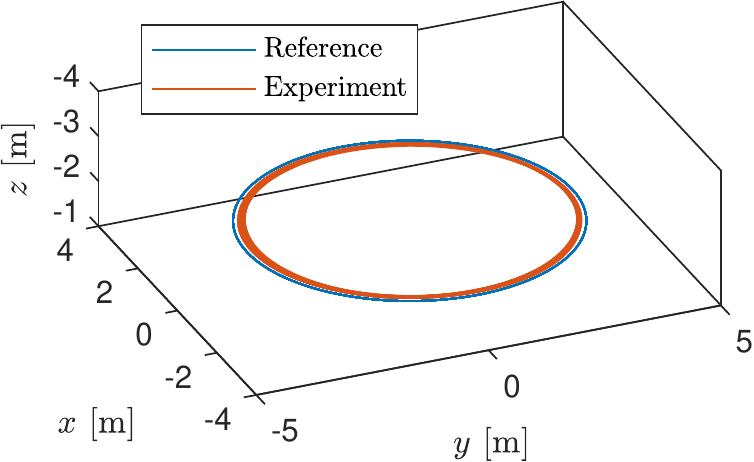}
		\caption{Position.}
		\label{fig:circle_position}
	\end{subfigure}%
	\quad
	\begin{subfigure}[t]{0.485\textwidth}
	\centering
	\includegraphics[width=\linewidth]{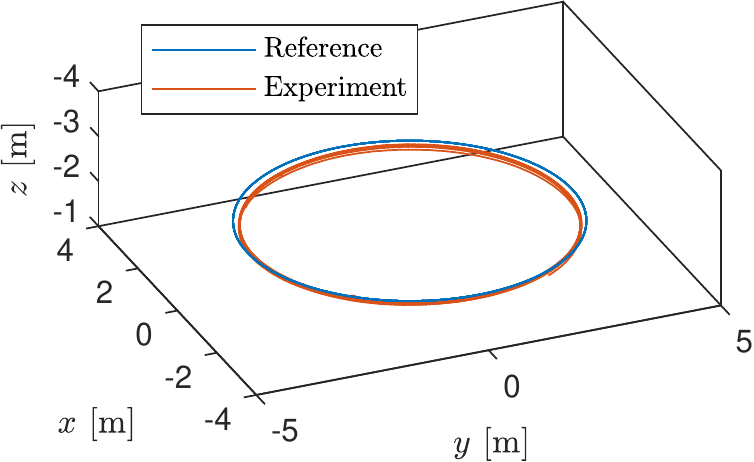}
	\caption{Position.}
	\label{fig:circleke_position}
	\end{subfigure}%
	
	\begin{subfigure}[t]{0.485\textwidth}
	\centering
	\includegraphics[width=\linewidth]{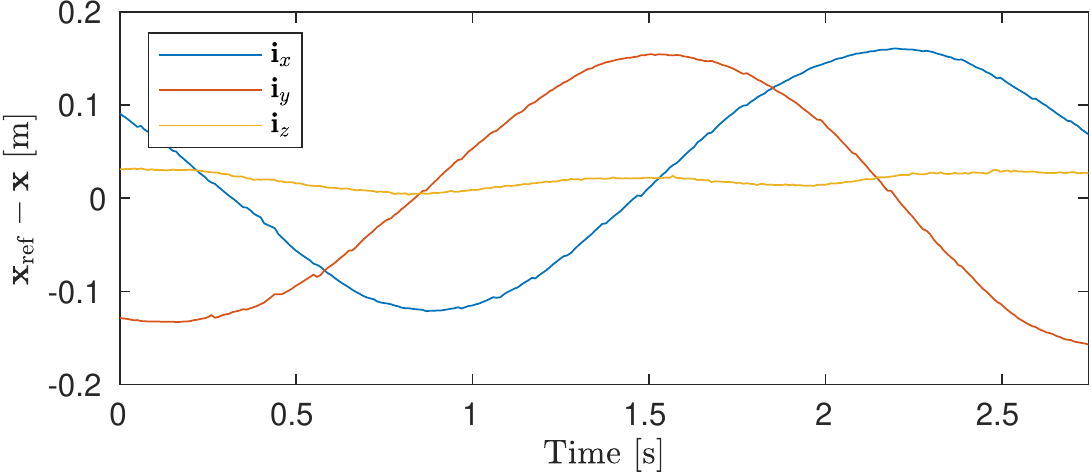}
	\caption{Position tracking error.}
	\label{fig:circle_poserror}
	\end{subfigure}%
	\quad
	\begin{subfigure}[t]{0.485\textwidth}
	\centering
	\includegraphics[width=\linewidth]{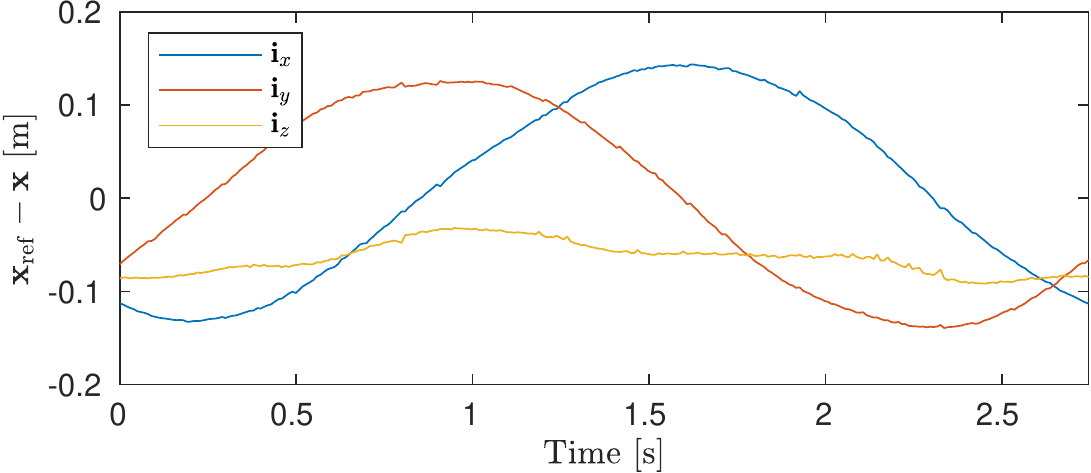}
	\caption{Position tracking error.}
	\label{fig:circleke_poserror}
	\end{subfigure}%

	\begin{subfigure}[t]{0.485\textwidth}
	\centering
	\includegraphics[width=\linewidth]{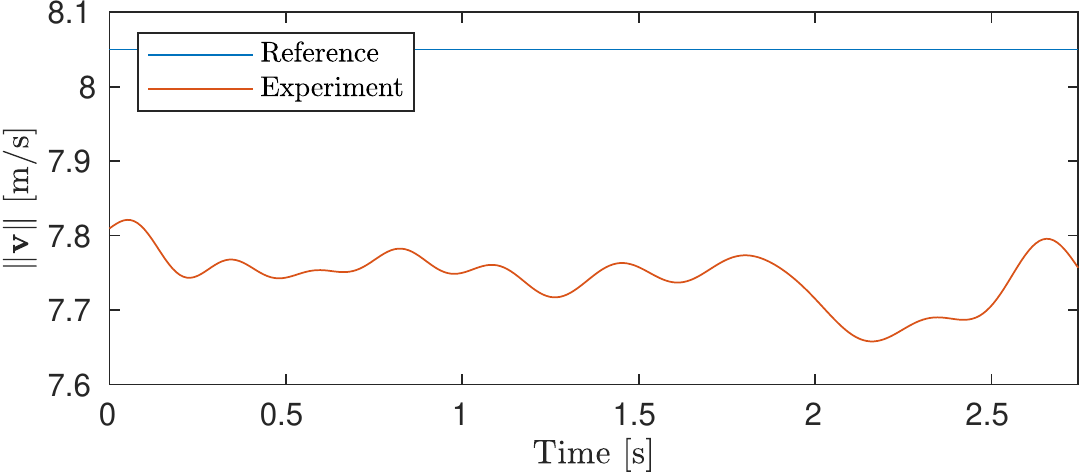}
	\caption{Speed.}
	\label{fig:circle_speed}
\end{subfigure}%
\quad
\begin{subfigure}[t]{0.485\textwidth}
	\centering
	\includegraphics[width=\linewidth]{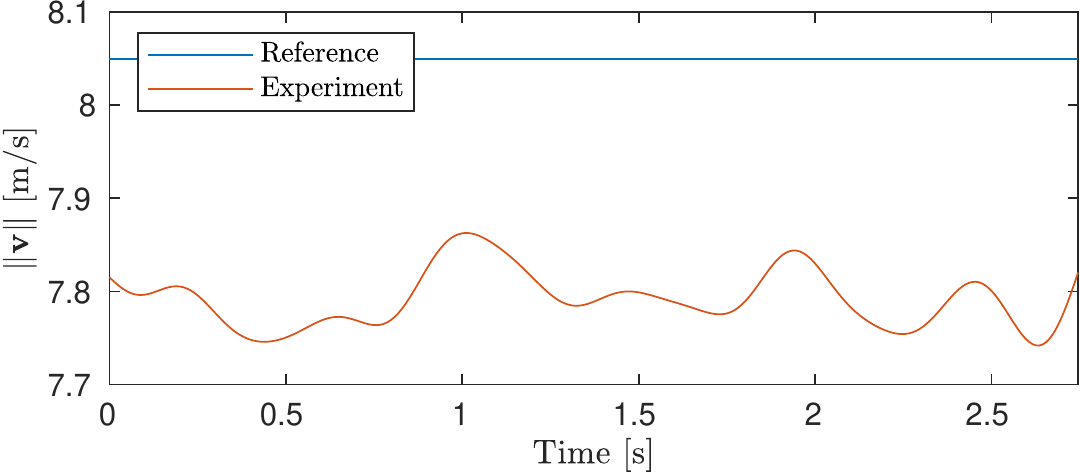}
	\caption{Speed.}
	\label{fig:circleke_speed}
\end{subfigure}%

	\begin{subfigure}[t]{0.485\textwidth}
	\centering
	\includegraphics[width=\linewidth]{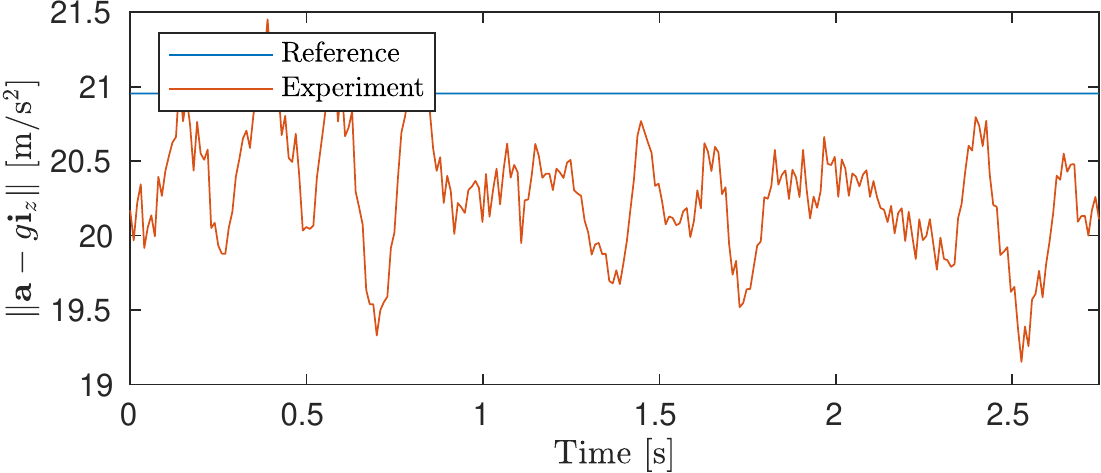}
	\caption{Acceleration.}
	\label{fig:circle_accel}
	\end{subfigure}%
	\quad
	\begin{subfigure}[t]{0.485\textwidth}
	\centering
	\includegraphics[width=\linewidth]{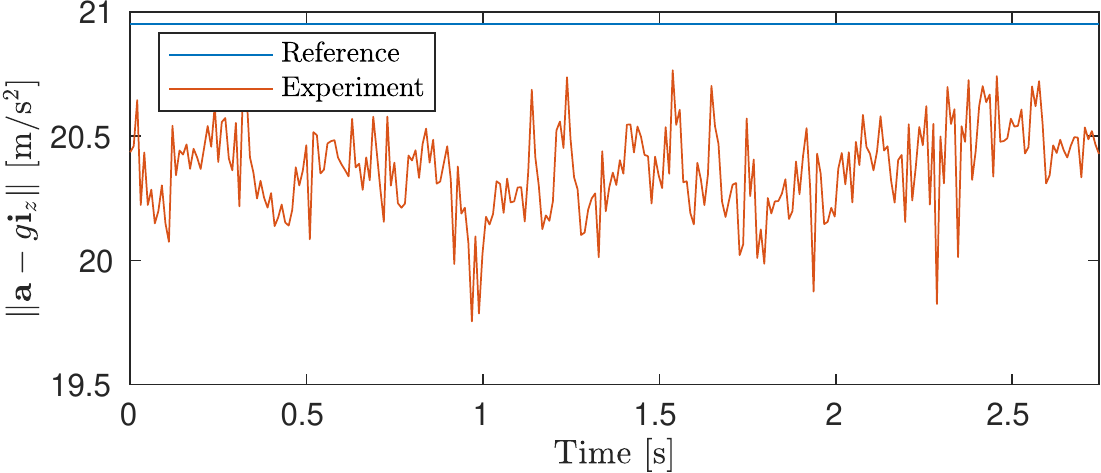}
	\caption{Acceleration.}
	\label{fig:circleke_accel}
	\end{subfigure}%

	\begin{subfigure}[t]{0.485\textwidth}
	\centering
	\includegraphics[width=\linewidth]{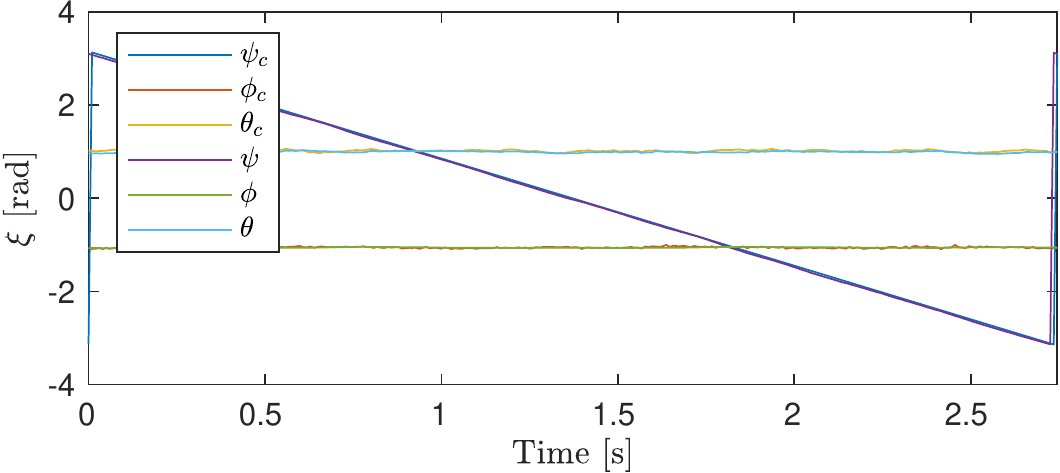}
	\caption{Attitude.}
	\label{fig:circle_attitude}
	\end{subfigure}%
	\quad
	\begin{subfigure}[t]{0.485\textwidth}
	\centering
	\includegraphics[width=\linewidth]{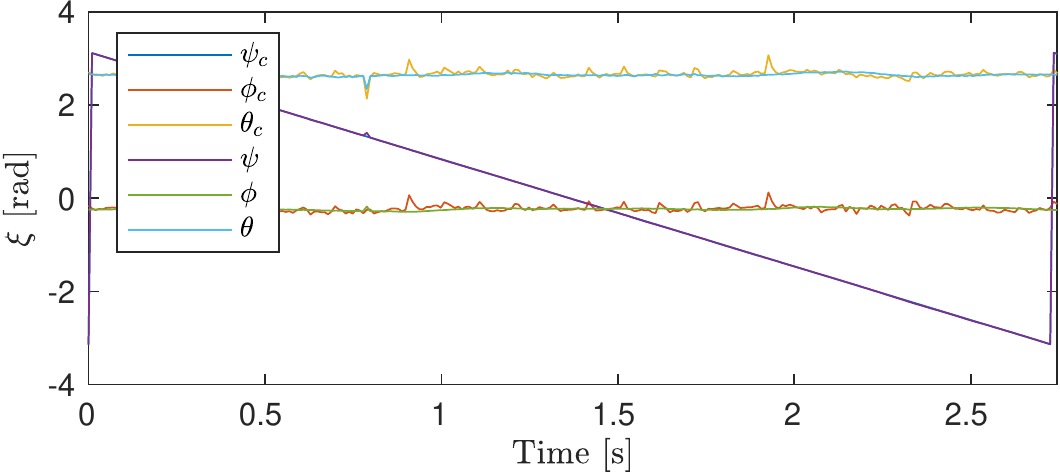}
	\caption{Attitude.}
	\label{fig:circleke_attitude}
	\end{subfigure}%

	\caption{Experimental results for circle trajectory at 7.8 m/s for coordinated flight in (a), (c), (e), (g), and (i); and for knife edge flight (b), (d), (f), (h), and (j).}
	\label{fig:circle}
\end{figure*}

\begin{figure*}
	\centering
	\begin{subfigure}[t]{0.485\textwidth}
	\centering
	\includegraphics[width=\linewidth]{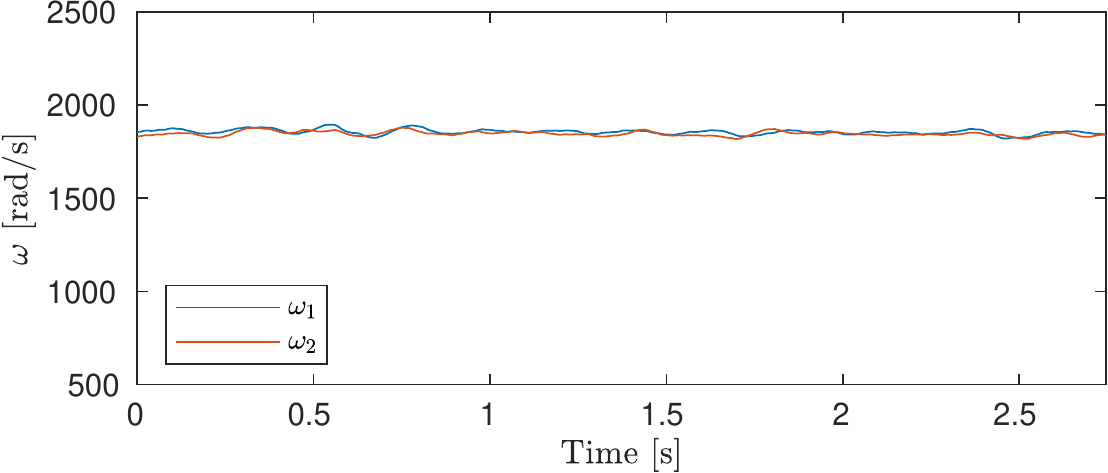}
	\caption{Rotor speed.}
	\label{fig:circle_rpm}
\end{subfigure}%
\quad
\begin{subfigure}[t]{0.485\textwidth}
	\centering
	\includegraphics[width=\linewidth]{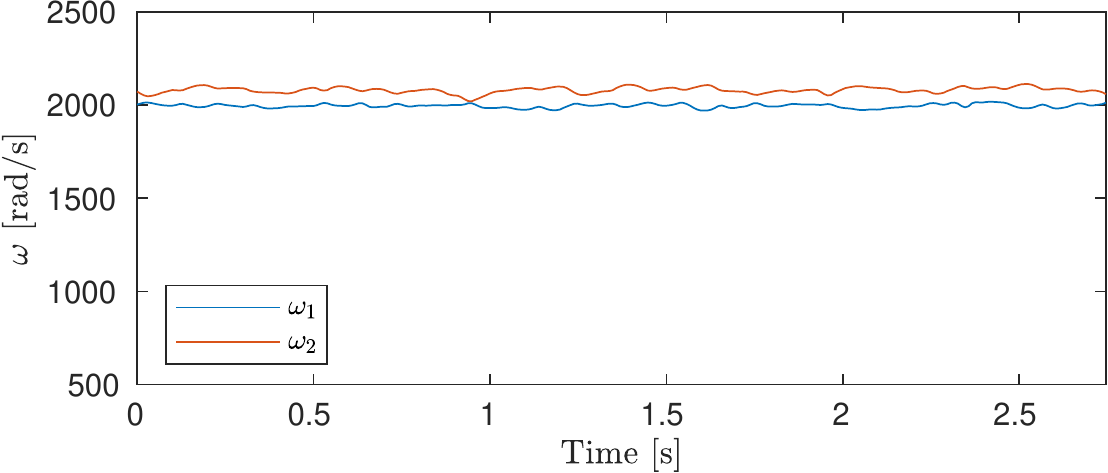}
	\caption{Rotor speed.}
	\label{fig:circleke_rpm}
\end{subfigure}%

\begin{subfigure}[t]{0.485\textwidth}
	\centering
	\includegraphics[width=\linewidth]{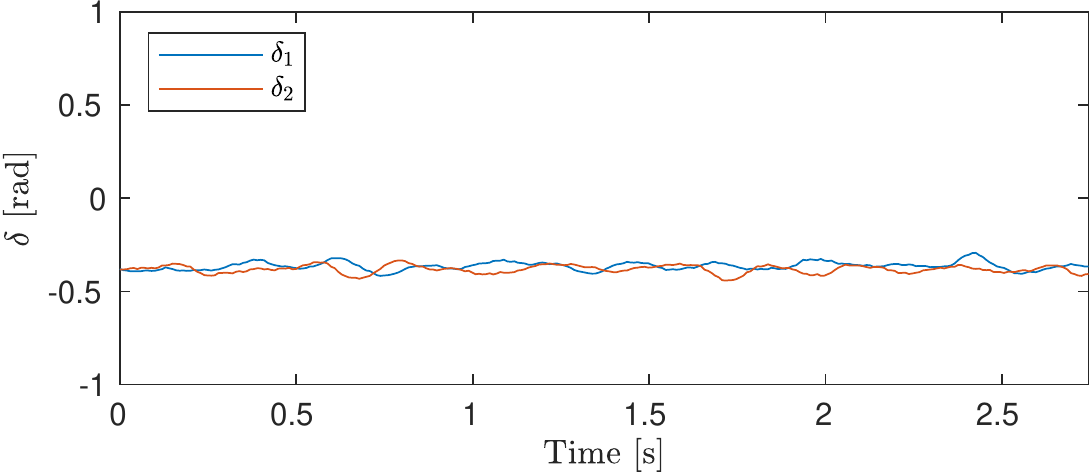}
	\caption{Flap deflection.}
	\label{fig:circle_flap}
\end{subfigure}
\quad
\begin{subfigure}[t]{0.485\textwidth}
	\centering
	\includegraphics[width=\linewidth]{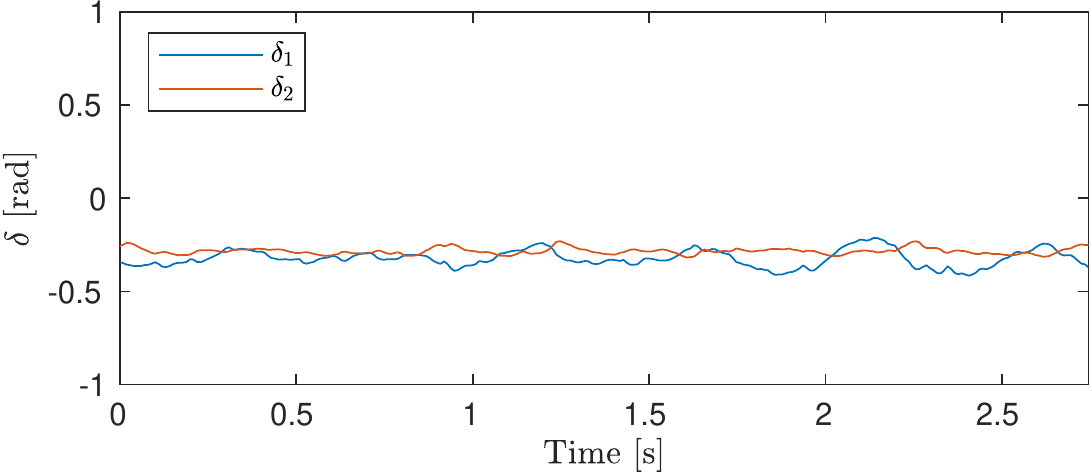}
	\caption{Flap deflection.}
	\label{fig:circleke_flap}
\end{subfigure}


\begin{subfigure}[t]{0.485\textwidth}
	\centering
	\includegraphics[width=\linewidth]{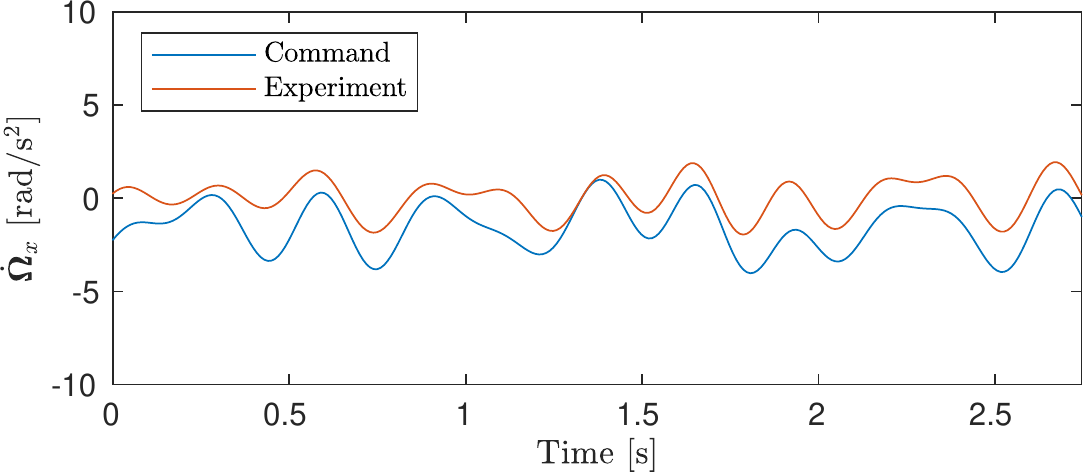}
	\caption{Angular acceleration around $\vect{b}_x$.}
	\label{fig:circle_angacc_x}
\end{subfigure}
\quad
\begin{subfigure}[t]{0.485\textwidth}
	\centering
	\includegraphics[width=\linewidth]{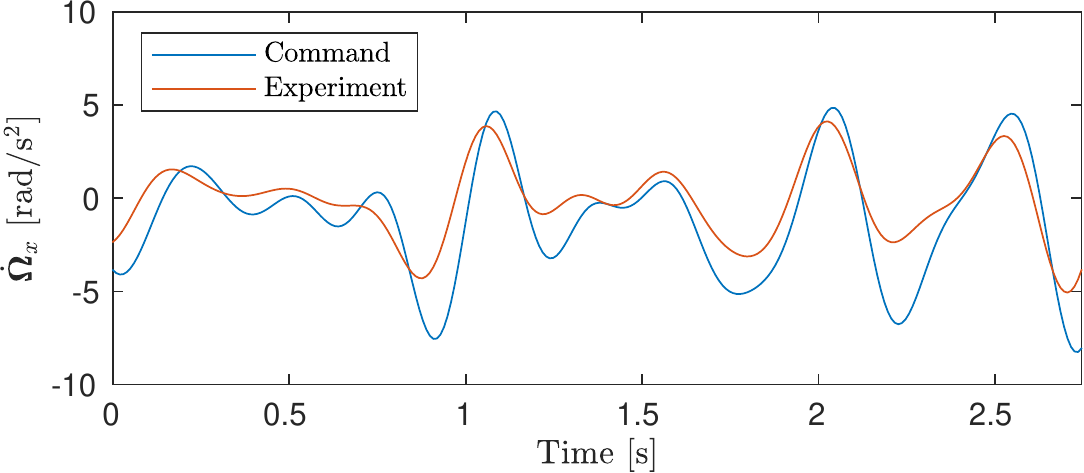}
	\caption{Angular acceleration around $\vect{b}_x$.}
	\label{fig:circleke_angacc_x}
\end{subfigure}

\begin{subfigure}[t]{0.485\textwidth}
	\centering
	\includegraphics[width=\linewidth]{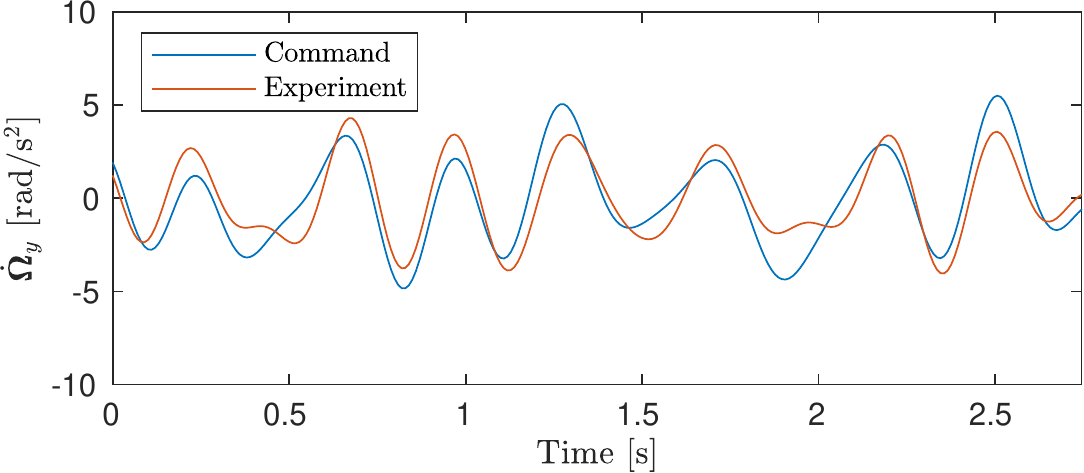}
	\caption{Angular acceleration around $\vect{b}_y$.}
	\label{fig:circle_angacc_y}
\end{subfigure}
\quad
\begin{subfigure}[t]{0.485\textwidth}
	\centering
	\includegraphics[width=\linewidth]{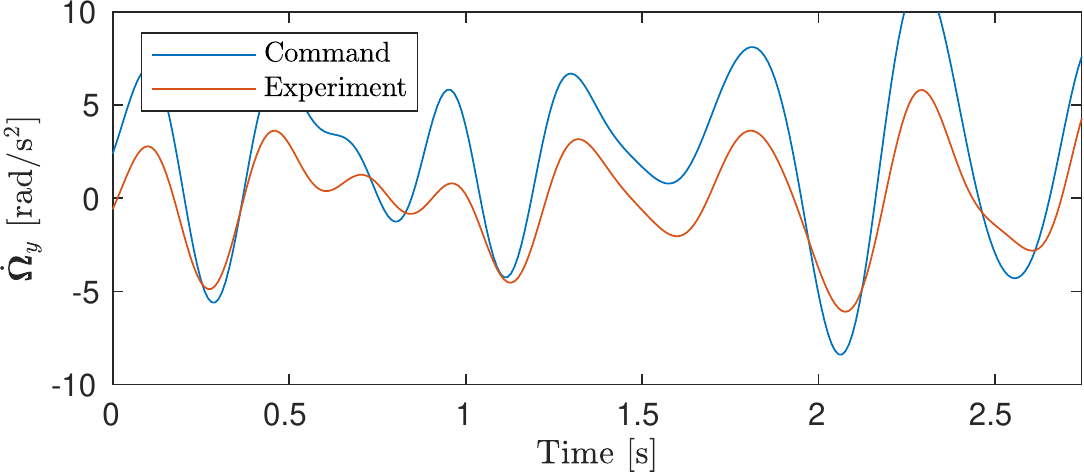}
	\caption{Angular acceleration around $\vect{b}_y$.}
	\label{fig:circleke_angacc_y}
\end{subfigure}

\begin{subfigure}[t]{0.485\textwidth}
	\centering
	\includegraphics[width=\linewidth]{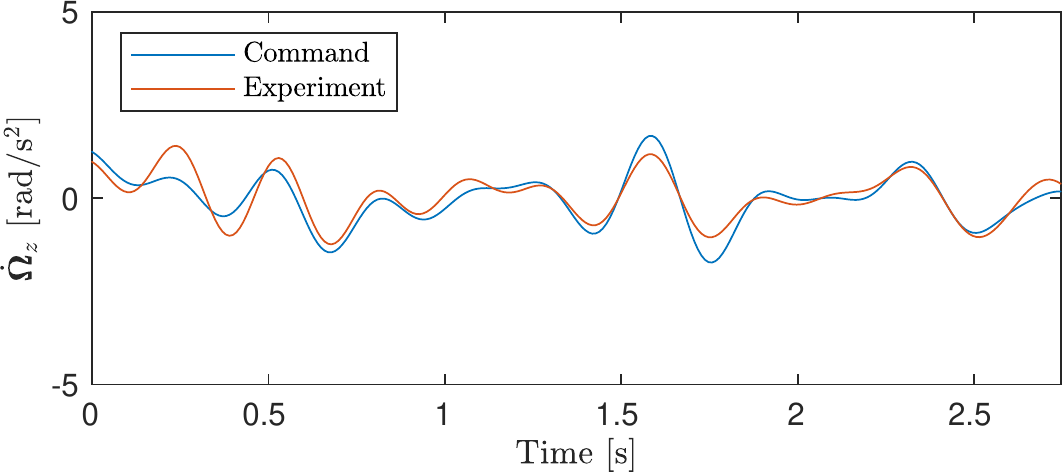}
	\caption{Angular acceleration around $\vect{b}_z$.}
	\label{fig:circle_angacc_z}
\end{subfigure}
\quad
\begin{subfigure}[t]{0.485\textwidth}
	\centering
	\includegraphics[width=\linewidth]{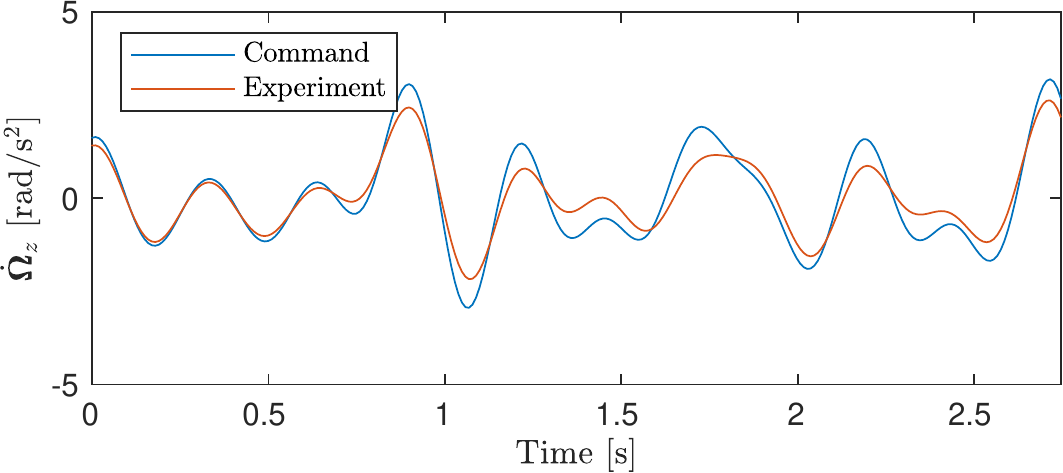}
	\caption{Angular acceleration around $\vect{b}_z$.}
	\label{fig:circleke_angacc_z}
\end{subfigure}

\caption{Experimental results for circle trajectory at 7.8 m/s for coordinated flight in (a), (c), (e), (g), and (i); and for knife edge flight (b), (d), (f), (h), and (j).} 
\label{fig:circle_ctrl}
\end{figure*}

Figure \ref{fig:circle} shows experimental results for tracking of a circular trajectory reference with a 3.5 m radius at a speed of 8.1 m/s.
The left column of figures corresponds to coordinated flight where the $\vect{b}_y$-axis is perpendicular to the circle tangent, and the right column corresponds to knife edge flight where the wing tip points along the tangent of the circle.
Position tracking performance is very similar between both flights.
In both cases, the flown trajectory has a slightly smaller radius than the reference, reducing the flight speed to 7.8 m/s.
The RMS position tracking error is 15 cm for both coordinated and knife edge flight.
The aircraft reaches a continuous acceleration of 2.1 $g$.
In coordinated flight, this requires a bank angle of 63 deg.
In knife edge flight, the aircraft is rolled 14 deg towards the direction of travel to compensate for drag and pitched over by 152 deg to provide thrust towards the circle center.
\revised{As shown in Fig. \ref{fig:circle_ctrl},} maintaining this state requires flap deflections of 20 deg and rotor speeds of over 2000 rad/s in knife edge flight.
In coordinated flight, the aircraft exploits the lift force to achieve circular flight more efficiently and requires rotor speeds of 1800 rad/s with nearly unchanged flap deflections.
In contrast to the trajectory described in Section \ref{sec:exp_ke}, the flap deflections are almost constant during the circular trajectory.
Hence, there are no transient accelerations caused by the flaps that are not considered in the position controller, and very accurate trajectory tracking is achieved in knife edge flight.
This shows that our controller is not only able to stabilize the knife edge flight condition, but also attains accurate trajectory tracking in knife edge flight at speeds close to 8 m/s.
\revised{Figure \ref{fig:circle_ctrl} also shows the commanded and achieved angular accelerations.
In order to improve the readability of these figures, we have shifted the commanded angular acceleration curves by up to 70 ms to account for the response time of the servos and rotors.
It can be seen that the controller is able to follow the commands with decent accuracy despite using only a simplistic model of the aerodynamic moment.
}

\subsection{Transitions}

\begin{figure*}
	\centering
	
	\begin{subfigure}[t]{0.485\textwidth}
		\centering
		\includegraphics[width=\linewidth]{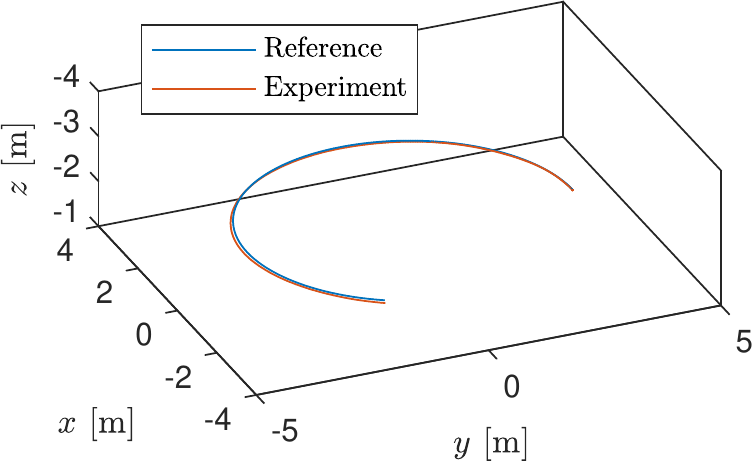}
		\caption{Position.}
		\label{fig:trans3_position}
	\end{subfigure}%
	\quad
	\begin{subfigure}[t]{0.485\textwidth}
		\centering
		\includegraphics[width=\linewidth]{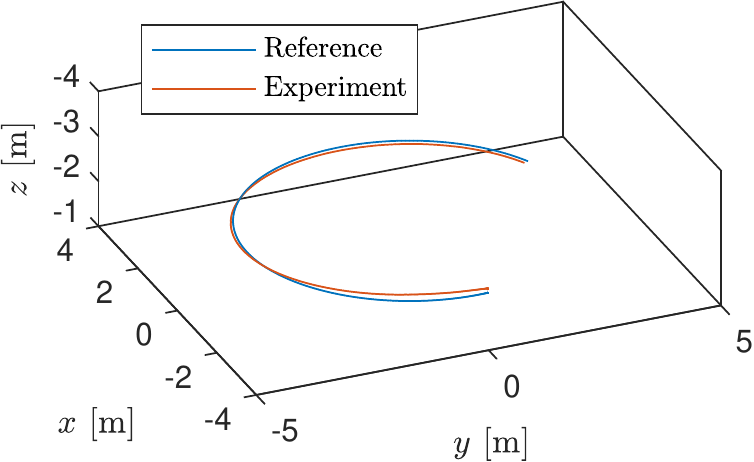}
		\caption{Position.}
		\label{fig:trans3out_position}
	\end{subfigure}%
	
	\begin{subfigure}[t]{0.485\textwidth}
		\centering
		\includegraphics[width=\linewidth]{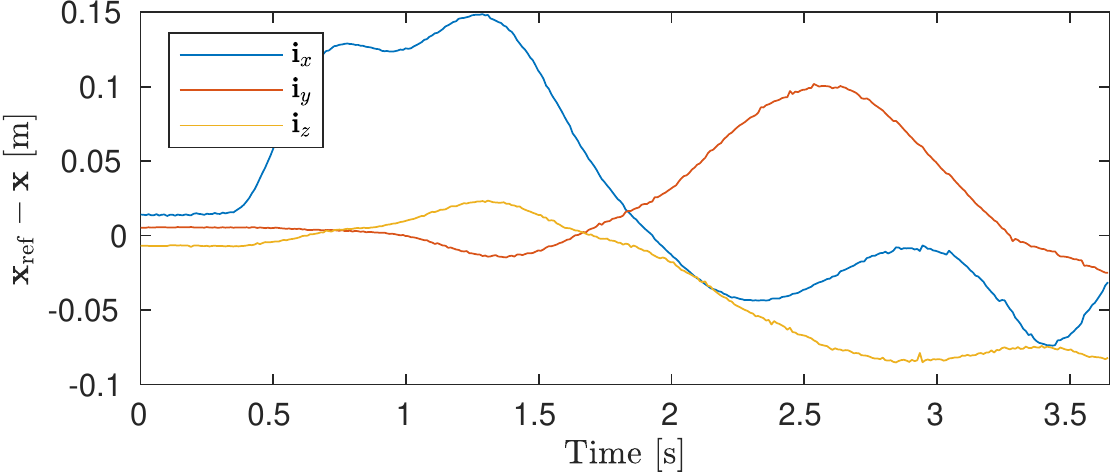}
		\caption{Position tracking error.}
		\label{fig:trans3_poserror}
	\end{subfigure}%
	\quad
	\begin{subfigure}[t]{0.485\textwidth}
		\centering
		\includegraphics[width=\linewidth]{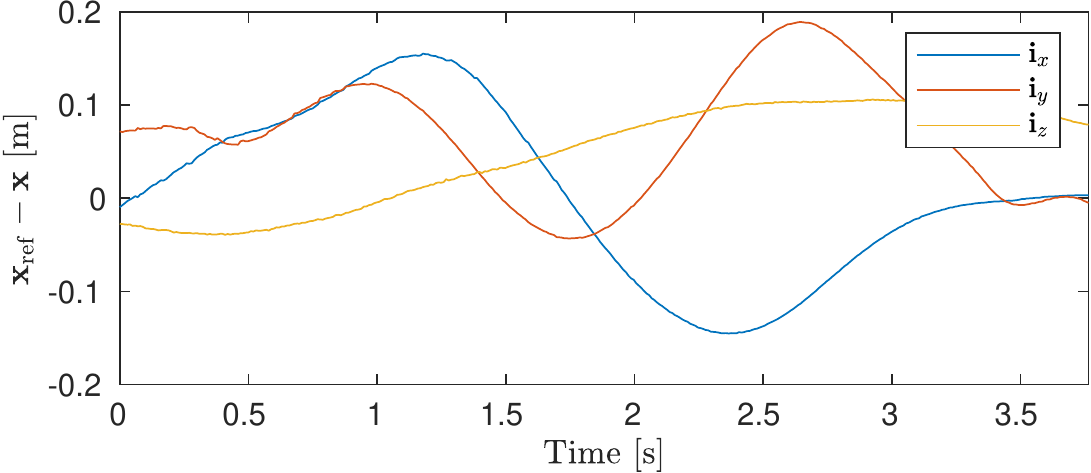}
		\caption{Position tracking error.}
		\label{fig:trans3out_poserror}
	\end{subfigure}%
	
	\begin{subfigure}[t]{0.485\textwidth}
		\centering
		\includegraphics[width=\linewidth]{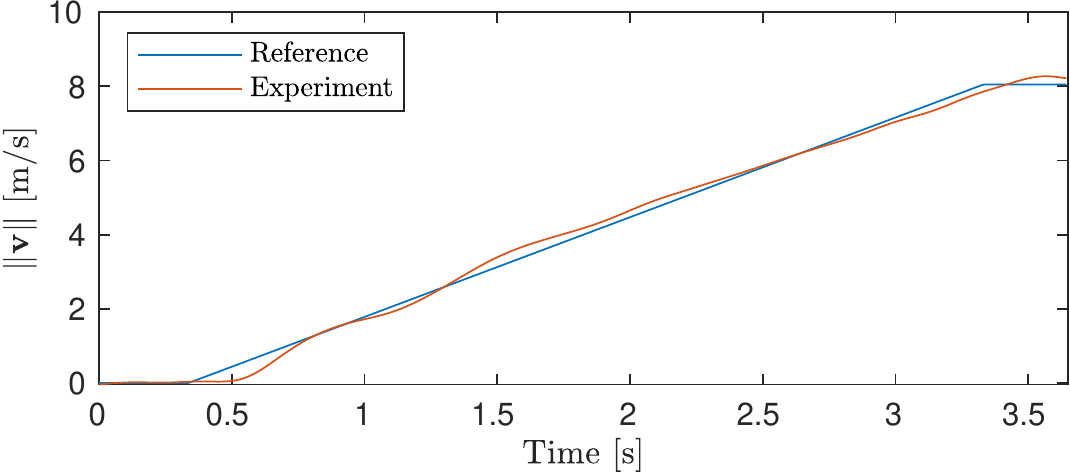}
		\caption{Speed.}
		\label{fig:trans3_speed}
	\end{subfigure}%
	\quad
	\begin{subfigure}[t]{0.485\textwidth}
		\centering
		\includegraphics[width=\linewidth]{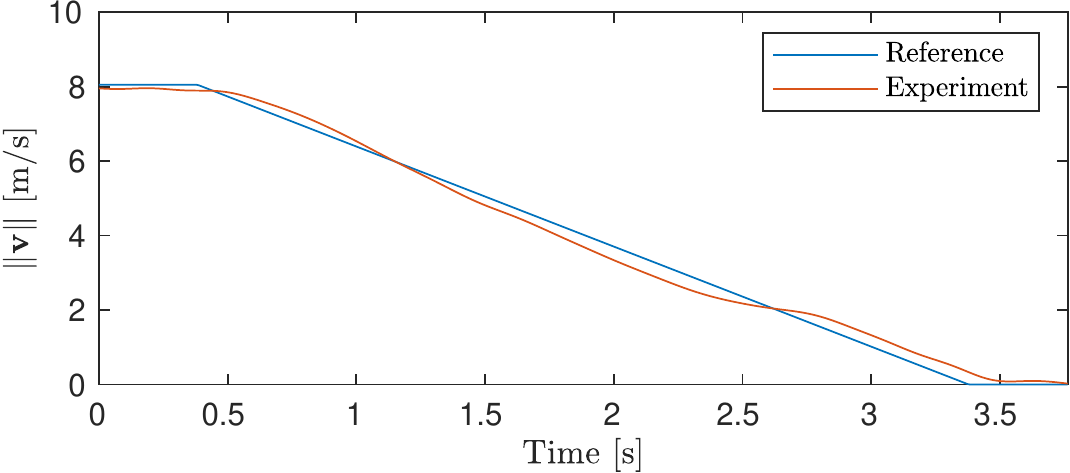}
		\caption{Speed.}
		\label{fig:trans3out_speed}
	\end{subfigure}%

	\begin{subfigure}[t]{0.485\textwidth}
		\centering
		\includegraphics[width=\linewidth]{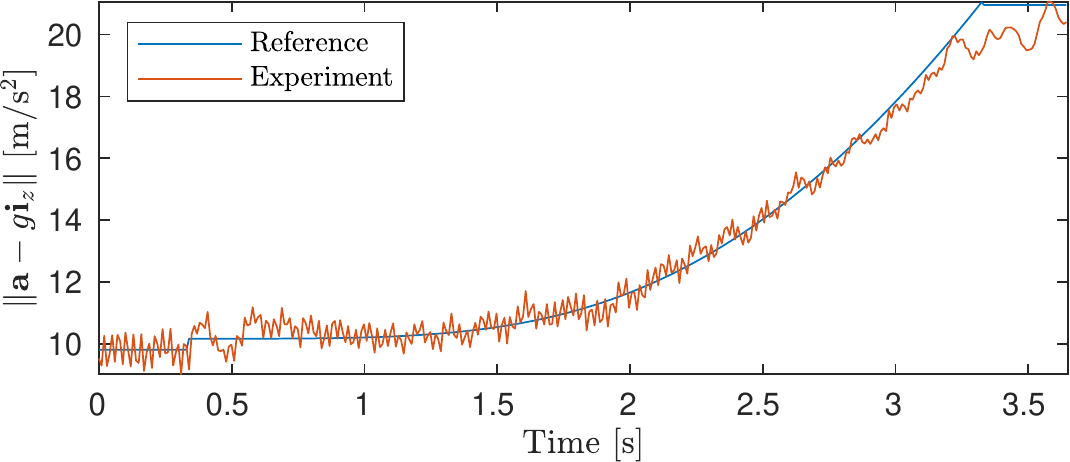}
		\caption{Acceleration.}
		\label{fig:trans3_accel}
	\end{subfigure}%
	\quad
	\begin{subfigure}[t]{0.485\textwidth}
		\centering
		\includegraphics[width=\linewidth]{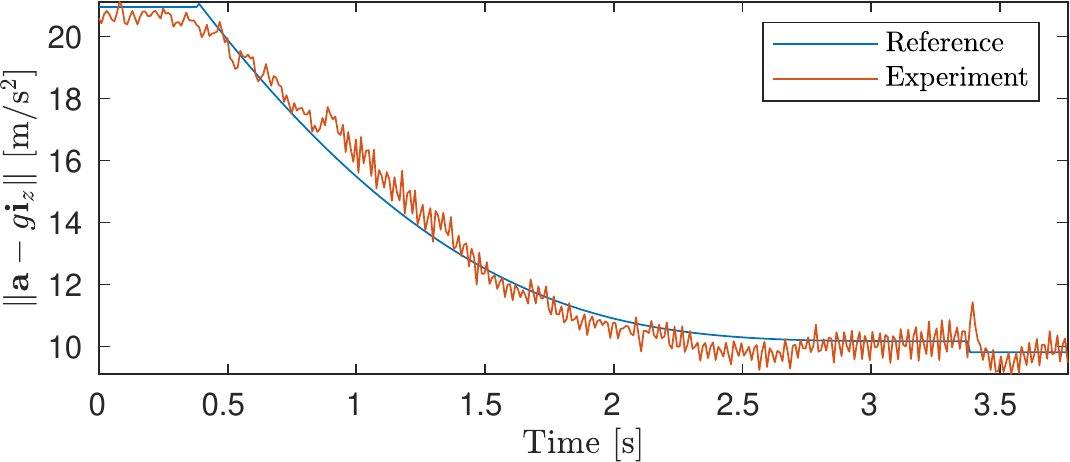}
		\caption{Acceleration.}
		\label{fig:trans3out_accel}
	\end{subfigure}%
	
	\begin{subfigure}[t]{0.485\textwidth}
		\centering
		\includegraphics[width=\linewidth]{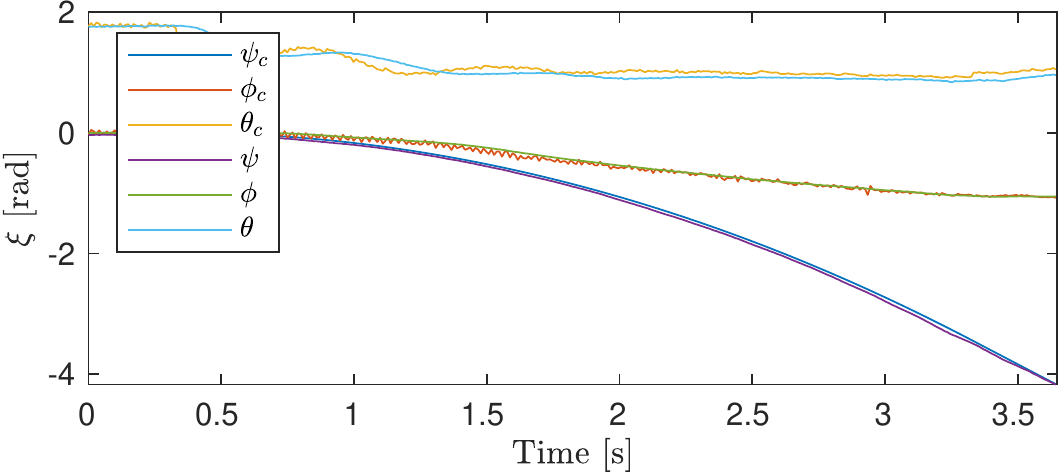}
		\caption{Attitude.}
		\label{fig:trans3_attitude}
	\end{subfigure}%
	\quad
	\begin{subfigure}[t]{0.485\textwidth}
		\centering
		\includegraphics[width=\linewidth]{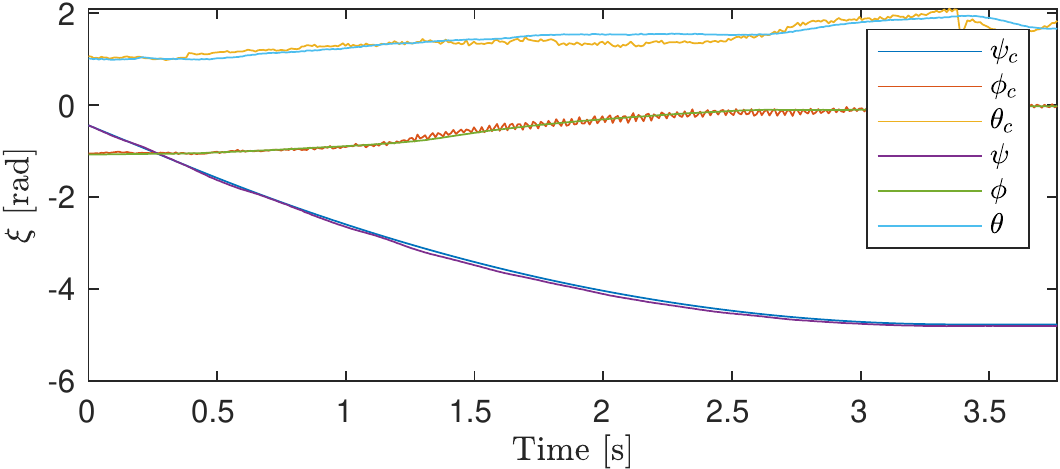}
		\caption{Attitude.}
		\label{fig:trans3out_attitude}
	\end{subfigure}%

	\caption{Experimental results for transition on a circle with radius 3.5 m from static hover to coordinated flight at 8 m/s (a), (c), (e), (g), and (i); and vice versa (b), (d), (f), (h), and (j).}
	\label{fig:trans3}
\end{figure*}

Figure \ref{fig:trans3} shows experimental results for transitions between static hover and coordinated flight at 8 m/s on a circular trajectory with 3.5 m radius.
Each transition takes 3 s at a constant tangential acceleration of 2.7 m/s$^2$ and is completed in 12 m.
The pitch angle varies over a range of 64 deg.
While transitioning from and to hover, the controller tracks the circle trajectory with respectively 10 cm and 15 cm RMS, and 15 cm and 24 cm maximum position error.
These maneuvers demonstrate that the controller is capable of performing aggressive transitions while simultaneously tracking turns with significant acceleration.

To evaluate the benefits of the feedforward input based on jerk and yaw rate, we also attempted to fly the same transitions without the angular velocity reference, \ie, with $\vect{\Omega}_{\sref} = \vect{0}$ in \eqref{eq:tsPD}.
We found that the controller is still able to perform 3 s transitions to speeds up to 3 m/s.
However, if the target speed is increased and the corresponding tangential acceleration exceeds 1 m/s$^2$, the absence of the feedforward term causes large deviations from the reference trajectory to the point where the vehicle cannot be stabilized anymore.
This confirms the benefit of jerk and yaw rate tracking when flying aggressive maneuvers.
Intuitively, the feedforward input enables the control to anticipate future accelerations by regulating not only the forces acting on the vehicle but also their temporal derivative.

\subsection{Differential Thrust Turning}\label{sec:diffthrustturn_exp}
\begin{figure*}
	\centering
	
	\begin{subfigure}[t]{0.5\textwidth}
		\centering
		\includegraphics[width=\linewidth]{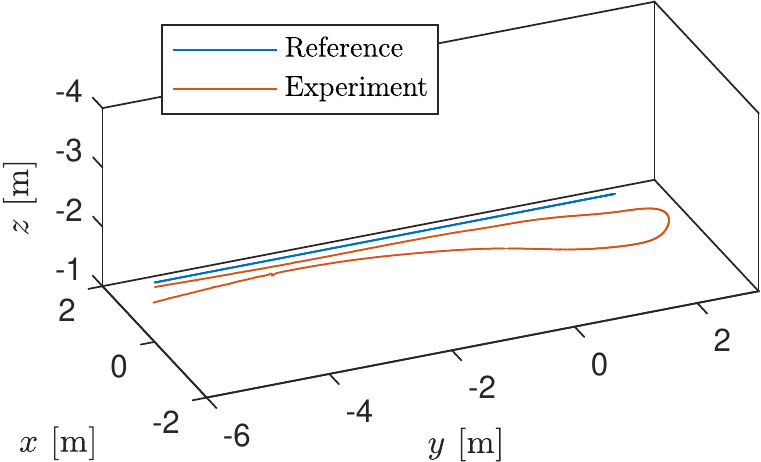}
		\caption{Position.}
		\label{fig:diffturn_position}
	\end{subfigure}%
	
	\begin{subfigure}[t]{0.485\textwidth}
		\centering
		\includegraphics[width=\linewidth]{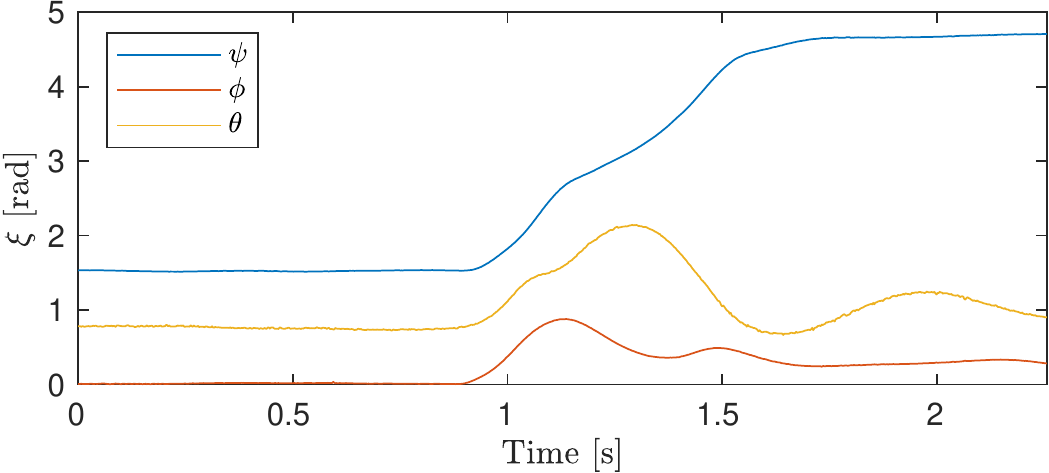}
		\caption{Attitude.}
		\label{fig:diffturn_attitude}
	\end{subfigure}%
	\quad
	\begin{subfigure}[t]{0.485\textwidth}
		\centering
		\includegraphics[width=\linewidth]{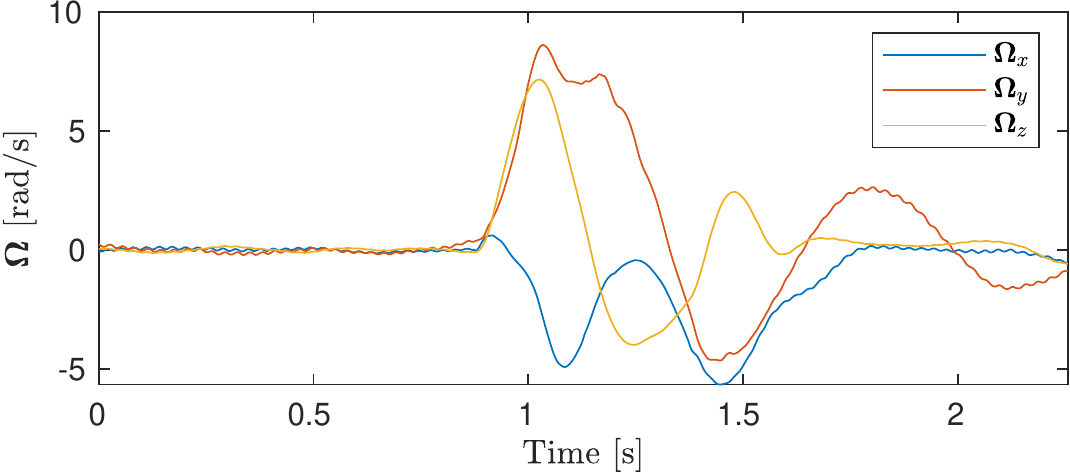}
		\caption{Angular velocity.}
		\label{fig:diffturn_angrate}
	\end{subfigure}
	
	\begin{subfigure}[t]{0.485\textwidth}
		\centering
		\includegraphics[width=\linewidth]{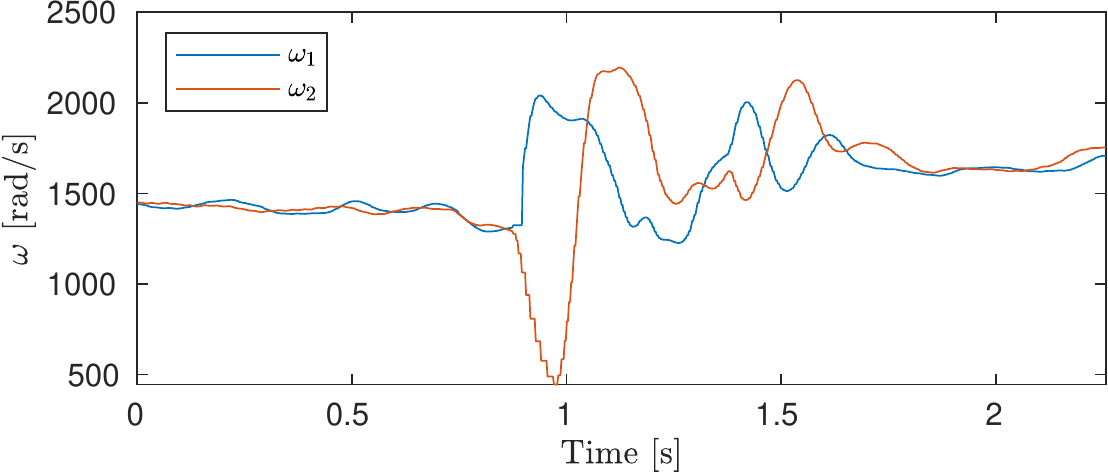}
		\caption{Rotor speed.}
		\label{fig:diffturn_rpm}
	\end{subfigure}%
	\quad
	\begin{subfigure}[t]{0.485\textwidth}
		\centering
		\includegraphics[width=\linewidth]{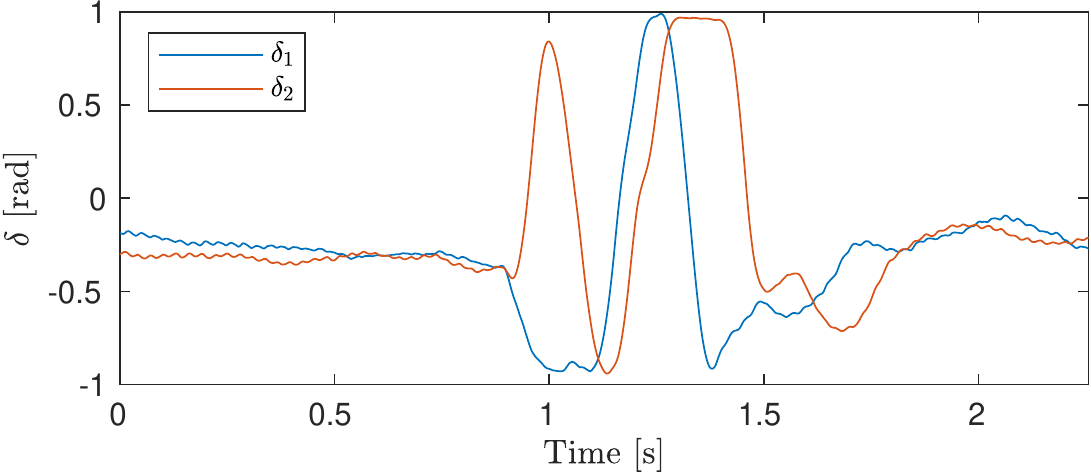}
		\caption{Flap deflection.}
		\label{fig:diffturn_flap}
	\end{subfigure}
	\caption{Experimental results for differential thrust turn.}
	\label{fig:diffturn}
\end{figure*}

Since the controller is not restricted to coordinated flight, it can perform turns without banking.
The tailsitter aircraft is particularly suitable for quick turns using yaw, because of the absence of any vertical surfaces and the availability of relatively powerful motors.
Figure \ref{fig:diffturn} shows a fast turn that is executed using differential thrust.
The reference trajectory, entered in coordinated flight at 7 m/s, changes direction without deviating from a straight line.
The controller responds with large differential thrust and flap deflections.
At the onset of the turn, both flaps are almost fully deflected in opposite directions and the motors produce a differential thrust of 6.1 N.
This causes the aircraft to turn at a maximum rate of 650 deg/s and point in the opposite direction within half a second, while remaining within 1 m of the position reference.

\section{Conclusion}\label{sec:tscontrol_conclusion}
\revised{This paper introduced a novel control design aimed at tracking agile trajectories using a tailsitter flying wing.
It was shown that combining $\varphi$-theory with robust incremental control enables effective global control that does not depend on extensive aerodynamic modeling.
In fact, the proposed controller achieves accurate tracking of challenging maneuvers using only coarse aerodynamic parameters, estimated from limited flight test data.
Hence, it has great potential for application to aircraft with (partially) unknown aerodynamic properties, allowing quick iteration of vehicle and control designs.

The current controller has some limitations that may be addressed to improve suitability for widespread deployment.
Firstly, the aerodynamic model assumes there are no wind or gusts.
It only considers the vehicle attitude and velocity with regard to the world-fixed reference frame, not considering the aerodynamic angles with regard to the freestream velocity.
While incremental control counteracts the resulting unmodeled force and moment to some extent, it would likely be beneficial to account for wind and gusts in the control design~\cite{pfeifle2021cascaded}.
Additionally, the definition of the yaw reference in the world-fixed frame does not readily transform into a coordinated flight constraint in windy conditions.
Thus, it may be practical to add an option to switch to a sideslip angle reference instead of the yaw reference.
Secondly, the controller assumes that the reference trajectory is dynamically feasible.
If this assumption is violated by the trajectory planner, saturation and clipping of the control inputs may occur, potentially resulting in loss of control.
An extension of the proposed controller that optimizes control allocation in case of saturation may be used to prioritize controlled and stable flight, while \rrevised{temporarily} sacrificing trajectory tracking accuracy.

Future work may include further exploration of the flight envelope in outdoor flights, including increased speeds that are unattainable in the constrained indoor space.
This may require consideration of wind, as described above, as well as reconfiguration of the test platform to add satellite navigation and airspeed sensors.
An extension to more conventional unmanned aircraft, \ie, with fuselage, tail, and vertical stabilizer, may also be considered.
Feedforward jerk and yaw rate references could significantly improve tracking of agile maneuvers on these aircraft, similarly to our findings for the tailsitter flying wing.}

\FloatBarrier
\appendices
\section{Angular Acceleration}\label{app:snap}
In this appendix, we derive an expression for the angular acceleration as a function of reference snap, \ie, the fourth temporal derivative of position, and yaw acceleration.
The transform can be used to obtain an angular acceleration feedforward input.
Along with the expressions given in Section \ref{sec:invdyn}, it can also be used to obtain the control inputs, \ie, the motor speeds and flap deflections, required to track a given trajectory $\vect{\sigma}_{\sref}(t)$.

By taking the derivative of \eqref{eq:phidot}, we obtain the following expression for the roll acceleration
\begin{multline}\label{eq:ddotphi}
\ddot{\phi} = -{\left(\beta_x^2+\beta_z^2\right)^{-2}}\left(\left(\ddot\beta_x\beta_z-\beta_x\ddot\beta_z\right)\left(\beta_x^2+\beta_z^2\right)\right.\\ 
\left. - \left(\dot\beta_x\beta_z-\beta_x\dot\beta_z\right)\left(2\beta_x\dot\beta_x+2\beta_z\dot\beta_z\right)\right),
\end{multline}
where $\beta_x$ and $\beta_z$ are respectively the first and second arguments of the $\atan2$ function in \eqref{eq:roll}.
Their first derivatives are given by \eqref{eq:betadot} and \eqref{eq:beta2dot}.
We obtain the second derivatives from these expressions as
\begin{align}
\begin{split}
\ddot \beta_x &= \left(\ssin{\psi}{\dot{\psi}}^2 - \ccos{\psi}\ddot\psi\right)\xelem\vect{f}^i - 2 \ccos{\psi} \dot\psi \xelem \dot{\vect{f}}^i - \ssin{\psi}\xelem \ddot{\vect{f}}^i\\
&\;\;\;\;\;- \left(\ccos{\psi}{\dot \psi}^2 + \ssin{\psi}\ddot \psi\right)\yelem \vect{f}^i -2 \ssin{\psi}\dot\psi\yelem\dot{\vect{f}}^i +\ccos{\psi}\yelem\ddot{\vect{f}}^i,
\end{split}
	\\
	\ddot \beta_z &= \zelem \ddot{\vect{f}}^i,
\end{align}
where the second force derivative is a function of snap
\begin{equation}
\ddot{\vect{f}}^i = m\vect{s}.
\end{equation}
Similarly, by taking the derivative of \eqref{eq:dottheta} we obtain the pitch acceleration
\begin{multline}\label{eq:ddottheta}
\ddot \theta = {\left(\sigma_x^2+\sigma_z^2\right)^{-2}}\bigl(
\left(\ddot\sigma_x\sigma_z-\sigma_x\ddot\sigma_z\right)\left(\sigma_x^2+\sigma_z^2\right)\\
-\left(\dot\sigma_x\sigma_z-\sigma_x\dot\sigma_z\right)\left(2\sigma_x\dot\sigma_x+2\sigma_z\dot\sigma_z\right)\bigr),
\end{multline}
where $\sigma_x$ and $\sigma_z$ are respectively the first and second arguments of the $\atan2$ function in \eqref{eq:theta}.
Their first derivatives are given by \eqref{eq:sigma1dot} and \eqref{eq:sigma2dot}, and their second derivatives are
\begin{align}
\ddot\sigma_x &= \eta \left(\xelem\ddot{\vect{f}}^\phi + c_{D_V} \dot{\tau}_x \right) - c^\delta_{L_V}  \delta \dot{\tau}_x - c_{L_V}\dot{\tau}_z - \zelem\ddot{\vect{f}}^\phi,\label{eq:sigma1ddot}\\
\ddot\sigma_z &= \eta \left(\zelem\ddot{\vect{f}}^\phi + c_{D_V} \dot{\tau}_z \right) - c^\delta_{L_V}  \delta \dot{\tau}_z + c_{L_V}\dot{\tau}_x + \xelem\ddot{\vect{f}}^\phi\label{eq:sigma2ddot}
\end{align}
with
\begin{align}
\dot{\tau}_x &= \ddot{\|\vect{v}\|} \xelem \vect{v}^\phi + 2\dot{\|\vect{v}\|} \xelem \dot{\vect{v}}^\phi + \|\vect{v}\|\xelem\ddot{\vect{v}}^\phi,\\
\dot{\tau}_z &= \ddot{\|\vect{v}\|} \zelem \vect{v}^\phi + 2\dot{\|\vect{v}\|} \zelem \dot{\vect{v}}^\phi + \|\vect{v}\|\zelem\ddot{\vect{v}}^\phi
\end{align}
and
\begin{align}
\ddot{\|\vect{v}\|} &= \frac{\vect{a}^\top\vect{a} + \vect{v}^\top\vect{j}}{\|\vect{v}\|} - \frac{\vect{v}^\top\vect{a} \dot{\|\vect{v}\|}}{\|\vect{v}\|^2},\\
\ddot{\vect{v}}^\phi &= \ddot{\vect{R}}^\phi_i\vect{v} + 2\dot{\vect{R}}^\phi_i\vect{a} + \vect{R}^\phi_i\vect{j}.\label{eq:jerk}
\end{align}
The expression for the force second derivative $\ddot{\vect{f}}^\phi$ is similar to \eqref{eq:jerk}.
In the force equations, we assume that the flap deflection is known and that its temporal derivatives are negligible, as described in Section \ref{sec:diff_attthrust}.
We combine the roll acceleration and pitch acceleration obtained from respectively \eqref{eq:ddotphi} and \eqref{eq:ddottheta} with the yaw acceleration $\ddot{\psi}$ to obtain the angular acceleration in the body-fixed reference frame.
We take the derivative of \eqref{eq:angrate} to obtain the following expression:
\begin{equation}\label{eq:angacc}
\dot{\vect{\Omega}} = \left[\begin{array}{c}
0\\\ddot \theta\\0
\end{array}\right] + \dot{\vect{R}}^{ \theta}_\phi \left[\begin{array}{c}
\dot \phi \\0\\0
\end{array}\right] + \vect{R}^{ \theta}_\phi \left[\begin{array}{c}
\ddot \phi \\0\\0
\end{array}\right] + \dot{\vect{R}}^{ \theta}_\psi \left[\begin{array}{c}
0\\0\\\dot \psi
\end{array}\right] + \vect{R}^{ \theta}_\psi \left[\begin{array}{c}
0\\0\\\ddot \psi
\end{array}\right].
\end{equation}

We can now find the aerodynamic and thrust moment in the body-fixed reference frame by rewriting \eqref{eq:Omegadot}, as follows:
\begin{equation}\label{eq:moment}
\vect{m} = \vect{J}\dot{\vect{\Omega}} + \vect{\Omega}\times \vect{J}\vect{\Omega} - \vect{m}_{\ext}.
\end{equation}
In practice, the unmodeled moment $\vect{m}_{\ext}$ is implicitly estimated and corrected for by incremental control, as described in Section \ref{sec:angacc_control}.
Based on the moment obtained from \eqref{eq:moment} and the total thrust obtained from \eqref{eq:T}, the required control inputs can be calculated using the expressions given in Section \ref{sec:diffcontrol}.
Note that---since the aerodynamic and thrust moment cannot instantaneously change---dynamic feasibility of $\vect{\sigma}_{\sref}$ requires continuity of \eqref{eq:moment}, and thereby at least fourth-order continuity of the position reference $\vect{x}_{\sref}$ and at least second-order continuity of the yaw reference $\psi_{\sref}$.
\section{Comparison to Baseline Controller}\label{sec:baselinecomp}
In order to experimentally demonstrate the advantages offered by two key aspects of our flight control design, we compare our proposed controller to a baseline version.
The baseline controller is identical to the proposed controller, except that (i) it does not incorporate feedforward jerk and yaw rate tracking, and (ii) it does not incrementally update the control inputs.
More concretely, we set the feedforward angular rate reference $\vect{\Omega}_{\sref}$ to zero in \eqref{eq:tsPD}, and we update the force and moment commands by direct inversion of \eqref{eq:vdot} and \eqref{eq:Omegadot}, respectively.
These commands are thus obtained as
\begin{align}\label{eq:nonincr}
\vect{f}^i_c &= m \left(\vect{a}_c - g \vect{i}_z\right),\\
\vect{m}_c &= \vect{J}\dot{\vect{\Omega}}_c + \vect{\Omega}_{\lpf} \times \vect{J}\vect{\Omega}_{\lpf},
\end{align}
in contrast to the incremental updates \eqref{eq:force_command} and \eqref{moment_command} used in our proposed controller.
We add integral action to the attitude controller \eqref{eq:tsPD} to improve the rejection of modeling errors and external disturbances in the absence of incremental control updates.
The baseline attitude controller is thus given by
\begin{equation}\label{eq:att_PID}
\dot{\vect{\Omega}}_c = \vect{K}_{\vect{\xi}} \vect{\zeta}_e - \vect{K}_{\vect{\Omega}} \vect{\Omega}_{\lpf} + {\vect{K}_{I_\xi}}\int \vect{\zeta}_e \;\;d t.
\end{equation}

Table \ref{tab:comp} gives key properties of the baseline controller, our proposed controller, and several tailsitter control designs presented in recent literature.
In order to enable comparison, we only include guidance and control designs that fully govern both translational and rotational motion, \ie, we omit works that consider only hover, only longitudinal control, only rotational control etc.
Also not included are control designs for (over-actuated) aircraft that possess actuators beyond the configuration we consider (\ie, two flaps and two rotors), since these aircraft typically present fundamentally different challenges in control design.

\begin{table*}[!ht]
	\centering
	\caption{Comparison of tailsitter flight control algorithms.}
	\footnotesize
	\label{tab:comp}
	\begin{tabular}{lllll}
		\hline
		&Methodology & Aerodynamic model & Robustification & Feedforward \\
		\hline
		Proposed & \makecell[lt]{INDI, and\\differential flatness} & Global $\varphi$-theory & Incremental & \makecell[lt]{Acceleration,\\ jerk, and\\ yaw rate}\\
		Baseline & Dynamic inversion & Global $\varphi$-theory & - & Acceleration\\
		\hline
		Lustosa, 2017 \cite{lustosa2017varphi} & Scheduled LQR & Linearized $\varphi$-theory & - & \makecell[lt]{Attitude, yaw\\ rate (at trim\\ points)}\\
		Ritz, 2017 \cite{ritz2017global} & \makecell[lt]{Dynamic inversion\\ (coordinated flight)} & Global first principles & - & Acceleration\\
		Chiappinelli, 2018 \cite{chiappinelli2018modeling} & PD & \makecell[lt]{Control effectiveness\\(moment only)} & - & Attitude\\
		Smeur, 2020 \cite{smeur2020incremental} & INDI & \makecell[lt]{Control effectiveness\\ (coordinated flight at\\ small flight path angle)}& Incremental & -\\	
		Barth, 2020 \cite{barth2020model} & \makecell[lt]{Model-free control\\ (coordinated flight)}& - & Model free & -\\
		\hline
	\end{tabular}
\end{table*}

While the baseline controller lacks key aspects of our proposed control design, it is still a state-of-the-art dynamic inversion controller based on a global aerodynamics model that incorporates the acceleration feedforward signal.
We note that none of the existing algorithms listed in the table combine robust control with feedforward control inputs to the same extent as our proposed controller.
Controllers that do not incorporate any robustification require an accurate aerodynamics model that covers the entire flight envelope or will incur significant systematic tracking errors.
Several algorithms include attitude or (equivalently) acceleration feedforward inputs, but none consider reference jerk, which is essential for accurate tracking of fast-changing accelerations.
Without the angular rate feedforward input corresponding to jerk, the derivative term of the attitude controller will counteract the nonzero angular rate required to track a dynamic attitude reference.
Finally, we note that existing controllers are often limited to coordinated flight because of limitations of the aerodynamic model or the control algorithm itself.
In typical cruise conditions, this limitation is not prohibitive, but when performing agile maneuvers it restricts the usable flight envelope, \eg, excluding knife edge flight.

\begin{figure*}
	\centering
	
	\begin{subfigure}[t]{0.295\textwidth}
		\centering
		\includegraphics[width=\linewidth]{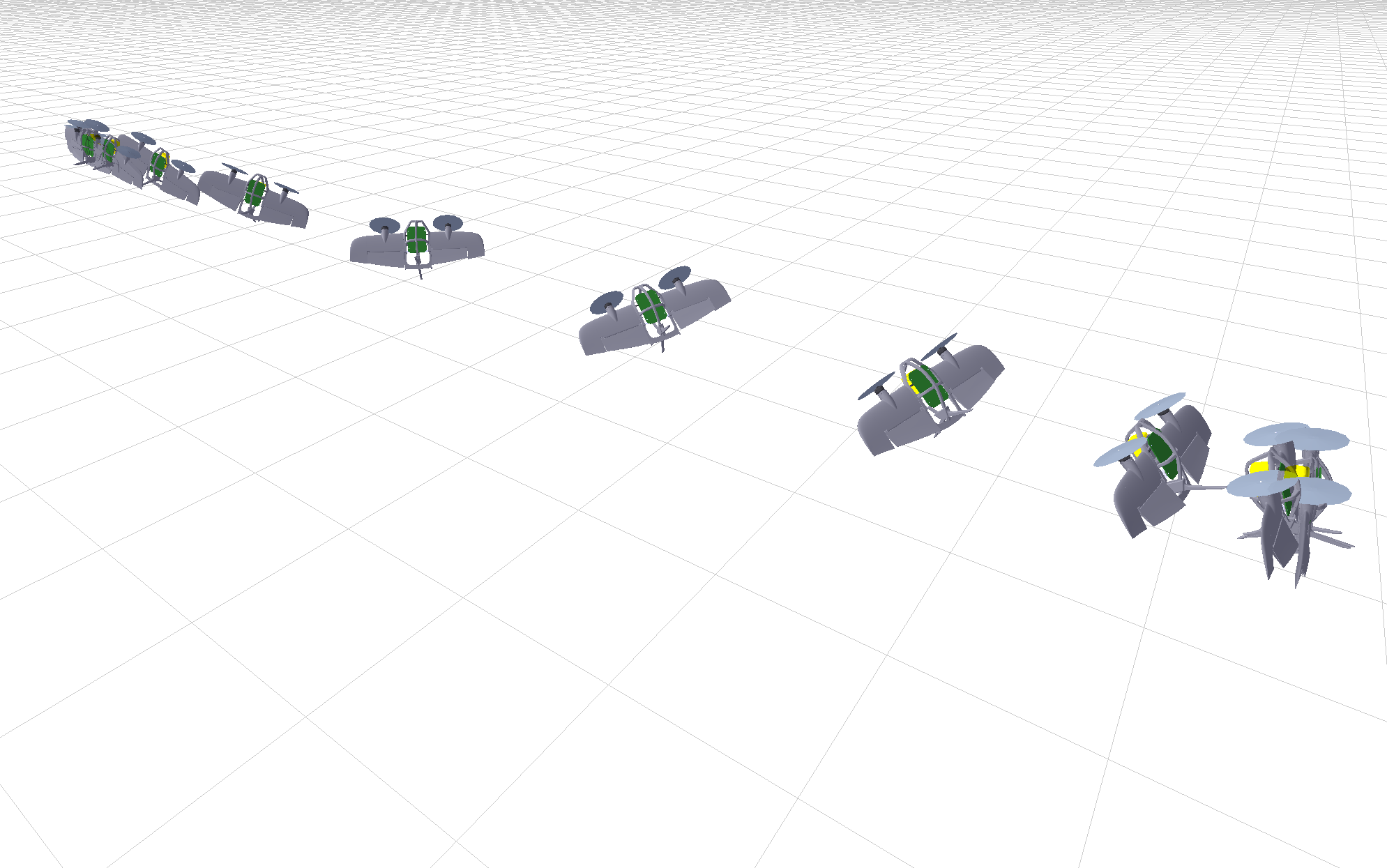}
		\caption{Hover-to-hover trajectory with $\nicefrac{\pi}{2}$ rad yaw.}
		\label{fig:comp_p2p}
	\end{subfigure}%
	\quad
	\begin{subfigure}[t]{0.34\textwidth}
		\centering
		\includegraphics[width=\linewidth]{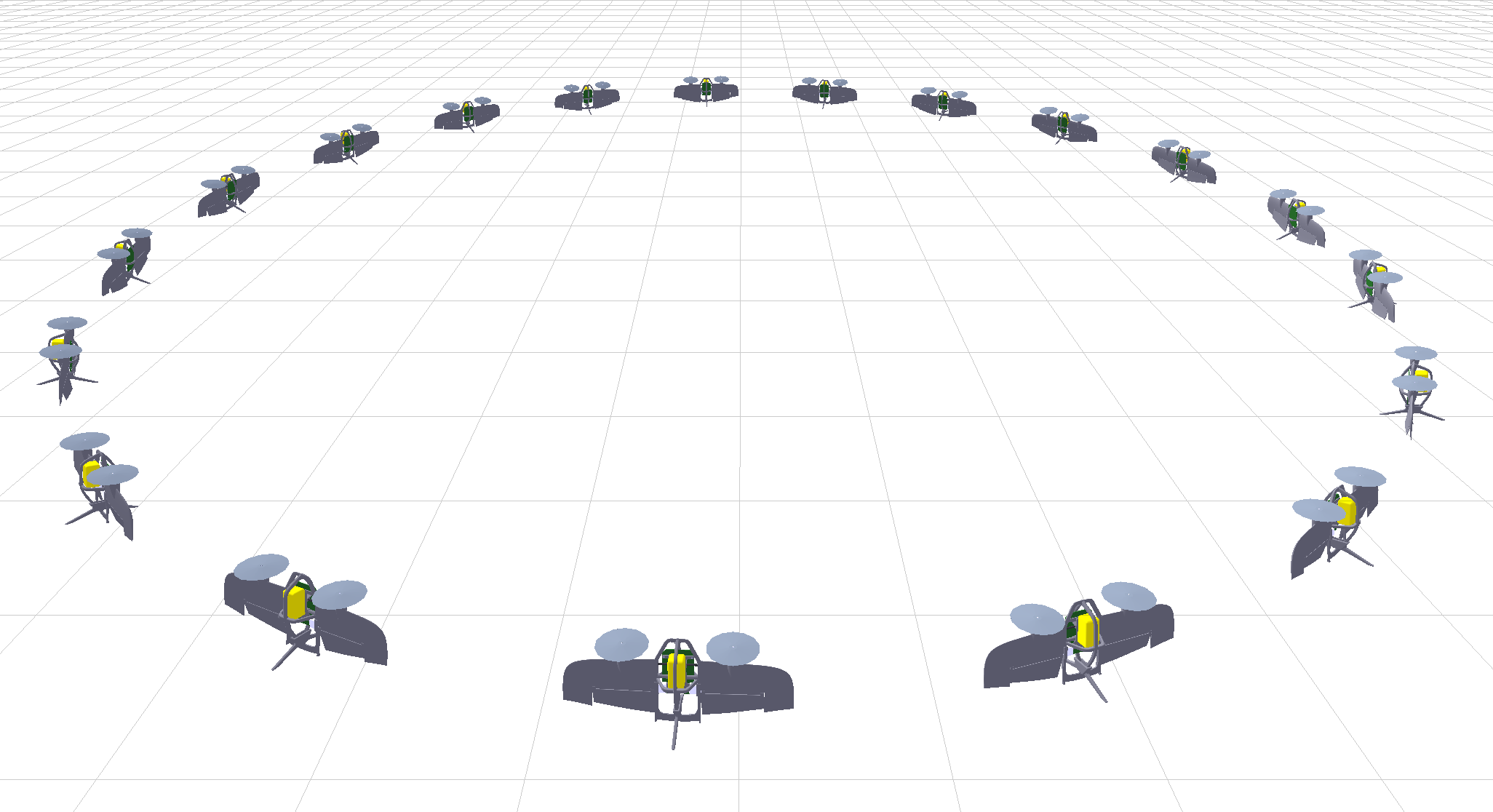}
		\caption{Circle trajectory in knife edge orientation.}
		\label{fig:comp_ke}
	\end{subfigure}%
	\quad
	\begin{subfigure}[t]{0.29\textwidth}
		\centering
		\includegraphics[width=\linewidth]{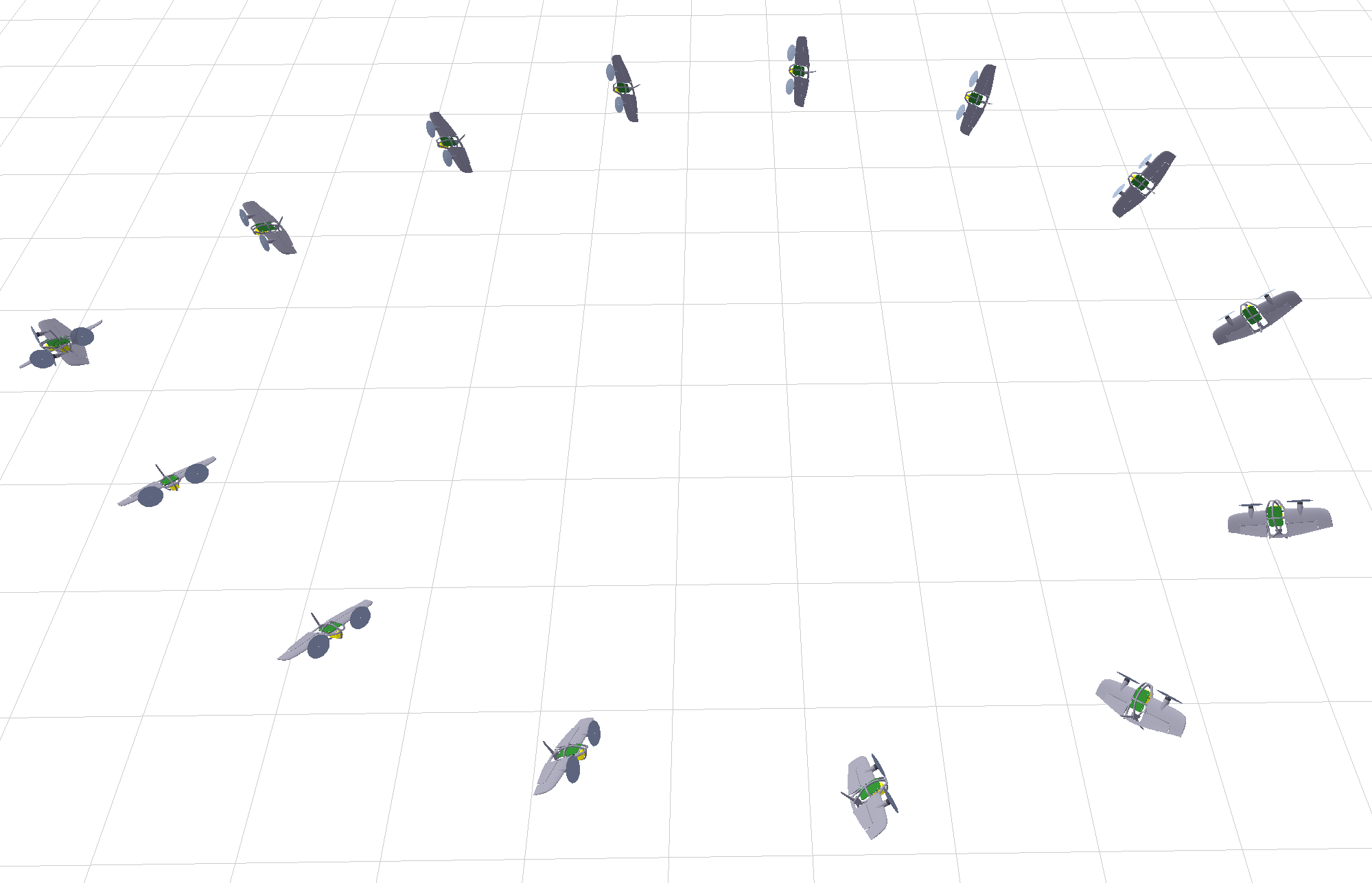}
		\caption{Coordinated flight on lemniscate trajectory (partially shown).}
		\label{fig:comp_lemniscate}
	\end{subfigure}%
	\caption{Reference trajectories with attitude and flap deflections obtained from differential flatness transform.}
	\label{fig:comp_traj}
\end{figure*}

In order to examine the combined and individual effects of incremental control and feedforward tracking, we employ two additional versions of the baseline controller:
baseline+FF includes feedforward tracking but no incremental control, while baseline+INDI includes incremental control but no feedforward tracking.
Table \ref{tab:comp_perf} presents tracking performance data for the proposed and baseline controllers on the three trajectories shown in Fig. \ref{fig:comp_traj}.

The hover-to-hover trajectory includes transitions and sideways flight.
When flown slowly (\ie, 5 s), the maximum acceleration of 2.3 m/s\textsuperscript{2} is similar to transition maneuvers found in literature.
All controllers track this slow trajectory with reasonable error.
At low speeds, incremental control already significantly reduces the error by accounting for unmodeled dynamics, while feedforward makes less difference as the angular rates remain low.
As expected, the importance of feedforward increases as the maneuver speeds up, evidenced by a dramatic increase in tracking error for the baseline and baseline+INDI controllers.
The error growth is much smaller for the proposed and baseline+FF controllers, confirming that jerk and yaw rate tracking is essential for accurate tracking of aggressive maneuvers.
As far as we are aware, our proposed control design is the first tailsitter controller to incorporate these feedforward inputs to enable agile maneuvers.

The importance of feedforward is further demonstrated on the knife edge circle trajectory.
We found that the baseline and baseline+INDI controllers are unable to accurately maintain knife edge orientation due to absence of the 76 deg/s yaw rate feedforward reference.
As knife edge flight is inherently unstable, failing yaw control leads to an overall failure to stabilize the vehicle.
The baseline+FF controller performs much better, but---due to the lack of incremental control---still incurs a significantly larger systematic tracking error than the proposed controller.

The lemniscate trajectory is flown at higher speed, where the controller increasingly relies on the aerodynamics model for accurate application of force and moment.
We found that the baseline controllers without INDI are unable to avoid crashing before the trajectory speed is reached.
This may be due to inaccurate aerodynamics parameters or due to significant aerodynamic effects that are not captured by the simplified dynamics model.
The baseline+INDI controller corrects for these modeling inaccuracies, and is able to avoid crashes.
However, it incurs large tracking error---due to the lack of jerk and yaw rate tracking---to the point that we were unable to obtain an error measurement because of flight space size limitations.

\begin{table*}[!ht]
	\centering
	\caption{Position and yaw tracking error for proposed and baseline controllers. The top three rows (corresponding to hover-to-hover trajectories) contain maximum values, while the bottom two rows (corresponding to periodic trajectories) contain RMS values.}
	\footnotesize
	\label{tab:comp_perf}
	\begin{tabular}{lll|rrrr|rrrr}
		\multicolumn{3}{c|}{}&\multicolumn{4}{c|}{$\|\vect{x}-\vect{x}_{\sref}\|_2$ [cm]} & \multicolumn{4}{c}{$|\psi-\psi_{\sref}|$ [deg]}\\
		& \rotatebox{90}{$\|\vect{v}_{\sref}\|_2$ [m/s]}& \rotatebox{90}{$\|\vect{a}_{\sref}\|_2$ [m/s$^2$]}& \rotatebox{90}{Proposed} & \rotatebox{90}{Baseline} & \rotatebox{90}{Baseline+FF} & \rotatebox{90}{Baseline+INDI} & \rotatebox{90}{Proposed} & \rotatebox{90}{Baseline} & \rotatebox{90}{Baseline+FF} & \rotatebox{90}{Baseline+INDI}\\
		\hline
		Hover-to-hover (6 m, $\nicefrac{\pi}{2}$ rad, 5 s) & 3.0 & 2.3 & \textbf{7.4}&41.2 & 31.9 & 17.4 & \textbf{1.3} &21.7 &13.6&8.4\\
		Hover-to-hover (6 m, $\nicefrac{\pi}{2}$ rad, 4 s) & 3.7 & 3.5 & \textbf{15.5}&63.0 & 33.8 & 34.9 & \textbf{2.0 }&21.5 &19.8&10.1\\
		Hover-to-hover (6 m, $\nicefrac{\pi}{2}$ rad, 3 s) & 4.9 & 6.2 &\textbf{23.3} & $>$400 & 40.4 & 64.7 & \textbf{10.4} & $>$25 &17.4& 20.6\\
		Knife edge circle (3 m radius) & 4.0 & 5.3 & \textbf{2.8} &  - &8.2 & - &\textbf{0.6}&-&1.8&-\\
		Lemniscate & 6.0 & 9.1 & \textbf{16.6} & - & - & $>$200 & \textbf{2.8} &-& - &$>$25\\
		\hline
	\end{tabular}
\end{table*}

\section*{Acknowledgments}
We thank Murat Bronz and John Aleman for the design, fabrication and assembly of the aircraft, and we thank Lukas Lao Beyer and Nadya Balabanska for the implementation of a flight dynamics simulation tool. This work was supported by the Army Research Office through grant W911NF1910322.

\FloatBarrier

\bibliographystyle{IEEEtran}
\bibliography{refs}

\end{document}